\documentclass{article}
\usepackage{iclr2027_conference,times}

\usepackage{amsmath,amsfonts,bm}

\def\eqref#1{equation~\ref{#1}}

\def\1{\bm{1}}

\DeclareMathAlphabet{\mathsfit}{\encodingdefault}{\sfdefault}{m}{sl}
\SetMathAlphabet{\mathsfit}{bold}{\encodingdefault}{\sfdefault}{bx}{n}

\usepackage[hidelinks]{hyperref}
\usepackage{url}
\usepackage{booktabs}
\usepackage{graphicx}
\usepackage{float}
\usepackage{amssymb}
\usepackage{xcolor}
\usepackage{multirow}
\usepackage{enumitem}
\usepackage{pifont}
\usepackage{listings}
\usepackage{xspace}
\usepackage{tikz}
\usepackage{fontawesome5}
\usepackage{array}
\usetikzlibrary{fit}

\definecolor{linkblue}{RGB}{0,82,155}

\newcommand{\method}{\textsc{MatLoom}\xspace}
\title{\method: Layered Text-to-Material\\Generation in a Compact Program Space}

\author{Anson Y. Lam, Shuqing Li\textsuperscript{$\ast$} \& Michael R. Lyu \\
Department of Computer Science and Engineering \\
The Chinese University of Hong Kong \\
Hong Kong, China \\
\{yflam1,sqli21,lyu\}@cse.cuhk.edu.hk \\[-2em]
}

\graphicspath{{figures/}}

\definecolor{figtag}{gray}{0.45}
\newlength{\cmpcellW}
\newcommand{\cmpcell}[2]{\begin{tabular}[c]{@{}c@{}}\includegraphics[width=\cmpcellW]{#1}\\[2pt]\includegraphics[width=\cmpcellW]{#2}\end{tabular}}
\newcommand{\bandtag}[1]{\parbox[c][\cmpcellW][c]{3.4em}{\hfill\color{figtag}\scriptsize #1}}
\newcommand{\cmprowlabel}[1]{\parbox[c][\dimexpr 2\cmpcellW+2pt\relax][c]{0.112\linewidth}{\raggedright\scriptsize #1}}
\newcommand{\cmptags}{\begin{tabular}[c]{@{}c@{}}\bandtag{flat}\\[2pt]\bandtag{staged}\end{tabular}}
\newcommand{\teasercell}[2]{%
    \begin{minipage}[t]{0.151\textwidth}%
        \centering
        \includegraphics[width=\linewidth]{#1}\\[0pt]%
        {\scriptsize\setlength{\baselineskip}{7.5pt}#2\par}%
    \end{minipage}%
}

\lstdefinestyle{dsl}{basicstyle=\ttfamily\small,breaklines=true,columns=fullflexible,keepspaces=true}
\newcommand{\dsl}[1]{\texttt{#1}}

\iclrfinalcopy %
\begin{document}

\maketitle

\lhead{}
{\renewcommand{\thefootnote}{\fnsymbol{footnote}}%
    \setcounter{footnote}{1}\footnotetext{Corresponding author.}%
    \setcounter{footnote}{0}}

\begin{center}
    \textcolor{linkblue}{\faHome}~\href{https://yflam1.github.io/matloom/}{\textcolor{linkblue}{\textbf{Project Page}}}
    \quad
    \textcolor{linkblue}{\faGithub}~\href{https://github.com/yflam1/matloom}{\textcolor{linkblue}{\textbf{Code}}}
\end{center}

\begin{abstract}
    Material generation should produce not only an appearance, but also the rules that construct it.
    We introduce \method, a compact, layer-oriented language for text-to-material generation with pretrained language models.
    Each program composes alpha-masked layers whose shared spatial expressions define coverage and physically based rendering (PBR) channels, making dependencies between patterns, color, and relief explicit.
    A standalone interpreter evaluates the program into material maps, while the source retains named fields and layer parameters for subsequent authoring.
    Without task-specific fine-tuning, our pipeline uses parser-guided repair and preview-based critique to revise material designs, then searches noise seeds while keeping each candidate's remaining source fixed.
    On a curated benchmark of $141$ prompts evaluated with six backbones, our best-performing configuration achieves higher mean scores than three diffusion baselines on all four flat-layout prompt-alignment metrics, and the margin persists on seven of the eight metric-layout cells under feasible equal-budget selection including best-of-$10$ sampling with the same quick scorer, with staged CLIPScore a statistical tie. A per-metric oracle bound over the same ten samples, which requires test labels, narrows the judge margins to $6.1$ points flat and a staged statistical tie.
    Its initial programs already exceed all three baselines on mean BLIPScore, before critique or seed search.
    Retained programs have a median length of $21$ lines when pooled across backbones.
    In a blind four-way comparison involving $30$ participants and $20$ prompts, our renders receive $59.2\%$ of choices, compared with $19.3\%$ for the most-preferred baseline.
    Compact executable programs thus offer a way to generate prompt-aligned materials while retaining their construction as part of the asset.
\end{abstract}

\begin{figure}[!ht]
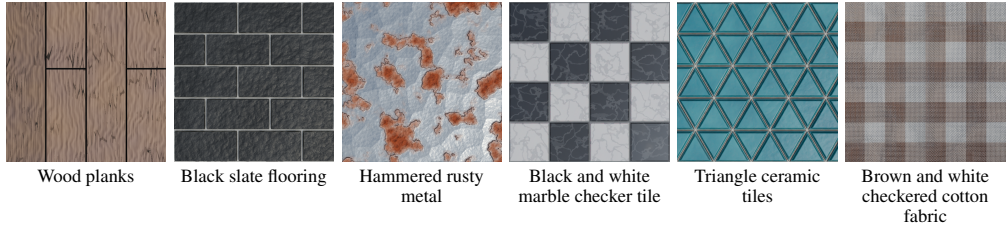

    \centering
    \teasercell{teaser/wood-planks.png}{Wood planks}\hspace{0.008\textwidth}%
    \teasercell{teaser/black-slate-flooring.png}{Black slate flooring}\hspace{0.008\textwidth}%
    \teasercell{teaser/hammered-rusty-metal.png}{Hammered rusty metal}\hspace{0.008\textwidth}%
    \teasercell{teaser/black-and-white-marble-checker-tile.png}{Black and white marble checker tile}\hspace{0.008\textwidth}%
    \teasercell{teaser/triangle-ceramic-tiles.png}{Triangle ceramic tiles}\hspace{0.008\textwidth}%
    \teasercell{teaser/brown-and-white-checkered-cotton-fabric.png}{Brown and white checkered cotton fabric}
    \caption{Selected text-to-material outputs from \method, rendered in the flat layout. Each render comes from a compact executable material program.}
    \label{fig:teaser}
\end{figure}

\section{Introduction}
\label{sec:intro}

Material authoring rarely ends with the first render.
A designer may ask to widen the grout between tiles, roughen a coating, or change a glaze color while preserving the rest of the surface (Figure~\ref{fig:explicit-program-edits}).
Text-to-material diffusion methods generate detailed physically based rendering (PBR) maps~\citep{vecchio2024matfuse,vecchio2024stablematerials,kocsis2025intrinsix}, but their raster outputs do not expose the construction rules behind spatial layout, relief, and reflectance.
Procedural programs and graphs keep those rules as operations and parameters~\citep{guerrero2022matformer,li2024procmatrl,li2025vlmaterial}, making the generated asset easier to inspect, re-evaluate, and revise.

\emph{Can compact material programs serve as an effective output space for pretrained language models, combining text-to-material fidelity with explicit authoring structure?}
This question follows a broader line of work on programs as visual representations~\citep{sharma2018csgnet,jones2020shapeassembly,zhang2024scenelanguage,wu2025chat2svg}.
For materials, the representation must express coupled decisions: a tile lattice defines both coverage and height, a crack has shape and relief, and a coating changes reflectance only where it is applied.
Existing material languages and systems expose different parts of this structure.
MDL supports declarative material definitions and layering~\citep{kettner2015mdl,mdl}; MultiMat serializes Substance graphs in a compact DSL with validation and visual feedback~\citep{belouadi2025multimat}; and Material Apprentice synthesizes procedural materials from text by retrieving expert process traces~\citep{gupta2026apprentice}.
We study a more restricted point in this design space: a layer-oriented field language between a natural-language request and rendered material maps.

We introduce \method, a compact language that represents a material as a stack of alpha-masked layers (Figure~\ref{fig:teaser}).
Each layer assigns PBR channels through expressions over a two-dimensional domain, and named spatial fields can be shared across masks, color, roughness, and height.
This design exposes three authoring decisions to the model: which spatial patterns to define, how layers use those patterns, and which stochastic realization of the noise fields to render.
A standalone interpreter evaluates the program into material maps at a chosen raster resolution, with pinned noise seeds for repeatable execution within a fixed implementation.
The restricted vocabulary trades the breadth of a general graph or shader language for concise programs whose dependencies remain visible in source.

Our synthesis procedure uses that separation to organize inference.
A pretrained language model writes and repairs a program using parser feedback.
A critic then inspects a fast preview, channel statistics, and source code where configured, and revises the material design.
Finally, a text--image scorer selects candidates from the revision trajectory and searches noise seeds while keeping each candidate's other expressions fixed.
The same explicit program is therefore the object of generation, repair, critique, selection, seed exploration, and later rendering.

We evaluate \method on a curated benchmark of $141$ prompts with six language-model backbones and three diffusion baselines.
Across retained programs from all backbones, the median length is $21$ lines.
The best configuration exceeds the baselines in mean score on all four flat-layout alignment metrics, and its initial programs already exceed all baselines on mean BLIPScore before critique or seed search.
In a blind four-way study with $30$ participants and $20$ prompts, \method receives $59.2\%$ of choices, compared with $19.3\%$ for the most-preferred baseline.
These results show that compact executable programs can be competitive for prompt-aligned material generation while preserving construction as part of the asset.
They do not by themselves establish a causal advantage over every procedural representation, so we add matched Blender-Python, budget, and channel controls plus a native Material Apprentice run, leaving MDL and MaterialX untested. \looseness=-1

Our contributions are:
\begin{itemize}[leftmargin=*,itemsep=2pt]
    \item \textbf{A layered authoring representation with a standalone evaluator.} Shared spatial expressions connect coverage and PBR channels through explicit compositing rules, and a standalone interpreter evaluates programs into maps with no digital content creation tool or path tracer in the loop, about $70$ ms per scored preview candidate (Section~\ref{sec:dsl}, Appendix~\ref{app:rep-method-details}).
    \item \textbf{Evidence that the representation carries the pipeline rather than the loop.} Repair, critique, and search follow established components (Section~\ref{sec:related}). We assemble these components around the representation and test that question through shared-backbone Blender-Python controls, an automated edit suite, budget replays, shared rendering channels, and the released Material Apprentice system (Appendices~\ref{app:bpy}, \ref{app:editing}, \ref{app:e2}, \ref{app:budget}, and~\ref{app:ma}).
    \item \textbf{An empirical analysis across backbones, budgets, and rendering channels.} A $141$-prompt benchmark, budget- and channel-controlled comparisons, an edit suite, and a blind study locate where compact programs win and where they do not (Section~\ref{sec:eval}).
\end{itemize}

\section{Related Work}
\label{sec:related}

\paragraph{Image-based material generation.}
Inverse rendering estimates spatially varying reflectance from photographs~\citep{deschaintre2018single,deschaintre2019flexible,guo2020materialgan,lopes2024materialpalette}, and text-conditioned systems synthesize PBR maps from language or other visual inputs~\citep{he2023text2mat,vecchio2024matfuse,vecchio2024stablematerials,kocsis2025intrinsix,luo2026matpedia}.
These raster maps support rendering, relighting, and some map-level editing.
\method instead retains an executable source program, so the construction of the maps can be inspected, re-evaluated, and resampled.
Our diffusion comparisons therefore test rendered appearance against strong raster baselines, not superiority over procedural generators.

\paragraph{Procedural and program-based material generation.}
MATch optimizes existing procedural graphs~\citep{shi2020match}, MatFormer and conditional MatFormer generate graph structure and parameters~\citep{guerrero2022matformer,hu2023generating}, and ProcMatRL improves image-conditioned parameter prediction~\citep{li2024procmatrl}.
VLMaterial generates Blender Python programs from images with a fine-tuned vision-language model~\citep{li2025vlmaterial}.
MultiMat introduces CompactSBS, a compact YAML representation of Substance graphs with visual feedback, validation, and repair, evaluated on image-conditioned and unconditional tasks~\citep{belouadi2025multimat}.
MatLayerNet plans layered aging materials with LLM agents and a curated mask library~\citep{cai2026matlayernet}.
Material Apprentice is closest in task definition because it supports text-to-procedural generation and editing by retrieving expert process traces and compiling them into Blender graphs~\citep{gupta2026apprentice}.
Independently, ShaderAgent writes a compact, validated shader DSL before Blender execution and searches parameters with a fine-tuned scorer~\citep{yuan2026shaderagent}, which supports restricting the program space as a design direction.
\method focuses on a different representation choice: a restricted field vocabulary with ordered alpha-composited layers, standalone execution, explicit layer parameters, and text-conditioned synthesis by pretrained language models.
MatLayerNet and MultiMat have released no public code, and MultiMat is not text-conditioned, so we compare against systems we can execute: the three diffusion baselines, direct Blender-Python generation under a shared backbone (Appendix~\ref{app:bpy}), and Material Apprentice, the only released text-to-procedural system (Appendix~\ref{app:ma}).
Material Apprentice reports that direct shader-graph generation fails in $30\%$ of cases and reaches a $100\%$ execution rate through best-of-$10$ compiler sampling at about $\$2$ and $30$ minutes per material with a GPT-5 backbone~\citep{gupta2026apprentice}.

\paragraph{Programs as visual representations, and standards.}
Graphics-program inference~\citep{ellis2018tikz}, ShapeAssembly~\citep{jones2020shapeassembly}, and Scene Language~\citep{zhang2024scenelanguage} connect learned generation to editable visual structure.
LAPS and LILO show how language can guide reusable program abstractions~\citep{wong2021laps,grand2024lilo}, while grammar prompting provides structured constraints for DSL generation~\citep{wang2023grammarprompting}.
Program repair~\citep{xia2023apr}, iterative critique~\citep{madaan2023selfrefine,gou2024critic}, and rendered feedback~\citep{wu2025chat2svg} are established components.
MDL separates declarative material definitions, including layering, from rendering algorithms~\citep{kettner2015mdl}; MaterialX supports portable material graphs~\citep{materialx}; and OpenPBR specifies a surface shading model~\citep{portsmouth2025openpbr}.
\method combines program repair, rendered critique, and execution feedback inside a deliberately smaller material vocabulary.
The goal is to measure the costs and benefits of that vocabulary, rather than to claim invention of portable material languages.
Table~\ref{tab:positioning} summarizes the task and representation boundaries.

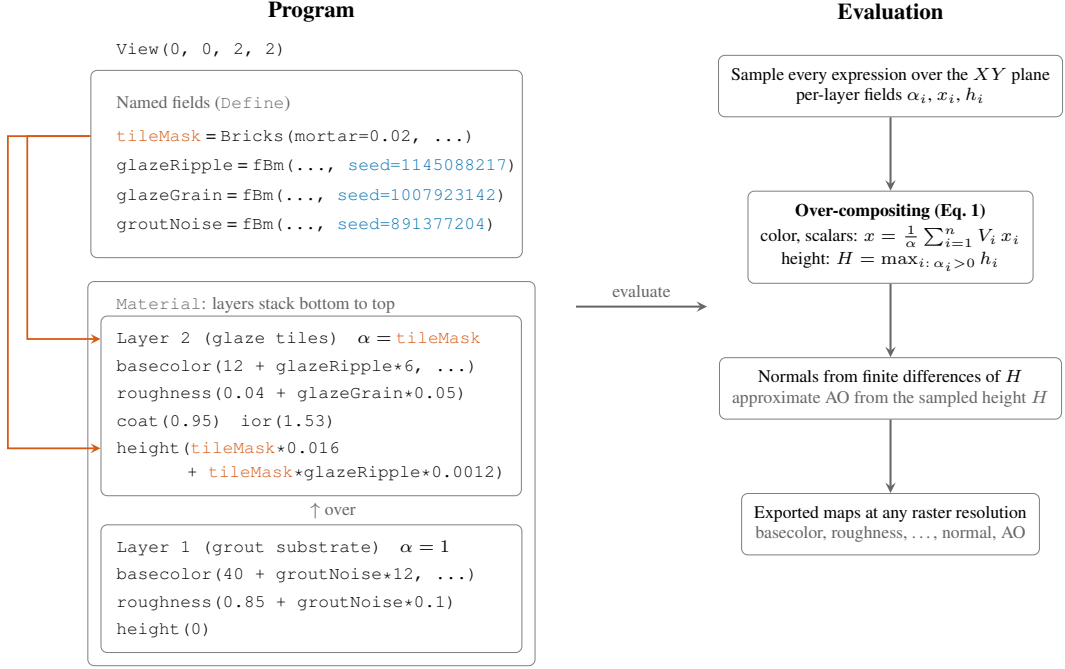
\begin{figure}[t]
    \centering
    \begingroup
\definecolor{mfshare}{RGB}{214,90,20}%
\definecolor{mfseed}{RGB}{0,114,178}%
\centering
\resizebox{\linewidth}{!}{%
    \begin{tikzpicture}[font=\scriptsize,
            row/.style={inner sep=1pt, align=left},
            fbox/.style={draw=black!50, rounded corners=2.5pt, inner sep=5pt},
            cbox/.style={draw=black!50, rounded corners=2.5pt, inner sep=5pt, align=center},
            flow/.style={draw=black!60, thick, -stealth},
            share/.style={draw=mfshare, line width=0.7pt, -stealth},
        ]
        \node[row,anchor=west] (view) at (0.85,-0.52) {\dsl{View(0, 0, 2, 2)}};
        \node[row,anchor=west] (d0) at (0.85,-1.28) {{\color{black!60}Named fields (\dsl{Define})}};
        \node[row,anchor=west] (d1) at (0.85,-1.73)
        {{\color{mfshare}\dsl{tileMask}} = \dsl{Bricks(mortar=0.02, ...)}};
        \node[row,anchor=west] (d2) at (0.85,-2.14)
        {\dsl{glazeRipple} = \dsl{fBm(..., {\color{mfseed}seed=1145088217})}};
        \node[row,anchor=west] (d3) at (0.85,-2.55)
        {\dsl{glazeGrain} = \dsl{fBm(..., {\color{mfseed}seed=1007923142})}};
        \node[row,anchor=west] (d4) at (0.85,-2.96)
        {\dsl{groutNoise} = \dsl{fBm(..., {\color{mfseed}seed=891377204})}};
        \node[fbox,fit=(d0)(d4),inner sep=9pt] (dbox) {};
        \node[row,anchor=west] (m0) at (0.85,-4.08) {{\color{black!60}\dsl{Material}: layers stack bottom to top}};
        \node[row,anchor=west] (l2a) at (0.85,-4.55)
        {\dsl{Layer 2 (glaze tiles)}\quad $\alpha=$ {\color{mfshare}\dsl{tileMask}}};
        \node[row,anchor=west] (l2b) at (0.85,-4.93)
        {\dsl{basecolor(12 + glazeRipple*6, ...)}};
        \node[row,anchor=west] (l2c) at (0.85,-5.31)
        {\dsl{roughness(0.04 + glazeGrain*0.05)}};
        \node[row,anchor=west] (l2d) at (0.85,-5.69)
        {\dsl{coat(0.95)}\quad\dsl{ior(1.53)}};
        \node[row,anchor=west] (l2e) at (0.85,-6.07)
        {\dsl{height(}{\color{mfshare}\dsl{tileMask}}\dsl{*0.016}};
        \node[row,anchor=west] (l2f) at (0.85,-6.41)
        {\dsl{\hspace*{0.98cm}+ }{\color{mfshare}\dsl{tileMask}}\dsl{*glazeRipple*0.0012)}};
        \node[fbox,fit=(l2a)(l2f),inner sep=5pt] (l2box) {};
        \node[row,anchor=west] (l1a) at (0.85,-7.45)
        {\dsl{Layer 1 (grout substrate)}\quad $\alpha=1$};
        \node[row,anchor=west] (l1b) at (0.85,-7.83)
        {\dsl{basecolor(40 + groutNoise*12, ...)}};
        \node[row,anchor=west] (l1c) at (0.85,-8.21)
        {\dsl{roughness(0.85 + groutNoise*0.1)}};
        \node[row,anchor=west] (l1d) at (0.85,-8.59)
        {\dsl{height(0)}};
        \coordinate (l1wide) at ([xshift=-5pt]l2box.east |- l1a);
        \node[fbox,fit=(l1a)(l1d)(l1wide),inner sep=5pt] (l1box) {};
        \node[fbox,fit=(m0)(l2box)(l1box),inner sep=5pt] (mbox) {};
        \node[font=\footnotesize\bfseries] (ptitle) at (mbox.center |- 0,0) {Program};
        \node[text=black!60] at (3.9,-6.93) {$\uparrow$ over};
        \draw[share] (dbox.west |- d1) -- (-0.35,-1.73) -- (-0.35,-4.55) -- (l2box.west |- l2a);
        \draw[share] (dbox.west |- d1) -- (-0.62,-1.73) -- (-0.62,-6.07) -- (l2box.west |- l2e);
        \node[font=\footnotesize\bfseries] (etitle) at (11.65,0) {Evaluation};
        \node[cbox, anchor=north] (eA) at (11.65,-0.60)
        {Sample every expression over the $XY$ plane\\
        {per-layer fields $\alpha_i$, $x_i$, $h_i$}};
        \node[cbox, anchor=north] (eB) at ([yshift=-1.02cm]eA.south -| 11.65,0)
        {\textbf{Over-compositing (Eq.~\ref{eq:slate-over})}\\[2pt]
            color, scalars: $x=\frac{1}{\alpha}\sum_{i=1}^{n}V_i\,x_i$\\[1pt]
            height: $H=\max_{i:\,\alpha_i>0}h_i$};
        \node[cbox, anchor=north] (eC) at ([yshift=-1.02cm]eB.south -| 11.65,0)
        {Normals from finite differences of $H$\\{\color{black!60}approximate AO from the sampled height $H$}};
        \node[cbox, anchor=north] (eD) at ([yshift=-1.02cm]eC.south -| 11.65,0)
        {Exported maps at any raster resolution\\
        {\color{black!60}basecolor, roughness, \ldots, normal, AO}};
        \draw[flow] (eA.south) -- (eB.north);
        \draw[flow] (eB.south) -- (eC.north);
        \draw[flow] (eC.south) -- (eD.north);
        \draw[flow, line width=0.8pt] ([xshift=16pt]mbox.east |- 0,-4.10) -- ([xshift=-16pt]eB.west |- 0,-4.10)
        node[midway, above, black!60] {evaluate};
    \end{tikzpicture}%
}%
\endgroup
    \vspace{-2em}
    \caption{Overview of the \method representation using the tile program of Listing~\ref{lst:example}. A program defines an optional \dsl{View}, reusable fields, and a bottom-to-top \dsl{Material} stack. Shared fields couple coverage with color, roughness, or height, explicit noise seeds pin a realization, scalar and color channels use the over rule of Eq.~\ref{eq:slate-over}, and height uses a separate maximum rule before normals and approximate ambient occlusion are derived. The shared \dsl{tileMask} (orange) drives both the glaze layer's coverage and its height, and explicit noise \dsl{seed} values (blue) pin one stochastic realization.}
    \label{fig:representation-overview}
\end{figure}

\section{The \method Representation}
\label{sec:dsl}

The representation separates editable source from sampled material maps (Figure~\ref{fig:representation-overview}).
A program defines spatial fields, assigns them to layer coverage and PBR channels, and evaluates them at a chosen resolution.
Layers group coverage with surface properties, named fields expose cross-channel dependencies, and explicit seeds distinguish a design from one stochastic realization.

\paragraph{Why this representation?}
A material concept such as grout or glaze can affect several properties in one region.
Layers keep those properties together, while a shared field lets one motif drive multiple channels.
These are inspectable source dependencies, and the matched controls in Appendices~\ref{app:bpy} and~\ref{app:editing} measure their effect. \looseness=-1

Changing explicit seeds explores realizations of a fixed program, but seeded noise can affect coverage, color, and height as well as fine texture.

\subsection{Program Structure}
\label{sec:dsl-structure}

A program has an optional sampling window (\dsl{View}), named expressions (\dsl{Define}), and bottom-to-top layers (\dsl{Material}).
Expressions are fields over the $XY$ plane, with $+x$ right and $+y$ up.
Changing raster size re-samples the same source fields, although normals, approximate ambient occlusion, and exported pixels depend on resolution.
Listing~\ref{lst:example} gives a complete two-layer tile program.

\paragraph{Layer channels and shared fields.}
\label{sec:dsl-channels}
\label{sec:dsl-expr}
Each layer assigns coverage $\alpha\in[0,1]$, base color, roughness, metallicity, emission, other surface-response channels, and height $h$.
Values can be constants or expressions built from transforms, noise, periodic patterns, and shapes.
Named definitions form an acyclic graph and can be reused, as when the tile mask drives both coverage and height.
Exported channels map to renderer inputs~\citep{burley2012principled,portsmouth2025openpbr}; this is an authoring convention rather than a physically derived coating model.
Channel ranges and validation are specified in Appendix~\ref{app:rep-channels}.

\subsection{Compositing Semantics}
\label{sec:dsl-compositing}

At each position, let layer $1$ be the bottom layer and define the visible coverage of layer $i$ as $V_i=\alpha_i\prod_{j=i+1}^{n}(1-\alpha_j)$ and total coverage as $\alpha=1-\prod_{i=1}^{n}(1-\alpha_i)$. For finite layer values, a scalar surface channel $x$ resolves as
\begin{equation}
    x=\begin{cases}
        \displaystyle\frac{1}{\alpha}\sum_{i=1}^{n}V_i x_i, & \alpha>0, \\
        0,                                                  & \alpha=0.
    \end{cases}
    \label{eq:slate-over}
\end{equation}
Since $V_i\geq0$ and $\sum_i V_i=\alpha$, covered pixels receive a convex blend of layer values.
Base color blends in linear light; this rule describes authoring visibility rather than radiative transfer.

Height instead takes the maximum finite height among positively covered layers, or zero if none contributes.
Fractional alpha therefore blends surface channels without attenuating relief; edge tapering must be encoded in the height expression.

\subsection{Dependencies and Controlled Edits}
\label{sec:dsl-dependencies}

For fixed execution settings, changing only parameters outside a channel's transitive source dependencies preserves its exported map.
Its rendered appearance can still change through another channel's effect on shading.
Appendix~\ref{app:rep-dependencies} gives the formal dependency statement.

In the tile program, the upper layer uses \dsl{tileMask} for coverage and height, so changing mortar width changes grout exposure and tile-edge relief together.
The one-case edits in Section~\ref{sec:manual-edits} test such dependencies; they do not establish a general editing advantage.

Inlining named definitions preserves evaluation semantics on the tested programs, enabling a future controlled comparison of source reuse (Appendix~\ref{app:rep-serialization}).

\subsection{Execution and Validation}
\label{sec:dsl-grammar}
\label{sec:dsl-pitfalls}
\label{sec:dsl-engine}
\label{sec:engine-render}

The Python engine exports sampled maps, and a TypeScript port supports browser inspection.
Parser checks and finite regression tests do not prove numerical validity or successful export for every accepted program (Appendices~\ref{app:grammar} and~\ref{app:engine}).
Evaluation uses Blender in a head-on \textbf{flat} layout and a \textbf{staged} layout that exposes relief and transmission.

\section{Text-to-Material Generation}
\label{sec:method}

\begin{figure}[H]
    \centering
    \vspace{-1.5em}
    \input{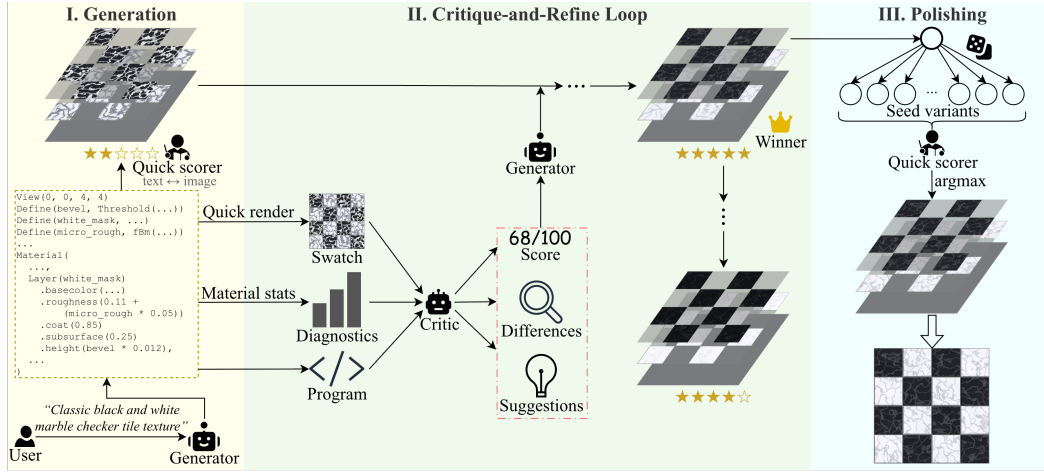}
    \vspace{-1em}
    \caption{\textbf{A visual walkthrough of material authoring.}
        A text request becomes a layered program, whose preview, channel statistics, and source inform critique and revision.
        Stage III depicts one candidate's seed sweep.
        The full search applies this sweep to the distinct initial, selected, and final programs (Section~\ref{sec:method-polish}).
        Each sweep includes its original program and changes only noise seeds.
        The program excerpt, intermediate appearances, stars, and $68/100$ score are schematic illustrations of information flow, not a measured trajectory or evidence of monotonic improvement.}
    \label{fig:overview}
    \vspace{-1.8em}
\end{figure}

The pipeline generates and repairs a program, revises it using preview feedback, then selects among program designs and their seed variants (Figure~\ref{fig:overview}).
Appendix~\ref{app:method-mechanics} gives implementation details.

\subsection{Stage I: Generation}
\label{sec:method-prompt}
\label{sec:method-retry}

The language model receives a DSL reference, six examples, and an organic-texture playbook (Appendix~\ref{app:prompts}).
The parser checks its program's syntax, references, and constructors, returning errors for up to three corrective responses.

Accepted programs are canonicalized, making noise seeds explicit before critique and selection.
All $2538$ retained main-run conversations completed without aborting and produced renderable final programs under the recorded settings.

\subsection{Stage II: Critique and Refinement}
\label{sec:method-quickrender}
\label{sec:method-stats}
\label{sec:method-critique}

The critic can receive a quick render, $28$ unlit material statistics, and source code with named fields and parameters.
The preview shades composited maps and height-derived normals under a fixed directional light.

The critic normally shares the generator's backbone and sampling settings, omitting image input for text-only backbones.
Each fresh critique returns a model-assessed match score, visual differences, and revision suggestions; the generator is asked to preserve unchanged layers and seeds.
Five revisions produce $P_0,\ldots,P_5$, with one critique per trajectory program and no revision after the last critique.

\subsection{Stage III: Selection and Seed Search}
\label{sec:method-polish}

A late revision can regress, so the last round is not chosen automatically.
Let $q_r(P,t)$ be the MobileCLIP2~\citep{mobileclip2} score of an $r\times r$ quick render for text $t$.
The selector picks the best of $P_0,\ldots,P_5$ using $q_{512}$, breaking ties toward the earliest round.
Seed search uses $q_{256}$ on the distinct initial, selected, and last programs, indexed by $C=\{0,j,5\}$.

For each $d\in C$, let $\mathcal{S}_K(P_d)$ contain the original program and $K$ variants obtained by changing only noise seeds. The returned program is $P^\star\in\arg\max_{P\in\bigcup_{d\in C}\mathcal{S}_K(P_d)}q_{256}(P,t)$.
With $K=1000$, the search scores at most $3(K+1)$ candidates.
Repeated seed references change together, while all non-seed source remains fixed within each candidate sweep.
The winner can come from any of the three trajectory positions.

Including each original prevents a decrease in $q_{256}$ relative to that candidate pool, but it does not guarantee better external metrics or human preference.
Appendices~\ref{app:refine} and~\ref{app:cma} analyze search budgets and proxy mismatch.

\section{Experiments}
\label{sec:eval}

Our evaluation asks two questions: whether the complete authoring system produces renders that match material descriptions, and whether its recorded programs expose useful post-generation control.
We therefore separate prompt-alignment results, which the benchmark, the trajectory analysis, and the blind study measure on rendered appearance, from evidence about the representation itself, which controlled generation, editing, budget, and channel comparisons now measure. \looseness=-1
Section~\ref{sec:results-main} answers the first question, and Sections~\ref{sec:results-stages} through~\ref{sec:results-threats} cover budget and revision, editing, the blind study, and the limitations.
All means average the three recorded runs for each prompt and then average over $141$ prompts, unless stated otherwise.

\subsection{Setup}
\label{sec:eval-setup}
\label{sec:scope-eval}
\label{sec:scope-ceiling}

The benchmark contains $141$ prompts from four public sources: category prompts from \citet{hu2023generating} ($30$), MatSynth descriptions~\citep{vecchio2024matsynth} ($11$), StableMaterials prompts~\citep{vecchio2024stablematerials} ($50$), and text2fabric descriptions~\citep{deschaintre2023visual} ($50$).
Sources with more than $50$ descriptions are reduced by farthest-point sampling in sentence-embedding space, with the selected prompt list fixed across methods.
This curated benchmark includes figurative and specific motifs, so its average characterizes this prompt distribution rather than material requests in general.

We compare against three released text-to-material-map systems: \textbf{MatFuse}~\citep{vecchio2024matfuse}, \textbf{StableMaterials}~\citep{vecchio2024stablematerials}, and \textbf{IntrinsiX}~\citep{kocsis2025intrinsix}.
All systems produce $512{\times}512$ maps rendered in the same Blender scenes and in both flat and staged layouts.
The layouts are useful but not representation-isolated: \method's height drives geometric displacement, StableMaterials' height drives bump shading, MatFuse and IntrinsiX emit no height map, and only \method exposes transmission.
We therefore interpret the main comparison as complete-system appearance under the recorded routes (Appendix~\ref{app:baselines}), with a shared-channel control in Appendix~\ref{app:e2}.
Rerendered over the common channel intersection, the flagship still leads every metric-layout cell, with the per-channel cost quantified in Section~\ref{sec:results-threats}.

Prompt alignment is scored with BLIPScore~\citep{li2023blip2}, CLIPScore~\citep{hessel2021clipscore}, VQAScore~\citep{lin2024vqascore}, and an MLLM judge (\texttt{claude-sonnet-5}, temperature zero), each scaled to $[0,100]$ with higher better.
These are learned visual proxies, not measures of physical accuracy or edit utility.
The six \method backbones each use three recorded authoring runs per prompt with procedural-noise seeds $42$, $123$, and $2026$.
All \method main runs fix generation temperature at $0.7$ and use reasoning effort high where the provider exposes it, with Gemma and Qwen inheriting the provider default.
Each main run uses an initial generation, 5 critique--revision rounds, trajectory selection, and up to $1000$ noise-seed variants for each of the first, selected, and last programs, at most $3{,}003$ scored candidates per run at about $70$ ms each.
This unmatched budget is quantified in Appendix~\ref{app:budget} against best-of-$N$ selection for the baselines.
For paired inference we use per-prompt run means, two-sided Wilcoxon signed-rank tests, and $10{,}000$-resample percentile bootstrap confidence intervals, with Holm correction over the $24$ flagship-versus-baseline contrasts.
Because the flagship was selected after comparing all six backbones, these tests remain exploratory.

\subsection{Prompt Alignment Across Methods}
\label{sec:results-main}

The strongest configuration, \texttt{gemini-3.6-flash}, has the highest mean on all four alignment metrics in both layouts (Table~\ref{tab:main}).
Against StableMaterials in the flat layout, its mean differences are BLIPScore $+27.5$ ($95\%$ CI $[21.8,33.3]$), CLIPScore $+3.1$, VQAScore $+9.3$, and judge $+9.6$, with all four Holm-adjusted $p$-values below $0.001$.
Every judge delta in this paper is computed within a single scoring session, because judge scores recorded in different sessions are not level-comparable (Appendix~\ref{app:e2}).
Across the three baselines, two layouts, and four metrics, $23$ of $24$ corrected contrasts remain significant.
The exception is the staged judge comparison with StableMaterials: $+3.0$, $95\%$ CI $[-1.2,7.3]$, $p{=}0.21$.
This interval supports neither superiority nor equivalence.

Backbone choice and prompt source both matter.
All six program-generating configurations exceed the strongest diffusion baseline on flat BLIPScore, but only the flagship exceeds StableMaterials on mean flat judge score; the other 5 range from $45.8$ to $56.8$ against $57.4$.
Per-source means also qualify the aggregate result: the flagship's flat judge score trails StableMaterials on the StableMaterials-source and category prompts, while its overall judge advantage comes from MatSynth and text2fabric (Appendix~\ref{app:persource}).
The benchmark thus supports competitive prompt alignment for the full authoring system, tested directly against same-loop Blender-Python in Appendix~\ref{app:bpy}. \looseness=-1

\begin{table}[tb]
    \caption{Main comparison on the $141$-prompt benchmark under four alignment metrics (mean over three recorded runs, higher is better). Flat and Staged give the two rendering layouts. Our rows use the full three-stage pipeline with the named generator and the same model as critic, with text-only critique for DeepSeek and GLM. Bold marks the best value per metric within each layout block. Baselines sample three seeds per prompt. \method rows use three recorded authoring runs with procedural seeds $42$, $123$, and $2026$. Means average the runs within each prompt before averaging prompts.}
    \label{tab:main}
    \centering
    \footnotesize
    \setlength{\tabcolsep}{3.2pt}
    \resizebox{\linewidth}{!}{%
        \begin{tabular}{l cccc cccc}
            \toprule
                                                               & \multicolumn{4}{c}{Flat} & \multicolumn{4}{c}{Staged}                                                                                                       \\
            \cmidrule(lr){2-5}\cmidrule(lr){6-9}
            Method                                             & BLIPScore                & CLIPScore                  & VQAScore       & Judge          & BLIPScore      & CLIPScore      & VQAScore       & Judge          \\
            \midrule
            IntrinsiX~\citep{kocsis2025intrinsix}              & 29.94                    & 24.29                      & 43.61          & 56.10          & 20.96          & 21.80          & 41.97          & 49.60          \\
            MatFuse~\citep{vecchio2024matfuse}                 & 8.27                     & 20.12                      & 30.61          & 36.36          & 8.58           & 19.25          & 30.56          & 33.63          \\
            StableMaterials~\citep{vecchio2024stablematerials} & 28.57                    & 25.66                      & 45.36          & 57.41          & 21.33          & 23.59          & 45.53          & 54.29          \\
            \midrule
            \method{} (gemma-4-26b-a4b-it)                     & 33.56                    & 25.46                      & 47.12          & 47.48          & 22.07          & 22.48          & 47.43          & 40.17          \\
            \method{} (qwen3.6-35b-a3b)                        & 35.64                    & 25.70                      & 46.36          & 45.80          & 21.93          & 22.55          & 46.60          & 37.47          \\
            \method{} (deepseek-v4-flash-0731)                 & 46.42                    & 27.29                      & 51.06          & 53.38          & 30.49          & 23.93          & 51.04          & 45.11          \\
            \method{} (glm-5.2)                                & 42.44                    & 26.78                      & 50.36          & 53.88          & 27.85          & 23.78          & 50.50          & 46.28          \\
            \method{} (gpt-5.6-luna)                           & 48.78                    & 27.61                      & 49.56          & 56.80          & 31.55          & 24.51          & 49.75          & 47.02          \\
            \method{} (gemini-3.6-flash)                       & \textbf{56.06}           & \textbf{28.80}             & \textbf{54.71} & \textbf{67.01} & \textbf{36.14} & \textbf{25.30} & \textbf{54.26} & \textbf{57.34} \\
            \bottomrule
        \end{tabular}}
\end{table}

\begin{figure}[tb]
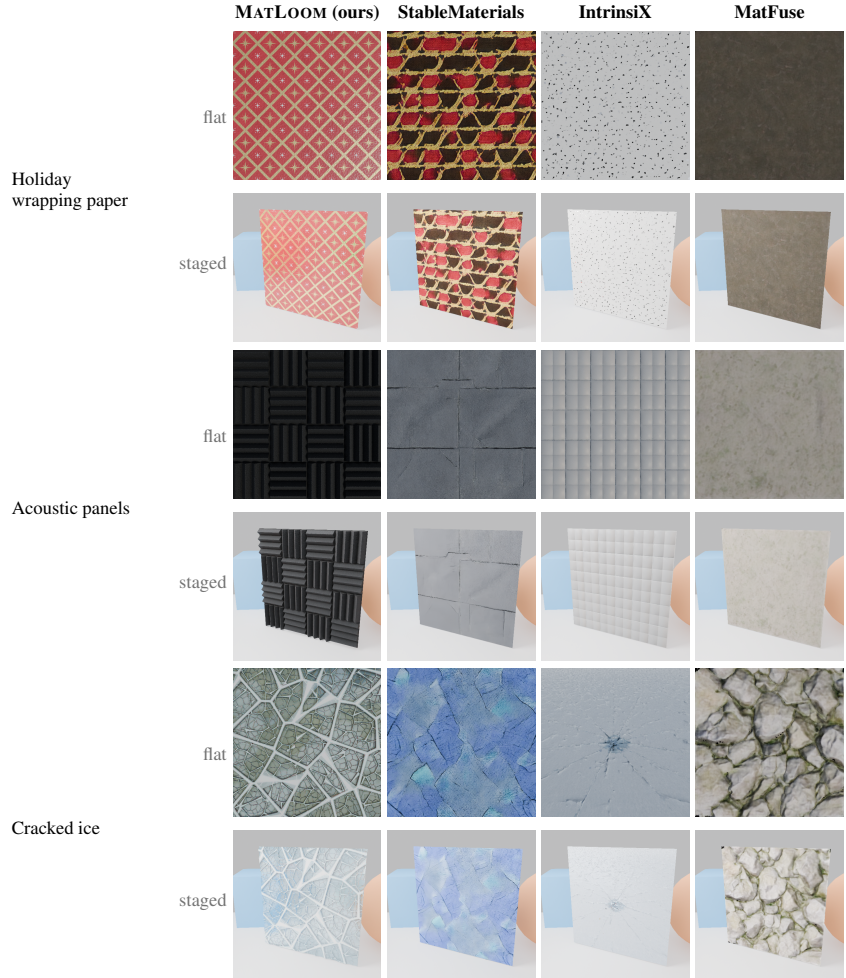

    \centering
    \setlength{\cmpcellW}{0.140\linewidth}%
    \setlength{\tabcolsep}{0.5pt}%
    \begin{tabular}{@{}c@{\hspace{0.006\linewidth}}c@{\hspace{0.006\linewidth}}c@{\hspace{0.006\linewidth}}c@{\hspace{0.006\linewidth}}c@{\hspace{0.006\linewidth}}c@{}}
                                                                                                                                           &          & \textbf{\scriptsize \method (ours)} & \textbf{\scriptsize StableMaterials} & \textbf{\scriptsize IntrinsiX} & \textbf{\scriptsize MatFuse} \\[2pt]
        \cmprowlabel{Holiday wrapping paper}                                                                                               & \cmptags &
        \cmpcell{comparison/holiday-wrapping-paper-ours-flat.png}{comparison/holiday-wrapping-paper-ours-staged.png}                       &
        \cmpcell{comparison/holiday-wrapping-paper-stablematerials-flat.png}{comparison/holiday-wrapping-paper-stablematerials-staged.png} &
        \cmpcell{comparison/holiday-wrapping-paper-intrinsix-flat.png}{comparison/holiday-wrapping-paper-intrinsix-staged.png}             &
        \cmpcell{comparison/holiday-wrapping-paper-matfuse-flat.png}{comparison/holiday-wrapping-paper-matfuse-staged.png}                                                                                                                                                                         \\[2pt]
        \cmprowlabel{Acoustic panels}                                                                                                      & \cmptags &
        \cmpcell{comparison/acoustic-panels-ours-flat.png}{comparison/acoustic-panels-ours-staged.png}                                     &
        \cmpcell{comparison/acoustic-panels-stablematerials-flat.png}{comparison/acoustic-panels-stablematerials-staged.png}               &
        \cmpcell{comparison/acoustic-panels-intrinsix-flat.png}{comparison/acoustic-panels-intrinsix-staged.png}                           &
        \cmpcell{comparison/acoustic-panels-matfuse-flat.png}{comparison/acoustic-panels-matfuse-staged.png}                                                                                                                                                                                       \\[2pt]
        \cmprowlabel{Cracked ice}                                                                                                          & \cmptags &
        \cmpcell{comparison/cracked-ice-ours-flat.png}{comparison/cracked-ice-ours-staged.png}                                             &
        \cmpcell{comparison/cracked-ice-stablematerials-flat.png}{comparison/cracked-ice-stablematerials-staged.png}                       &
        \cmpcell{comparison/cracked-ice-intrinsix-flat.png}{comparison/cracked-ice-intrinsix-staged.png}                                   &
        \cmpcell{comparison/cracked-ice-matfuse-flat.png}{comparison/cracked-ice-matfuse-staged.png}
    \end{tabular}
    \caption{Qualitative comparison. Each cell stacks the flat layout over the staged layout. The examples were selected to expose structure and channel behavior, not to estimate typical performance or isolate the representation.}
    \label{fig:comparison}
\end{figure}

Figure~\ref{fig:comparison} shows selected examples where explicit spatial structure is visible.
For holiday wrapping paper, \method produces a repeated diamond-and-star design; for acoustic panels, alternating wedge orientations form a checkerboard with visible relief; for cracked ice, a nested crack network appears over a transmissive sheet.
The staged views reflect the unequal routing noted above.

\subsection{Revision Trajectory and Selection Behavior}
\label{sec:results-stages}

The flagship's stored trajectory gives a within-run readout of the authoring loop (Table~\ref{tab:stages}).
Round $0$ already reaches $48.31$ BLIPScore and $61.35$ judge in the flat layout, above all three diffusion baselines on those means, significantly so on every non-judge cell (Table~\ref{tab:budget0}).
Five critique--revision rounds add $+3.5$ BLIPScore and $+6.3$ judge, while trajectory selection adds $+2.3$ BLIPScore over the last revision.
The final polished output reaches $56.06$ BLIPScore, but its flat judge score is slightly below round $5$.
The selector beats the last revision on flat BLIPScore in $178$ of $423$ runs ($42\%$), loses in $137$, and ties in $108$; it also reduces mean flat judge by about one point.
The trajectory therefore improves some alignment proxies while revealing disagreement between the quick scorer and other judgments (Appendix~\ref{app:refine}).

\paragraph{Budget-matched controls.}
Round $0$ (one LLM call, one scored candidate) isolates the representation from the authoring budget, and there the flagship leads StableMaterials and IntrinsiX on six of eight metric-layout cells (two-sided Wilcoxon) and MatFuse on all eight (Table~\ref{tab:budget0}).
The two exceptions are the judge cells against StableMaterials and IntrinsiX, which remain statistical ties (flat $+3.9$, $p{=}0.13$ and $+5.2$, $p{=}0.055$, staged $-1.3$, $p{=}0.55$ and $+3.4$, $p{=}0.19$), so the flat judge margin over StableMaterials needs critique and search.
The full pipeline spends $12$ LLM calls and at most $3{,}003$ scored candidates per prompt, with $75\%$ of its $44.7$ hour total in seed search, against StableMaterials' $0.5$ seconds per sample.
A final-metric replay shows one variant per candidate already reaches $99\%$ of the sweep's flat BLIPScore ($55.47$ at $K{=}1$ versus $56.06$), a marginal final-metric yield of $0.6$ points (Appendix~\ref{app:budget}).
Granting each baseline best-of-$3$ selection over its three seeds with our own MobileCLIP2 quick scorer, the flagship still leads StableMaterials on seven of eight cells, the staged judge ($-1.1$) excepted, and it leads MatFuse on all eight cells at best-of-$9$ (Appendix~\ref{app:budget}).
At StableMaterials best-of-$10$, the strongest feasible selection we test, only staged CLIPScore remains a tie ($-0.2$) while both judge cells become significant leads ($+15.6$ flat and $+12.2$ staged, both $p{<}0.001$, paired within a single scoring session, Appendix~\ref{app:budget}).
A per-metric oracle over the three samples, an infeasible upper bound that requires the test labels, would flip the two judge cells against StableMaterials ($-3.5$ flat, $-9.2$ staged) while the BLIPScore, CLIPScore, and VQAScore leads persist ($+14.7$, $+1.2$, $+2.7$), as Appendix~\ref{app:budget} and Figure~\ref{fig:budget-curve} disclose.
Over the same ten samples the oracle narrows the judge cells to $+6.1$ flat and a $+1.2$ staged tie while several staged proxy cells move to StableMaterials (Appendix~\ref{app:budget}). \looseness=-1

\begin{table}[htbp]
    \caption{Budget-matched comparison at round $0$: the \texttt{gemini-3.6-flash} flagship's initial program (one LLM call, one scored candidate, no critique, no seed search), mean over its three recorded runs, against each baseline's three-seed mean ($n{=}141$ paired prompts). Cells are paired mean differences, ours minus baseline ($^{*}p{<}0.05$, $^{**}p{<}0.01$, $^{***}p{<}0.001$, two-sided Wilcoxon), and the four cells without stars carry bracketed $95\%$ bootstrap intervals.}
    \label{tab:budget0}
    \centering
    \footnotesize
    \renewcommand{\arraystretch}{0.92}
    \setlength{\tabcolsep}{2.5pt}
    \resizebox{\linewidth}{!}{%
        \begin{tabular}{l cccc cccc}
            \toprule
             & \multicolumn{4}{c}{Flat} & \multicolumn{4}{c}{Staged} \\
            \cmidrule(lr){2-5}\cmidrule(lr){6-9}
            Baseline & BLIP & CLIP & VQA & Judge & BLIP & CLIP & VQA & Judge \\
            \midrule
            IntrinsiX & $+$18.4$^{***}$ & $+$3.4$^{***}$ & $+$9.7$^{***}$ & $+$5.2 {[}$+$0,$+$10{]} & $+$10.0$^{***}$ & $+$2.6$^{***}$ & $+$10.7$^{***}$ & $+$3.4 {[}$-$1,$+$8{]} \\
            MatFuse & $+$40.0$^{***}$ & $+$7.6$^{***}$ & $+$22.7$^{***}$ & $+$25.0$^{***}$ & $+$22.4$^{***}$ & $+$5.1$^{***}$ & $+$22.1$^{***}$ & $+$19.4$^{***}$ \\
            StableMaterials & $+$19.7$^{***}$ & $+$2.0$^{***}$ & $+$8.0$^{***}$ & $+$3.9 {[}$-$1,$+$9{]} & $+$9.6$^{***}$ & $+$0.8$^{*}$ & $+$7.1$^{***}$ & $-$1.3 {[}$-$6,$+$3{]} \\
            \bottomrule
        \end{tabular}}
\end{table}

\paragraph{Component diagnostics.}
\label{sec:results-ablation}
The stored component comparisons are exploratory: their full reference fixes temperature at $0.7$ while variant runs inherit the provider default, so they do not isolate component effects.
A settings-matched rerun (six configurations, $40$ prompts, three seeds, temperature $0.7$ on both sides) now provides that evidence: the load-bearing components are the prompt playbook ($-4.8$ flat BLIPScore) and the render diagnostics panel ($-5.8$ flat, $-7.3$ staged with $p{<}0.001$ after Holm correction), while the render-critique effect shrinks to $-3.1$ (not significant) and the code-critique anomaly fades to $+1.5$ (not significant).
The exploratory records are retained, with the settings mismatch stated alongside them (Appendix~\ref{app:ablation}).

\begin{table}[htbp]
    \caption{Stage-wise scores for \texttt{gemini-3.6-flash}, averaged over $423$ runs. Round $0$ precedes critique, Round $5$ is the last revision, Selected maximizes the MobileCLIP2 quick scorer $q_{512}$ (Section~\ref{sec:method-polish}) over the trajectory, and Final additionally searches seeds of the first, selected, and last programs. Best per metric in \textbf{bold}.}
    \label{tab:stages}
    \centering
    \footnotesize
    \renewcommand{\arraystretch}{0.92}
    \setlength{\tabcolsep}{4pt}
    \begin{tabular}{l cccc cccc}
        \toprule
                     & \multicolumn{4}{c}{Flat} & \multicolumn{4}{c}{Staged}                                                                                                       \\
        \cmidrule(lr){2-5}\cmidrule(lr){6-9}
        Stage        & BLIPScore                & CLIPScore                  & VQAScore       & Judge          & BLIPScore      & CLIPScore      & VQAScore       & Judge          \\
        \midrule
        Round 0      & 48.31                    & 27.70                      & 53.34          & 61.35          & 30.95          & 24.37          & 52.67          & 52.98          \\
        Round 5      & 51.86                    & 28.25                      & 52.93          & \textbf{67.61} & 33.23          & 24.91          & 53.43          & 56.82          \\
        Selected     & 54.17                    & 28.71                      & 53.97          & 66.56          & 35.50          & 25.25          & 54.07          & 57.20          \\
        Final (full) & \textbf{56.06}           & \textbf{28.80}             & \textbf{54.71} & 67.01          & \textbf{36.14} & \textbf{25.30} & \textbf{54.26} & \textbf{57.34} \\
        \bottomrule
    \end{tabular}
\end{table}

\subsection{Program Control in One Material}
\label{sec:manual-edits}

Figure~\ref{fig:explicit-program-edits} (Appendix~\ref{app:example}) illustrates how an authored program can be edited after generation. \looseness=-1
Starting from the final program for a two-layer tile material, we manually increase the tile roughness offset from $0.04$ to $0.75$, widen grout by changing the shared mortar parameter from $0.02$ to $0.06$, and shift tile base-color offsets to a blue glaze.
All other source bytes, including noise seeds and the sampling window, are preserved. \looseness=-1
At both $256{\times}256$ and $1024{\times}1024$, repeated evaluation in the recorded environment produces byte-identical previews; roughness and color edits change only their intended composited channels, while the grout edit deliberately propagates through channels that share the named mask (Appendix~\ref{app:example}).
Appendix~\ref{app:editing} extends this single case to a frozen $172$-task editing benchmark that also runs both raster baselines through re-prompting, where joint success is $84.3\%$ for the DSL arm against $9.9\%$ and $4.1\%$.

\subsection{Human Evaluation}
\label{sec:user-study}

We ran a blind online preference study over $20$ prompts, five from each benchmark source, drawn from the $141$-prompt benchmark with a seeded diversity-gated selection (Appendix~\ref{app:userstudy}).
Each trial showed four unlabeled $512{\times}512$ flat-layout renders in a $2{\times}2$ grid, \method with the flagship backbone against the StableMaterials, IntrinsiX, and MatFuse outputs, all at procedural seed $123$, and all thirty volunteers completed all trials, passed the attention check, and none was excluded. \looseness=-1

\method won $59.2\%$ of the $600$ forced choices ($95\%$ CI $[55.8,62.7]$ by participant-cluster bootstrap), ahead of StableMaterials at $19.3\%$, IntrinsiX at $15.3\%$, and MatFuse at $6.2\%$.
Every participant chose \method more often than any one baseline across the $20$ trials, with $8$--$17$ \method choices per participant.
Mean fidelity ratings were $4.98$ for \method, $3.60$ for StableMaterials, $3.01$ for IntrinsiX, and $2.27$ for MatFuse.
The mean within-trial rating advantage over StableMaterials was $1.37$ points on the $1$--$7$ scale ($95\%$ participant-cluster bootstrap CI $[1.15,1.61]$), positive under Wilcoxon signed-rank tests both by prompt ($n{=}20$, $p{=}4.4\times 10^{-4}$, rank-biserial $0.83$) and by participant mean ($n{=}30$, $p{=}6.5\times 10^{-7}$).
The preference held descriptively without ($60.0\%$ \method win rate) and with ($58.8\%$) image-creation experience. \looseness=-1
The study measures prompt-conditioned appearance preference on this subset, not editing utility or physical accuracy. A second blind study with the same instrument covers the new arms: the DSL arm wins $65.5\%$ of choices against both Blender-Python arms and $83.3\%$ against Material Apprentice ($n{=}21$; Appendix~\ref{app:userstudy}).

\subsection{Limitations and Threats to Validity}
\label{sec:results-threats}
\label{sec:results-limitations}

The fixed primitive and channel vocabulary constrains fine microstructure, specific figurative motifs, participating media, and arbitrary reflectance models.
The benchmark measures prompt alignment, not physical material accuracy, edit success, appearance consistency across resolutions, or causal value of the representation.
Appendix~\ref{app:editing} measures edit success over multiple materials and edit families. \looseness=-1
Holm correction covers the $24$ flagship-versus-baseline contrasts, but it does not account for selecting the flagship from six backbones on this benchmark, so these tests remain exploratory.
The authoring pipeline uses a substantial and unmatched seed-search budget, and the diffusion baselines do not receive an equalized search or channel-routing treatment.
Appendix~\ref{app:budget} replays this budget on the final metrics and adds best-of-$N$ selection for the baselines, showing alignment-proxy advantages present at one-sample budget while the flat judge margin over StableMaterials is partly inference expenditure (Section~\ref{sec:results-stages}).
The staged layout combines generation with renderer differences, including displacement, bump, height availability, and transmission, and the shared-channel rerender of Appendix~\ref{app:e2} partially separates them: dropping to the strict channel intersection costs \method about $6$ flat and $3$ staged judge points against at most $2.1$ for each baseline.
Component comparisons are exploratory because variant runs use one run per prompt, with the differing settings stated alongside the values, and they guide follow-up experiments rather than proving component necessity (Appendix~\ref{app:ablation}).
Directly testing the representation hypothesis requires controlled comparisons against other executable material forms under matched backbones, budgets, and rendering channels.
Appendix~\ref{app:bpy} adds Blender-Python under the shared backbone and Appendix~\ref{app:ma} a native Material Apprentice run under its own GPT-5 backbone, the Blender-Python control scoped to \texttt{gemini-3.6-flash}, seed $42$, $45$ prompts. \looseness=-1

\section{Conclusion}
\label{sec:conclusion}

\method studies text-to-material generation as executable authoring: compact layered programs retain the construction of each rendered asset.
On the $141$-prompt benchmark, its strongest configuration exceeds three diffusion baselines on all four flat-layout alignment proxies, and the margin survives feasible equal-budget selection at StableMaterials best-of-$10$ on seven of eight cells, staged CLIPScore excepted. A per-metric oracle over the same ten samples (requiring test labels) narrows the judge margins to $6.1$ points flat and a staged statistical tie, with $59.2\%$ of choices in a $30$-person blind study. \looseness=-1
Revision and seed search improve different proxies. The matched controls locate where the representation carries the advantage.

\subsection*{AI use statement}
Generative AI is part of the experimental method: hosted language models generate and critique material programs, and a separate multimodal model supplies one evaluation score.
AI tools also assisted with code development, literature search and synthesis, and manuscript drafting and editing.
The authors take responsibility for the final content, including the cited literature, mathematical statements, reported measurements, participant-study data, and released code.

\subsection*{Ethics statement}
This work studies the generation of digital surface materials from text.
The release is limited to source code and does not redistribute benchmark prompts, baseline models, or third-party data or model weights, which remain subject to their original licenses.
An authoring system may reproduce protected surface designs or generate assets that appear plausible but have unverified physical properties, so rendered appearance should not be treated as a measured material specification.
Both preference studies were voluntary and began with informed consent.
Participants could optionally provide an email address for follow-up, so responses were confidential rather than anonymous.
The reported analysis excluded the email field and used de-identified responses.

\subsection*{Reproducibility statement}
The representation and compositing rules are specified in Section~\ref{sec:dsl}, with an example program and grammar summary in Appendices~\ref{app:example} and~\ref{app:grammar}.
We will release the source code, including the engine with its parity fixture suite and the scoring pipeline.
Pinned program seeds support repeated evaluation without another language-model call, provided the engine, export settings, and rendering environment are fixed.
Cross-engine agreement is tested on a finite fixture set with explicit tolerances and exclusions (Appendix~\ref{app:engine}).
The quantitative tables and paired comparisons are computed from the stored per-run scores and the exported study responses (Appendix~\ref{app:provenance}).

\bibliographystyle{iclr2027_conference}
\bibliography{iclr2027_conference}

\appendix
\section{Positioning Against Prior Work}
\label{app:positioning}

Table~\ref{tab:positioning} separates the user input, generated representation, and execution requirements of relevant methods and standards.
Text conditioning, compact programs, material layers, and visual feedback each have precedents, so the intended comparison is the effect of \method's restricted layer vocabulary under matched generation and editing conditions, and Appendices~\ref{app:bpy} and~\ref{app:editing} provide those matched comparisons.
\begin{table}[t]
    \caption{Task and representation boundaries. User conditioning is distinct from textual program encoding: MultiMat consumes compact textual programs during synthesis without evaluating natural-language-conditioned generation. Language standards are representation precedents, not learned generation systems. Public code refers to a runnable implementation released by the authors, with MDL and MaterialX governed as open standards. These properties do not imply a quality ranking, and matched generation and editing comparisons appear in Appendices~\ref{app:bpy} and~\ref{app:editing}.}
    \label{tab:positioning}
    \centering
    \footnotesize
    \setlength{\tabcolsep}{2.5pt}
    \renewcommand{\arraystretch}{1.15}
    \begin{tabular}{@{}p{2.0cm}p{1.9cm}p{2.6cm}p{1.9cm}p{1.2cm}p{2.8cm}@{}}
        \toprule
        Work & User conditioning                                                                     & Generated representation & Execution dependence & Public code & Relevant distinction \\
        \midrule
        Raster generators~\citep{vecchio2024matfuse,vecchio2024stablematerials,kocsis2025intrinsix}
             & Text; additional modalities vary
             & PBR raster maps
             & Renderer consuming the maps
             & Yes
             & Material appearance without an explicit generative program                                                                                                     \\
        Conditional MatFormer~\citep{hu2023generating}
             & Text, image, or partial graph
             & Procedural node graph
             & Substance ecosystem
             & No
             & Prior text-conditioned procedural synthesis                                                                                                                    \\
        VLMaterial~\citep{li2025vlmaterial}
             & Image
             & Python program constructing a shader graph
             & Blender API
             & Yes
             & Learned image-to-program synthesis                                                                                                                             \\
        MultiMat~\citep{belouadi2025multimat}
             & Image or unconditional
             & CompactSBS, a compact YAML graph program
             & Substance Designer
             & No
             & Intermediate visual feedback and incremental validation                                                                                                        \\
        MatLayerNet
             & Text
             & Layer plans, per-layer parameters, and library-derived masks over PBR maps
             & MetaGPT multi-agent pipeline with a curated mask-generator library
             & No
             & Prior language-guided substrate, texture, and aging layers~\citep{cai2026matlayernet}                                                                          \\
        Material Apprentice~\citep{gupta2026apprentice}
             & Text; optional reference image; editing instruction
             & Expert process trace compiled to a shader graph
             & Blender API
             & Yes
             & Closest text-to-procedural system, using process retrieval                                                                                                     \\
        MDL~\citep{kettner2015mdl}
             & Author-supplied program
             & Declarative material with procedural functions
             & MDL-capable compiler or renderer
             & Open standard
             & Portable material language with layered scattering                                                                                                             \\
        MaterialX
             & Author-supplied graph
             & Material and look-development graph
             & MaterialX-capable tools
             & Open standard
             & Portable graph description and exchange~\citep{materialx}                                                                                                      \\
        \method
             & Text; subsequent layer revisions
             & Restricted field expressions and an ordered layer program
             & Standalone map executor
             & Yes
             & Studies layer-local authoring through an explicit restricted vocabulary                                                                                        \\
        \bottomrule
    \end{tabular}
\end{table}

\section{Example Program}
\label{app:example}

Listing~\ref{lst:example} gives the program for ``black shiny ceramic floor tiles,'' referenced in Section~\ref{sec:dsl-structure}.
Its text matches the final program in the retained \texttt{gemini-3.6-flash} seed-$123$ trajectory, ignoring outer whitespace.

\begin{lstlisting}[
    float=htbp,
    basicstyle=\ttfamily\scriptsize,
    breaklines=true,
    columns=fullflexible,
    keepspaces=true,
    caption={A complete two-layer example program for ``black shiny ceramic floor tiles.'' Black-glazed tiles composite \emph{over} a rough gray grout substrate. The \dsl{tileMask} lattice is shared by the upper layer's alpha mask and height expression, so editing that definition updates both fields.},
    label={lst:example},
    captionpos=b,
]
View(0, 0, 2, 2)
Define(tileMask, Bricks(brick_width=0.48, brick_height=0.48, offset=0, mortar=0.02, feather=0.008))
Define(glazeRipple, fBm(octaves=3, base_freq=7, to_01=True, seed=1145088217))
Define(glazeGrain, fBm(octaves=5, base_freq=35, to_01=True, seed=1007923142))
Define(groutNoise, fBm(octaves=4, base_freq=30, to_01=True, seed=891377204))
Material(
  Layer(1)
    .basecolor((40 + (groutNoise * 12)), (42 + (groutNoise * 12)), (45 + (groutNoise * 12)))
    .roughness((0.85 + (groutNoise * 0.1)))
    .height(0),
  Layer(tileMask)
    .basecolor((12 + (glazeRipple * 6)), (13 + (glazeRipple * 6)), (16 + (glazeRipple * 6)))
    .roughness((0.04 + (glazeGrain * 0.05)))
    .coat(0.95)
    .ior(1.53)
    .height(((tileMask * 0.016) + ((tileMask * glazeRipple) * 0.0012)))
)
\end{lstlisting}

\begin{figure}[htbp]
    \centering
    \begin{minipage}[t]{0.2\linewidth}
        \centering
        \includegraphics[width=\linewidth]{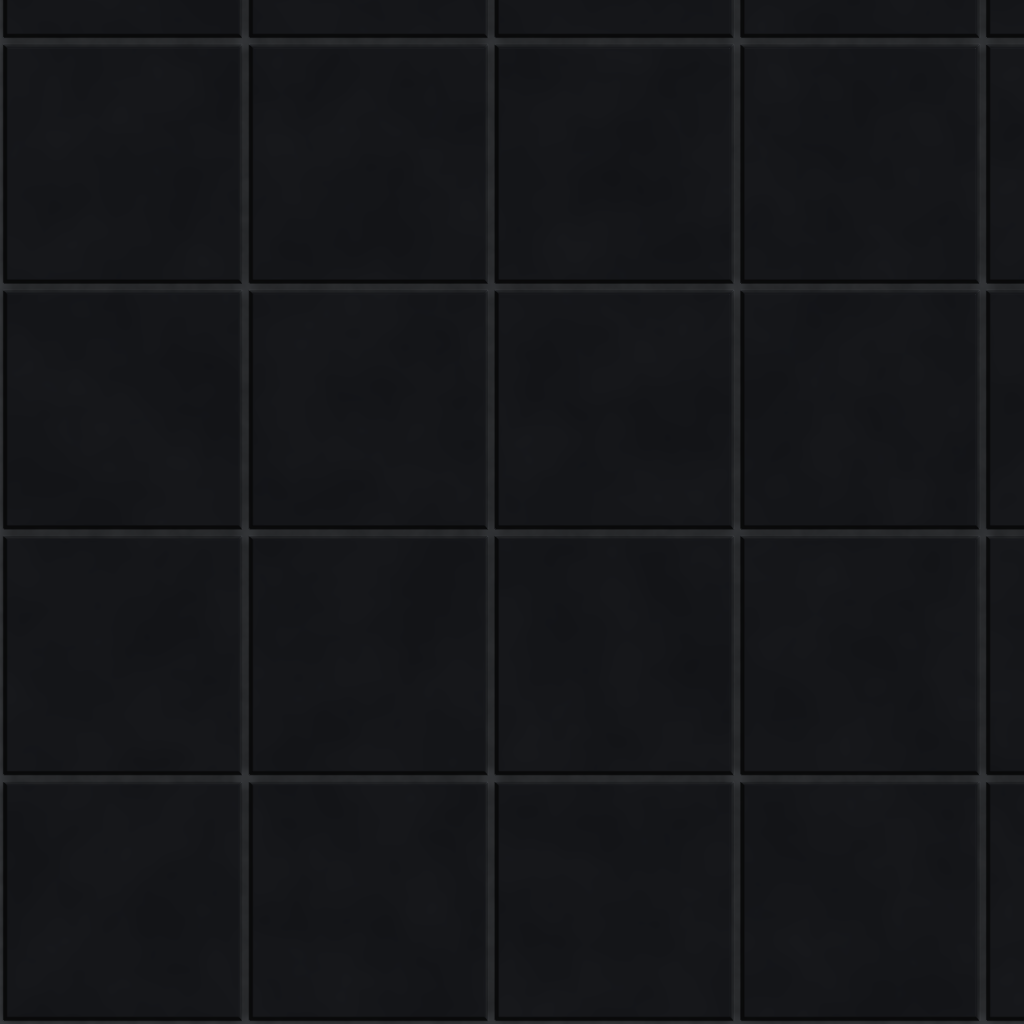}
        \small Original\\
        \scriptsize No edit
    \end{minipage}\hfill
    \begin{minipage}[t]{0.2\linewidth}
        \centering
        \includegraphics[width=\linewidth]{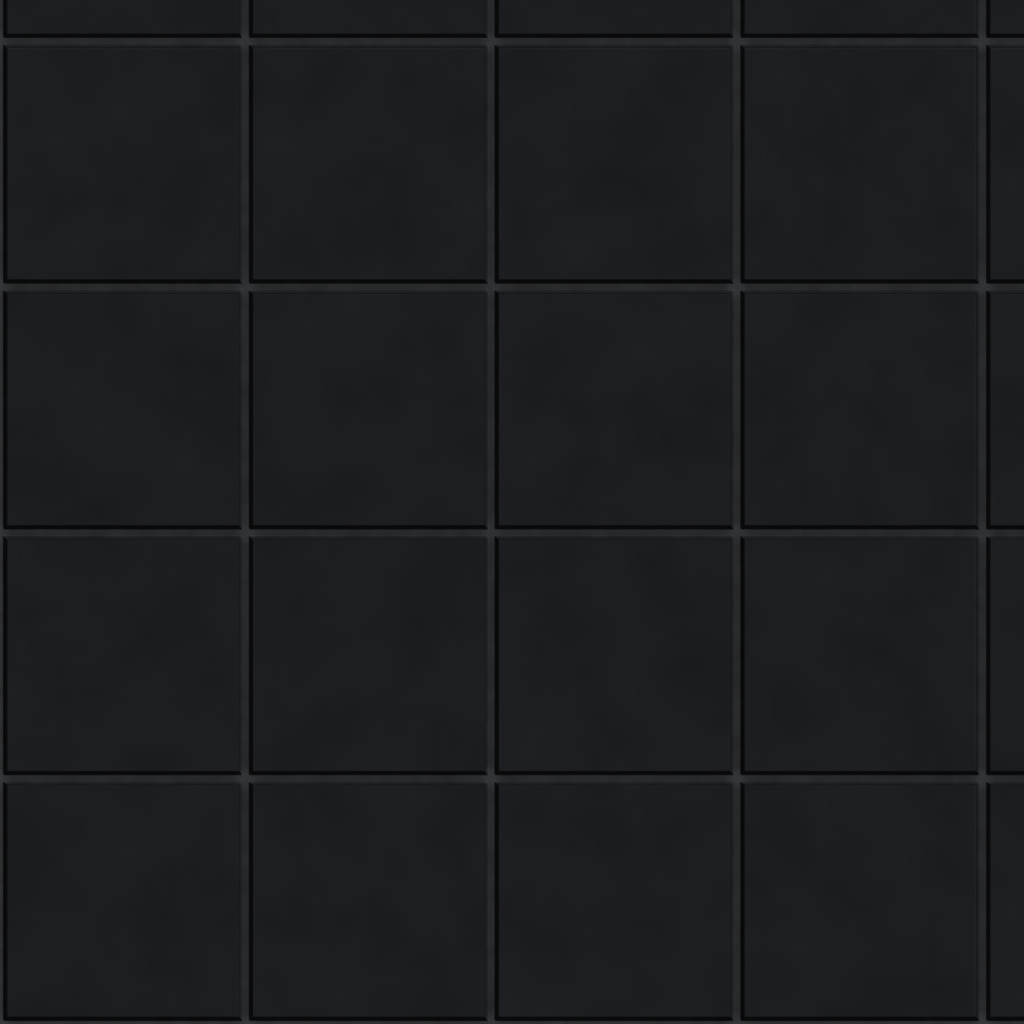}
        \small Higher roughness\\
        \scriptsize Offset: $0.04\!\rightarrow\!0.75$
    \end{minipage}\hfill
    \begin{minipage}[t]{0.2\linewidth}
        \centering
        \includegraphics[width=\linewidth]{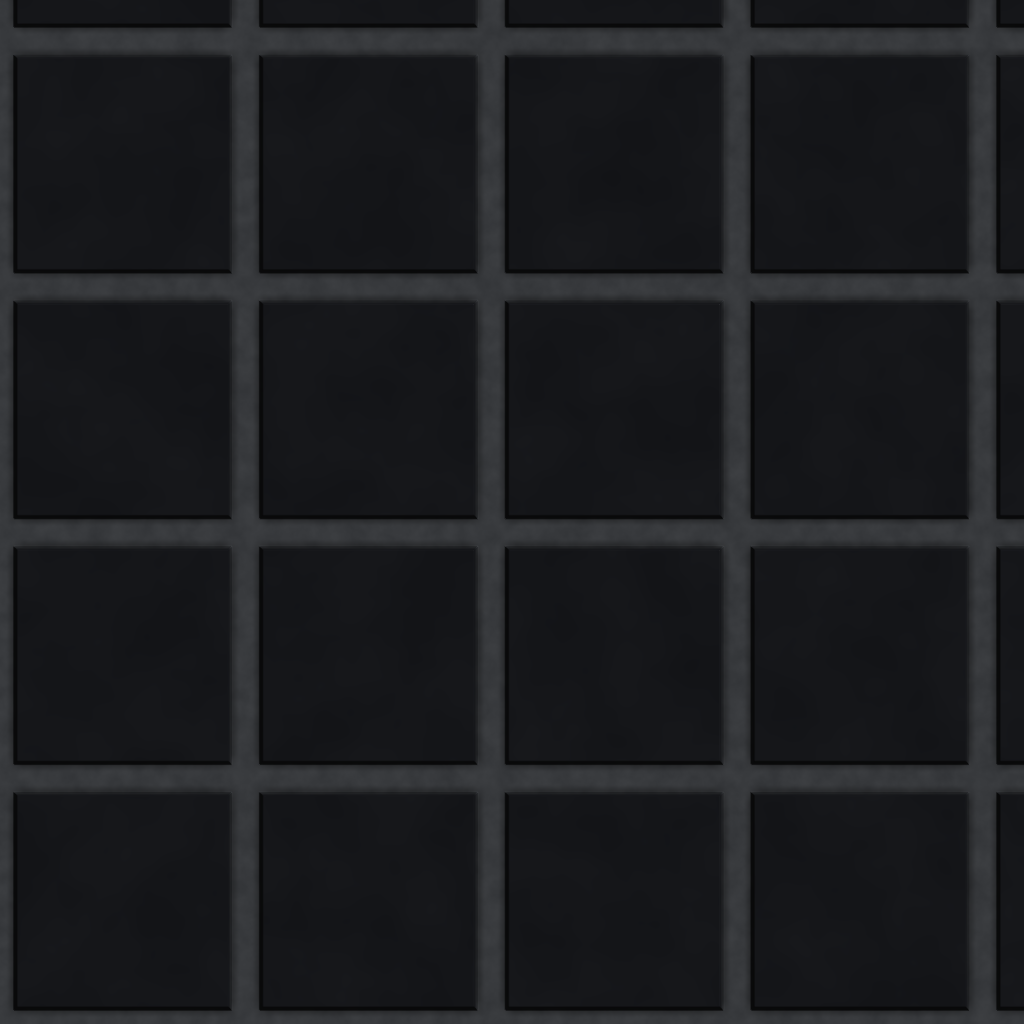}
        \small Wider grout\\
        \scriptsize Mortar: $0.02\!\rightarrow\!0.06$
    \end{minipage}\hfill
    \begin{minipage}[t]{0.2\linewidth}
        \centering
        \includegraphics[width=\linewidth]{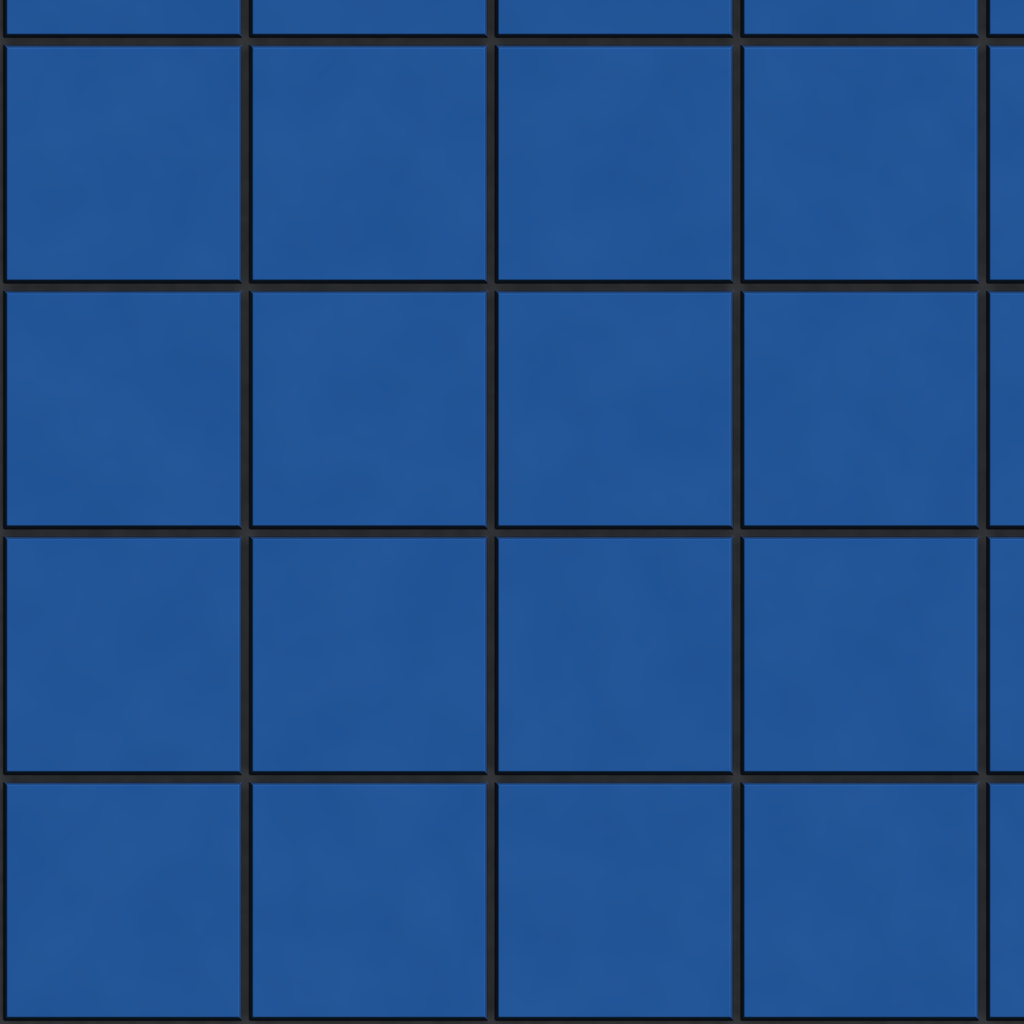}
        \small Blue glaze\\
        \scriptsize RGB offsets: $(25,75,140)$
    \end{minipage}
    \caption{\textbf{Explicit edits to one material program.} We manually change roughness, a shared grout parameter, or base-color offsets while preserving the remaining source and all three noise seeds. This is one illustrative case, not an automated-edit benchmark.}
    \label{fig:explicit-program-edits}
\end{figure}

Figure~\ref{fig:explicit-program-edits} uses this same source with three manually specified edits.
Re-evaluating the four programs in the released engine reproduces all eight previews byte-for-byte, so repeated evaluation is byte-identical.
At both resolutions, the roughness and color edits change only their respective composited channels. Widening grout deliberately changes the shared mask and its affected channels.
These checks establish the behavior of this case, not an editing-success rate or invariance across resolutions.

\section{Grammar Summary}
\label{app:grammar}

Listing~\ref{lst:grammar} summarizes canonical program syntax in EBNF, with keyword-argument and lexical productions condensed for space.
The full specification is included in the implementation's README, while the generator uses an operational language reference and the engine's parser.
The regression suite validates the serialization of $30$ generated examples against the full grammar and tests selected malformed expressions.
This finite test suite does not prove that every parser-accepted program is admitted by the canonical grammar or produces numerically valid fields.
Reference resolution and constructor constraints are checked while parsing, and channel values are clamped during evaluation (Section~\ref{sec:dsl-grammar}).

\begin{lstlisting}[
    float=tp,
    basicstyle=\ttfamily\scriptsize,
    breaklines=true,
    columns=fullflexible,
    keepspaces=true,
    caption={Condensed EBNF summary of the layered-material representation, with start symbol \texttt{program}. This listing omits lexical definitions and is not a standalone recognizer specification. The released source contains the full canonical grammar and its regression tests.},
    label={lst:grammar},
    captionpos=b,
]
program  = [ view ] , { define } , material ;
view     = "View" , "(" , real , "," , real , "," , real , "," , real , ")" ;
define   = "Define" , "(" , identifier , "," , expr , ")" ;
material = "Material" , "(" , [ layer , { "," , layer } ] , ")" ;
layer    = "Layer" , "(" , [ expr ] , ")" , { "." , channel } ;
channel  = "basecolor" , "(" , expr , "," , expr , "," , expr , ")"
         | ( "metallic" | "roughness" | "sheen" | "coat" | "transmission"
           | "subsurface" | "anisotropy" ) , "(" , expr , ")"
         | "ior" , "(" , expr , ")"
         | "emissive" , "(" , expr , "," , expr , "," , expr , "," , expr , ")"
         | "height" , "(" , expr , ")" ;

expr      = term , { ( "+" | "-" ) , term } ;
term      = factor , { ( "*" | "/" ) , factor } ;
factor    = [ "-" | "+" ] , power ;
power     = atom , [ "**" , factor ] ;
atom      = number | "pi" | "e" | reference | call | "(" , expr , ")" ;
reference = identifier ;

call = ( "X" | "Y" ) , "(" , ")"
     | ( "Abs" | "Sqrt" | "Floor" | "Ceil" | "Sin" | "Cos" ) , "(" , expr , ")"
     | "Log" , "(" , expr , [ "," , real ] , ")"
     | ( "Min" | "Max" ) , "(" , expr , "," , expr , ")"
     | "Threshold" , "(" , expr , { "," , kwarg } , ")"
     | ( "fBm" | "Worley" | "Bricks" | "Weave" ) , "(" , [ kwarg , { "," , kwarg } ] , ")"
     | ( "Translate" | "Scale" ) , "(" , expr , "," , real , "," , real , ")"
     | "Rotate" , "(" , expr , "," , real , ")"
     | "Fill" , "(" , shape , { "," , kwarg } , ")"
     | "Stroke" , "(" , shape , "," , pos_real , [ "," , kwarg ] , ")" ;
shape = "Rect" , "(" , real , "," , real , "," , pos_real , "," , pos_real ,
        [ "," , non_neg_real ] , ")"
      | "Ellipse" , "(" , real , "," , real , "," , pos_real , "," , pos_real , ")"
      | "Path" , "(" , real , "," , real , { "," , segment } , ")" ;
segment = "LineTo" , "(" , real , "," , real , ")"
        | "CubicTo" , "(" , real , "," , real , "," , real , "," , real , "," , real ,
          "," , real , ")" ;
kwarg = identifier , "=" , ( expr | enum ) ;
enum  = '"euclidean"' | '"manhattan"' | '"chebyshev"'
      | '"F1"' | '"F2"' | '"F2-F1"' | '"F2+F1"'
      | '"NonZero"' | '"EvenOdd"' | '"row"' | '"column"' | "True" | "False" ;
\end{lstlisting}

\section{Cross-engine Parity}
\label{app:engine}

The Python reference generates expected values for a shared fixture suite, and the TypeScript port is compared against those values.
The suite distinguishes exact comparisons, numerical tolerances, and excluded cases rather than asserting universal cross-engine identity.

\paragraph{Exact fixture comparisons.} Selected arithmetic operations, leaves, thresholding, hard-edged brick and weave patterns, separable transforms, shape fill decisions, canonical serialization, and packed surface-channel bytes use exact comparisons.
Hard-boundary samples are chosen away from shape edges, where coordinate precision can change a fill decision.
These checks establish agreement on the sampled cases.

\paragraph{Approximate fixture comparisons.} Scalar expressions involving transcendental functions and multi-octave noise use a tolerance of approximately $10^{-11}$.
Feathered shapes, weave coverage, gridded \dsl{fBm}, and relief use tolerances of approximately $10^{-6}$ to $10^{-5}$ to accommodate float32 storage and evaluation differences.
The thresholds are test tolerances, not error bounds proved for arbitrary programs.

\paragraph{Excluded cases and export differences.} Worley uses a 64-bit hash in Python and a 32-bit hash in TypeScript, so identical seeds do not imply identical fields for this primitive.
The browser's WebGPU preview is outside the cross-engine comparison, while CPU worker backends have separate within-TypeScript parity tests.
Browser channel packing quantizes IOR over $[1,3]$, whereas the Python engine supports IOR values above $3$ in its floating-point export.
The Python HDR fallback clips negative height values, which require EXR to preserve.
These differences limit claims of interchangeable numerical output.

\paragraph{Browser viewer.} The viewer loads supported programs and exposes source editing and lighting controls without a digital content creation application.
It provides an execution and inspection interface, rather than evidence that final images match Blender under different shading and lighting.
The released source pins the engine revision and includes the fixture suite with its tolerances.

\section{Refinement Depth, Selection, and Polish}
\label{app:refine}

Table~\ref{tab:stages} summarizes the flagship trajectory, Table~\ref{tab:refine-depth} gives its round-wise scores, and Tables~\ref{tab:selector} and~\ref{tab:polish} separate selection from seed search.
The trajectory improves several mean metrics through round $5$, but the round-wise changes are not uniformly monotonic.
All reported mean seed-search deltas are positive, ranging from $+0.04$ to $+2.22$ points, without implying that every run improves.

The selector's flat BLIPScore win rate against taking the last revision is $178/423$ ($42.1\%$), with $137$ losses ($32.4\%$) and $108$ ties ($25.5\%$).
Thus a win rate below $50\%$ does not mean that selection loses on most runs.
Its mean flat judge score decreases by $1.04$ points relative to the last revision, and the complete pipeline scores $67.01$ compared with $67.61$ at round $5$.
The per-metric oracle rows use evaluation scores to choose a different best program for each metric and therefore provide upper bounds, not feasible inference baselines.
Figure~\ref{fig:refine} illustrates a selected trajectory rather than typical monotonic improvement.

Table~\ref{tab:polish-budget} replays prefixes of the stored seed sweeps without new generation or scoring.
For each budget $K$, the replay retains the best recorded quick score among the originals and the first $K$ variants of each eligible candidate.
The gain is nonnegative because the originals remain eligible.
Ten variants recover $48\%$ of the full recorded surrogate gain and 500 recover $94\%$.
Re-scoring the replayed winners on final metrics confirms the tradeoff: a single variant per candidate reaches $55.47$ flat BLIPScore, $99\%$ of the full sweep's $56.06$, while flat CLIPScore and VQAScore means vary by at most $0.8$ points across budgets, the judge column of Table~\ref{tab:polish-budget} shows the same flat pattern at a $1.0$-point spread over $K{=}1$ to $500$, and the sweep's marginal final-metric BLIPScore yield beyond one variant is $0.6$ points.
Table~\ref{tab:polish} reports the external-metric deltas of the full sweep, and the final-metric columns of Table~\ref{tab:polish-budget} extend them to truncated budgets.

\begin{table}[htbp]
    \caption{Stage III's sweep budget replayed from the stored per-variant quick-scorer scores of the flagship's $423$ runs, with the replay winners re-scored on final flat metrics at each truncated budget. The gain is the quick-scorer margin over the run's best pre-polish program (median $31.5$), in quick-scorer points rather than evaluation metrics, non-negative because the selection keeps the original unless a variant beats it. ``Runs at best'' is the share of runs whose truncated sweep already contains the full sweep's winner. Flat BLIP and Flat VQA are external evaluation means of the replay winners (mean of per-prompt seed means over $141$ prompts and $3$ runs), with $K{=}1000$ giving the committed final means. The flat judge column is a single scoring session covering $K{=}1$ to $500$, so its entries compare with each other but not with the $K{=}1000$ judge mean from the original evaluation, and its $K{=}1000$ cell is therefore dashed. Polish wall share projects the measured $33.5$ hour full sweep, $75\%$ of the $44.7$ hour total, linearly in scored candidates.}
    \label{tab:polish-budget}
    \centering
    \footnotesize
    \setlength{\tabcolsep}{3.2pt}
    \resizebox{\linewidth}{!}{%
    \begin{tabular}{l ccc cccc}
        \toprule
        Budget $K$ & Mean gain & \% of $K{=}1000$ & Runs at best (\%) & Flat BLIP & Flat VQA & Flat judge & Polish wall share (\%) \\
        \midrule
        $1$        & $+0.25$   & 18               & 0                 & 55.47     & 53.97    & 70.81      & 1                        \\
        $10$       & $+0.68$   & 48               & 1                 & 56.44     & 54.17    & 70.35      & 3                        \\
        $50$       & $+0.96$   & 67               & 6                 & 55.99     & 53.94    & 70.22      & 13                       \\
        $100$      & $+1.07$   & 75               & 9                 & 56.35     & 54.37    & 70.95      & 23                       \\
        $250$      & $+1.21$   & 85               & 22                & 56.42     & 54.36    & 71.22      & 43                       \\
        $500$      & $+1.34$   & 94               & 50                & 56.37     & 54.71    & 70.25      & 60                       \\
        $1000$     & $+1.42$   & 100              & 100               & 56.06     & 54.71    & --         & 75                       \\
        \bottomrule
    \end{tabular}}
\end{table}

\begin{table}[htbp]
    \caption{Refinement rounds for \texttt{gemini-3.6-flash} (the Flat and Staged column groups give both layouts, mean over all runs). Round $0$ is the initial program's absolute score. Rounds $1$--$5$ are signed deltas from it. Best round per layout half in \textbf{bold}.}
    \label{tab:refine-depth}
    \centering
    \footnotesize
    \setlength{\tabcolsep}{4pt}
    \begin{tabular}{l cccc cccc}
        \toprule
              & \multicolumn{4}{c}{Flat} & \multicolumn{4}{c}{Staged}                                                                                                                 \\
        \cmidrule(lr){2-5}\cmidrule(lr){6-9}
        Round & BLIPScore                & CLIPScore                  & VQAScore       & Judge            & BLIPScore        & CLIPScore        & VQAScore         & Judge            \\
        \midrule
        0     & 48.31                    & 27.70                      & \textbf{53.34} & 61.35            & 30.95            & 24.37            & 52.67            & 52.98            \\
        1     & $+$1.30                  & $+$0.13                    & $-$0.76        & $+$3.71          & $+$0.57          & $+$0.27          & $+$0.02          & $+$1.80          \\
        2     & $+$1.74                  & $+$0.22                    & $-$0.41        & $+$4.19          & $+$1.38          & $+$0.40          & $+$0.22          & $+$2.28          \\
        3     & $+$1.89                  & $+$0.31                    & $-$0.40        & $+$3.87          & $+$1.66          & $+$0.46          & $+$0.37          & $+$3.32          \\
        4     & $+$2.95                  & $+$0.38                    & $-$0.75        & $+$3.32          & $+$2.02          & $+$0.46          & $-$0.02          & $+$3.04          \\
        5     & \textbf{$+$3.55}         & \textbf{$+$0.55}           & $-$0.41        & \textbf{$+$6.26} & \textbf{$+$2.28} & \textbf{$+$0.54} & \textbf{$+$0.76} & \textbf{$+$3.84} \\
        \bottomrule
    \end{tabular}
\end{table}

\begin{table}[htbp]
    \caption{Quick-scorer selection over the trajectory of \texttt{gemini-3.6-flash} in both layouts. Selected is the MobileCLIP2 argmax over six programs, last uses round $5$, and oracle uses each evaluation metric to select its own best program. The oracle is an upper bound unavailable to the deployed selector. Win rate counts strictly higher scores than last, with ties not treated as losses, and $\Delta$ is the mean signed difference.}
    \label{tab:selector}
    \centering
    \footnotesize
    \setlength{\tabcolsep}{4pt}
    \begin{tabular}{l cccc cccc}
        \toprule
                 & \multicolumn{4}{c}{Flat} & \multicolumn{4}{c}{Staged}                                                                   \\
        \cmidrule(lr){2-5}\cmidrule(lr){6-9}
        Program  & BLIPScore                & CLIPScore                  & VQAScore & Judge   & BLIPScore & CLIPScore & VQAScore & Judge   \\
        \midrule
        First    & 48.31                    & 27.70                      & 53.34    & 61.35   & 30.95     & 24.37     & 52.67    & 52.98   \\
        Selected & 54.17                    & 28.71                      & 53.97    & 66.56   & 35.50     & 25.25     & 54.07    & 57.20   \\
        Last     & 51.86                    & 28.25                      & 52.93    & 67.61   & 33.23     & 24.91     & 53.43    & 56.82   \\
        Oracle   & 64.80                    & 30.01                      & 60.71    & 79.24   & 44.48     & 26.31     & 60.42    & 69.07   \\
        \midrule
        Win rate & 42.1                     & 41.4                       & 38.1     & 30.5    & 42.8      & 43.3      & 36.4     & 35.0    \\
        $\Delta$ & $+$2.31                  & $+$0.45                    & $+$1.04  & $-$1.04 & $+$2.27   & $+$0.34   & $+$0.64  & $+$0.38 \\
        \bottomrule
    \end{tabular}
\end{table}

\begin{table}[htbp]
    \caption{Signed mean deltas from seed search ($1000$ variants, original retained) applied to the first, selected, and last trajectory programs, with a Mean row averaging those 3 conditions. All retained mean deltas are positive, but no per-run or statistical improvement guarantee follows.}
    \label{tab:polish}
    \centering
    \footnotesize
    \setlength{\tabcolsep}{4pt}
    \begin{tabular}{l cccc cccc}
        \toprule
                 & \multicolumn{4}{c}{Flat} & \multicolumn{4}{c}{Staged}                                                                   \\
        \cmidrule(lr){2-5}\cmidrule(lr){6-9}
        Polished & BLIPScore                & CLIPScore                  & VQAScore & Judge   & BLIPScore & CLIPScore & VQAScore & Judge   \\
        \midrule
        First    & $+$2.22                  & $+$0.17                    & $+$0.49  & $+$0.64 & $+$0.23   & $+$0.08   & $+$0.33  & $+$0.27 \\
        Selected & $+$1.54                  & $+$0.12                    & $+$0.22  & $+$0.13 & $+$0.17   & $+$0.04   & $+$0.09  & $+$0.59 \\
        Last     & $+$1.63                  & $+$0.18                    & $+$0.36  & $+$0.11 & $+$1.09   & $+$0.09   & $+$0.25  & $+$0.90 \\
        Mean     & $+$1.80                  & $+$0.16                    & $+$0.36  & $+$0.30 & $+$0.50   & $+$0.07   & $+$0.22  & $+$0.59 \\
        \bottomrule
    \end{tabular}
\end{table}

\section{Runtime Analysis}
\label{app:runtime}

Table~\ref{tab:runtime} reports recorded wall-clock times for the complete authoring runs, with every stage running sequentially in one process per prompt.
For the flagship, the median run takes $4.6$ minutes: $40$ seconds for the six generation calls, $42$ seconds for the six critique calls (one per trajectory program, the last triggering no revision), $1.7$ seconds for trajectory selection, and $181$ seconds for seed search over 2 or 3 candidates, with the remainder spent on lazy scorer loading and program canonicalization.
The initial proposal has a median latency of $11$ seconds, and each revision or critique call takes 5 to 7 seconds.
Each seed sweep renders $1001$ candidates at $256{\times}256$ and scores them in a median of $70$ seconds, about $70$ milliseconds per candidate, covering a preview evaluation and one text--image scorer encode.
Critique and trajectory-selection renders use $512{\times}512$ previews, and channel statistics are computed on a $128{\times}128$ grid.
The seed search is roughly constant across backbones, so end-to-end differences reflect the language-model calls, whose combined medians reach $910$ seconds for the slowest backbone.
The recorded hardware and scorer-device specifications are unavailable, so these wall-clock figures should not be used as cross-system latency comparisons.

\begin{table}[htbp]
    \caption{Recorded authoring latency per backbone over $423$ runs each, with 3 stochastic runs per prompt. The polish share is the fraction of total wall time spent in Stage III seed search, and p90 is the 90th percentile of that stage's duration.}
    \label{tab:runtime}
    \centering
    \footnotesize
    \setlength{\tabcolsep}{6pt}
    \begin{tabular}{l ccc}
        \toprule
        Backbone               & Median total (min) & Median polish share (\%) & Polish p90 (min) \\
        \midrule
        gemini-3.6-flash       & 4.6                & 68                       & 5.6              \\
        gpt-5.6-luna           & 6.3                & 43                       & 5.2              \\
        gemma-4-26b-a4b-it     & 7.7                & 33                       & 3.6              \\
        glm-5.2                & 10.3               & 27                       & 4.7              \\
        qwen3.6-35b-a3b        & 12.8               & 20                       & 3.6              \\
        deepseek-v4-flash-0731 & 20.0               & 18                       & 7.2              \\
        \bottomrule
    \end{tabular}
\end{table}

\section{Exploratory Component Comparisons}
\label{app:ablation}

Table~\ref{tab:ablation-matched} reports the settings-matched component rerun, and Tables~\ref{tab:ablation} and~\ref{tab:ablation-wilcoxon} retain the earlier exploratory records that motivated it.

\paragraph{Settings-matched rerun.}
The rerun re-runs six configurations under settings identical to the recorded main runs: generator \texttt{gpt-5.6-luna}, temperature fixed at $0.7$ on both sides, procedural seeds $42$, $123$, and $2026$, reasoning effort high, and every other recorded field unchanged.
It covers a deterministic $40$-prompt stratified subset ($10$ per source, selection rule committed before any scoring) with three seeds per configuration, $720$ runs in total and no failures.
Cells are per-prompt seed-mean deltas against the recorded three-seed \texttt{gpt-5.6-luna} mains restricted to the same subset, with $95\%$ percentile bootstrap intervals and Holm-corrected Wilcoxon stars.
Variant judge cells stay dashes on purpose: the variant judge scores come from a different scoring session than the reference scores, and judge scores from different sessions are not level-comparable (Appendix~\ref{app:e2}), so a cross-session judge delta would not isolate component effects.

Under matched settings the load-bearing components are the prompt playbook, at $-4.8$ flat ($[-9,-0]$) and $-5.7$ staged BLIPScore, and the render diagnostics panel, at $-5.8$ flat ($[-11,-1]$) and $-7.3$ staged BLIPScore.
After Holm correction over the table, the only surviving cells are the diagnostics panel's staged BLIPScore ($p{<}0.001$) and the render critique's flat CLIPScore ($-1.2$).
Removing the render critique costs $-3.1$ flat BLIPScore with an interval crossing zero, where the exploratory records showed $-6.8$, so the exploratory effect mixed the component with the temperature difference.
Removing the code critique moves flat BLIPScore by $+1.5$ and is not significant, where the exploratory records showed $+7.0$, so that anomaly also fades under matched sampling.
Few-shot removal stays small at $-1.9$ flat, and removing the code critique together with the diagnostics panel stays at $-0.8$ flat and $-4.3$ staged.

\begin{table}[htbp]
    \caption{Settings-matched component reruns: six configurations re-run at the recorded main-run settings (generator \texttt{gpt-5.6-luna}, temperature $0.7$ on both sides, seeds $42$, $123$, and $2026$) on the frozen $40$-prompt stratified subset, three seeds per configuration ($720$ runs). The first row gives the reference absolute means (recorded three-seed mains on the same subset, judge included), and other rows are signed per-prompt seed-mean deltas with $95\%$ percentile bootstrap intervals and Holm-corrected Wilcoxon stars ($^{*}p{<}0.05$, $^{**}p{<}0.01$, $^{***}p{<}0.001$). Variant judge cells stay dashed because the variant scores come from a different scoring session than the reference scores, and judge scores from different sessions are not level-comparable.}
    \label{tab:ablation-matched}
    \centering
    \footnotesize
    \setlength{\tabcolsep}{3.5pt}
    \resizebox{\linewidth}{!}{%
\begin{tabular}{l cccc cccc}
\toprule
 & \multicolumn{4}{c}{Flat} & \multicolumn{4}{c}{Staged} \\
\cmidrule(lr){2-5}\cmidrule(lr){6-9}
    Ablation & BLIP & CLIP & VQA & Judge & BLIP & CLIP & VQA & Judge \\
\midrule
    Full (ours, subset) & 52.83 & 27.72 & 50.27 & 57.47 & 31.28 & 24.49 & 49.02 & 43.41 \\
    w/o playbook & -4.8 [-9,-0] & -0.6 [-1,-0] & -0.4 [-3,+3] & -- & -5.7 [-10,-2] & -0.9 [-2,-0] & +0.1 [-3,+3] & -- \\
    w/o few-shot & -1.9 [-7,+3] & -0.1 [-1,+0] & +0.4 [-2,+3] & -- & -2.9 [-8,+2] & -0.4 [-1,+0] & +1.9 [-1,+5] & -- \\
    \midrule
    w/o render critique & -3.1 [-9,+3] & -1.2$^{**}$ [-2,-1] & -0.5 [-3,+2] & -- & -1.1 [-6,+4] & -0.5 [-1,+0] & +2.3 [-1,+5] & -- \\
    w/o code critique & +1.5 [-4,+7] & +0.2 [-0,+1] & +2.3 [-1,+6] & -- & -1.2 [-5,+3] & +0.0 [-1,+1] & +2.7 [-1,+7] & -- \\
    w/o diagnostics & -5.8 [-11,-1] & -0.3 [-1,+0] & -1.4 [-5,+2] & -- & -7.3$^{***}$ [-11,-4] & -0.6 [-1,-0] & -0.9 [-4,+2] & -- \\
    w/o code critique + diagnostics & -0.8 [-6,+5] & +0.6 [-0,+1] & +1.6 [-2,+5] & -- & -4.3 [-9,+1] & -0.4 [-1,+0] & -0.1 [-4,+3] & -- \\
\bottomrule
\end{tabular}}
\end{table}

\paragraph{Exploratory records.}
Table~\ref{tab:ablation} retains the configuration comparisons discussed in Section~\ref{sec:results-ablation}, using the \texttt{gpt-5.6-luna} generator and one authoring run per prompt.
The full reference is the seed-$42$ main run at temperature $0.7$, whereas variant records leave temperature unset and inherit the provider default.
Model, reasoning effort, revision depth, repair budget, selection and polish configuration, procedural seed, and the prompt list are otherwise identical, so the rows remain paired by prompt but do not isolate component effects.
The render-only critic has the largest observed flat judge delta ($+8.50$), and the render-plus-statistics critic has the largest flat BLIPScore delta ($+6.96$).
Paired per-prompt Wilcoxon tests accompany these comparisons (Table~\ref{tab:ablation-wilcoxon}): removing the render critique decreases flat BLIPScore ($p{=}0.015$) and flat CLIPScore ($p{<}0.001$), withholding source code raises the mean in all eight cells and significantly in seven, and comparisons that remain significant after within-configuration Bonferroni correction include the flat-judge delta without source access and the judge deltas of the text-only GLM and Gemma critics.
These exploratory deltas motivated the matched rerun of Table~\ref{tab:ablation-matched}, whose matched values supersede them as evidence.

\begin{table}[htbp]
    \caption{Exploratory component configurations in both layouts. The first row contains the full reference's absolute means for one authoring run per prompt, and subsequent rows are signed deltas. Diagnostics denotes the material-statistics panel.}
    \label{tab:ablation}
    \centering
    \footnotesize
    \setlength{\tabcolsep}{3.5pt}
    \resizebox{\linewidth}{!}{%
        \begin{tabular}{l cccc cccc}
            \toprule
                                & \multicolumn{4}{c}{Flat} & \multicolumn{4}{c}{Staged}                                                                   \\
            \cmidrule(lr){2-5}\cmidrule(lr){6-9}
            Configuration       & BLIPScore                & CLIPScore                  & VQAScore & Judge   & BLIPScore & CLIPScore & VQAScore & Judge   \\
            \midrule
            Full (ours)         & 51.23                    & 27.53                      & 50.01    & 56.76   & 32.90     & 24.48     & 49.95    & 47.60   \\
            \midrule
            w/o playbook        & $-$4.99                  & $-$0.52                    & $-$1.47  & $-$0.08 & $-$4.66   & $-$0.54   & $-$1.30  & $-$1.87 \\
            w/o few-shot        & $-$1.06                  & $+$0.32                    & $+$1.60  & $+$0.69 & $-$2.53   & $+$0.18   & $+$2.06  & $+$2.40 \\
            w/o both            & $+$0.52                  & $+$0.31                    & $+$0.78  & $+$0.21 & $+$0.17   & $+$0.14   & $+$0.61  & $+$0.64 \\
            \midrule
            w/o render critique & $-$6.84                  & $-$1.03                    & $+$0.08  & $-$2.42 & $-$4.27   & $-$0.55   & $-$0.10  & $-$1.64 \\
            w/o code critique   & $+$6.96                  & $+$0.99                    & $+$2.57  & $+$7.59 & $+$4.10   & $+$0.96   & $+$3.06  & $+$4.72 \\
            w/o diagnostics     & $-$1.32                  & $-$0.06                    & $-$0.34  & $+$2.75 & $-$3.67   & $-$0.21   & $+$0.29  & $-$0.71 \\
            w/o render + code   & $-$4.10                  & $+$0.30                    & $+$1.32  & $+$2.12 & $-$0.83   & $+$0.29   & $+$1.43  & $+$2.05 \\
            w/o render + diag.  & $-$4.92                  & $-$0.36                    & $-$0.70  & $-$3.28 & $-$4.04   & $-$0.32   & $-$1.59  & $-$1.13 \\
            w/o code + diag.    & $+$2.73                  & $+$1.62                    & $+$2.71  & $+$8.50 & $+$3.76   & $+$1.07   & $+$2.98  & $+$4.45 \\
            \bottomrule
        \end{tabular}}
\end{table}

\begin{table}[htbp]
    \caption{Paired two-sided Wilcoxon tests for the exploratory comparisons of Tables~\ref{tab:ablation} and~\ref{tab:crosscritic} against the \texttt{gpt-5.6-luna} seed-$42$ reference ($n{=}141$ paired prompts). Each cell repeats the signed mean delta with the $p$-value in parentheses, and the critic rows correspond to Table~\ref{tab:crosscritic}.}
    \label{tab:ablation-wilcoxon}
    \centering
    \footnotesize
    \setlength{\tabcolsep}{5pt}
    {\scriptsize (a) Flat layout.}\\[2pt]
        \begin{tabular}{l cccc}
            \toprule
            Ablation & BLIP & CLIP & VQA & Judge \\
            \midrule
            w/o playbook & $-$5.0 (.047) & $-$0.5 (.070) & $-$1.5 (.157) & $-$0.1 (.703) \\
            w/o few-shot & $-$1.1 (.483) & $+$0.3 (.292) & $+$1.6 (.270) & $+$0.7 (.436) \\
            w/o playbook $+$ few-shot & $+$0.5 (.999) & $+$0.3 (.398) & $+$0.8 (.547) & $+$0.2 (.540) \\
            \midrule
            w/o render critique & $-$6.8 (.015) & $-$1.0 ($<$0.001) & $+$0.1 (.973) & $-$2.4 (.428) \\
            w/o code critique & $+$7.0 (.022) & $+$1.0 (.007) & $+$2.6 (.033) & $+$7.6 ($<$0.001) \\
            w/o diagnostics & $-$1.3 (.358) & $-$0.1 (.746) & $-$0.3 (.706) & $+$2.8 (.115) \\
            w/o render $+$ code & $-$4.1 (.302) & $+$0.3 (.610) & $+$1.3 (.263) & $+$2.1 (.211) \\
            w/o render $+$ diag. & $-$4.9 (.126) & $-$0.4 (.132) & $-$0.7 (.501) & $-$3.3 (.066) \\
            w/o code $+$ diag. & $+$2.7 (.389) & $+$1.6 ($<$0.001) & $+$2.7 (.019) & $+$8.5 ($<$0.001) \\
            \midrule
            critic = Gemini & $+$3.6 (.014) & $+$1.0 (.010) & $+$1.7 (.184) & $+$5.1 (.004) \\
            critic = GLM & $-$8.6 (.001) & $-$0.5 (.061) & $+$0.9 (.782) & $+$7.2 ($<$0.001) \\
            critic = DeepSeek & $-$0.4 (.707) & $-$0.1 (.373) & $+$1.2 (.332) & $-$2.8 (.175) \\
            critic = Qwen & $-$5.7 (.061) & $-$0.5 (.082) & $-$1.2 (.356) & $-$4.7 (.057) \\
            critic = Gemma & $-$6.9 (.067) & $-$0.3 (.174) & $+$0.4 (.789) & $+$11.5 ($<$0.001) \\
            \bottomrule
        \end{tabular}\\[6pt]
    {\scriptsize (b) Staged layout.}\\[2pt]
        \begin{tabular}{l cccc}
            \toprule
            Ablation & BLIP & CLIP & VQA & Judge \\
            \midrule
            w/o playbook & $-$4.7 (.046) & $-$0.5 (.092) & $-$1.3 (.322) & $-$1.9 (.282) \\
            w/o few-shot & $-$2.5 (.368) & $+$0.2 (.827) & $+$2.1 (.172) & $+$2.4 (.202) \\
            w/o playbook $+$ few-shot & $+$0.2 (.231) & $+$0.1 (.419) & $+$0.6 (.955) & $+$0.6 (.675) \\
            \midrule
            w/o render critique & $-$4.3 (.172) & $-$0.6 (.050) & $-$0.1 (.790) & $-$1.6 (.383) \\
            w/o code critique & $+$4.1 (.106) & $+$1.0 (.002) & $+$3.1 (.011) & $+$4.7 (.007) \\
            w/o diagnostics & $-$3.7 (.124) & $-$0.2 (.207) & $+$0.3 (.772) & $-$0.7 (.761) \\
            w/o render $+$ code & $-$0.8 (.847) & $+$0.3 (.270) & $+$1.4 (.393) & $+$2.0 (.394) \\
            w/o render $+$ diag. & $-$4.0 (.039) & $-$0.3 (.275) & $-$1.6 (.135) & $-$1.1 (.510) \\
            w/o code $+$ diag. & $+$3.8 (.256) & $+$1.1 ($<$0.001) & $+$3.0 (.024) & $+$4.5 (.012) \\
            \midrule
            critic = Gemini & $+$1.8 (.182) & $+$0.8 ($<$0.001) & $+$2.0 (.142) & $+$3.0 (.100) \\
            critic = GLM & $-$2.2 (.321) & $-$0.1 (.625) & $+$2.2 (.098) & $+$10.2 ($<$0.001) \\
            critic = DeepSeek & $-$0.0 (.989) & $+$0.1 (.851) & $+$1.2 (.496) & $+$0.6 (.817) \\
            critic = Qwen & $-$4.4 (.065) & $-$0.3 (.343) & $+$1.7 (.494) & $-$2.4 (.192) \\
            critic = Gemma & $+$0.5 (.966) & $+$0.2 (.677) & $+$1.8 (.621) & $+$13.8 ($<$0.001) \\
            \bottomrule
        \end{tabular}
\end{table}

\section{Cross-critic Study}
\label{app:crosscritic}

Table~\ref{tab:crosscritic} uses \texttt{gpt-5.6-luna} as generator and varies the critic, with one authoring run per prompt.
DeepSeek and GLM omit image input, so their feedback also differs in modality.
As in the component table, the self-critique reference fixes temperature at $0.7$ while the cross-critic variants leave it unset.
The observed rankings therefore combine critic choice, input availability, and potentially sampling settings.
Gemini has the largest flat BLIPScore delta ($+3.56$), while Gemma has the largest judge deltas ($+11.53$ flat and $+13.77$ staged).
These layout- and metric-dependent rankings do not identify a uniformly best critic.

\begin{table}[htbp]
    \caption{Exploratory cross-critic configurations with the \texttt{gpt-5.6-luna} generator. The reference contains absolute means, and subsequent rows contain signed deltas. Critic modalities differ as described above. Largest observed delta per column is in \textbf{bold}.}
    \label{tab:crosscritic}
    \centering
    \footnotesize
    \setlength{\tabcolsep}{4pt}
    \resizebox{\linewidth}{!}{%
        \begin{tabular}{l cccc cccc}
            \toprule
                       & \multicolumn{4}{c}{Flat} & \multicolumn{4}{c}{Staged}                                                                                                                     \\
            \cmidrule(lr){2-5}\cmidrule(lr){6-9}
            Critic     & BLIPScore                & CLIPScore                  & VQAScore         & Judge             & BLIPScore        & CLIPScore        & VQAScore         & Judge             \\
            \midrule
            Self (gpt) & 51.23                    & 27.53                      & 50.01            & 56.76             & 32.90            & 24.48            & 49.95            & 47.60             \\
            \midrule
            Gemini     & \textbf{$+$3.56}         & \textbf{$+$0.99}           & \textbf{$+$1.74} & $+$5.08           & \textbf{$+$1.77} & \textbf{$+$0.83} & $+$2.01          & $+$2.99           \\
            DeepSeek   & $-$0.40                  & $-$0.13                    & $+$1.22          & $-$2.80           & $-$0.04          & $+$0.09          & $+$1.17          & $+$0.65           \\
            GLM        & $-$8.57                  & $-$0.51                    & $+$0.87          & $+$7.16           & $-$2.20          & $-$0.14          & \textbf{$+$2.23} & $+$10.20          \\
            Qwen       & $-$5.75                  & $-$0.53                    & $-$1.22          & $-$4.71           & $-$4.37          & $-$0.34          & $+$1.75          & $-$2.44           \\
            Gemma      & $-$6.92                  & $-$0.28                    & $+$0.40          & \textbf{$+$11.53} & $+$0.53          & $+$0.20          & $+$1.75          & \textbf{$+$13.77} \\
            \bottomrule
        \end{tabular}}
\end{table}

\section{Exploratory Parameter Search}
\label{app:cma}

\begin{figure}[htbp]
    \centering
    \includegraphics[width=0.26\linewidth]{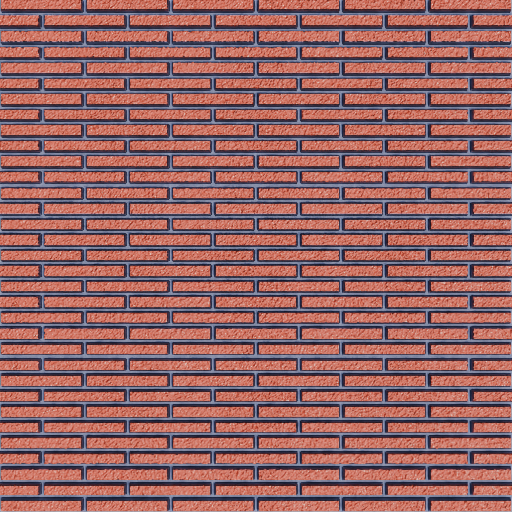}\hfill
    \includegraphics[width=0.26\linewidth]{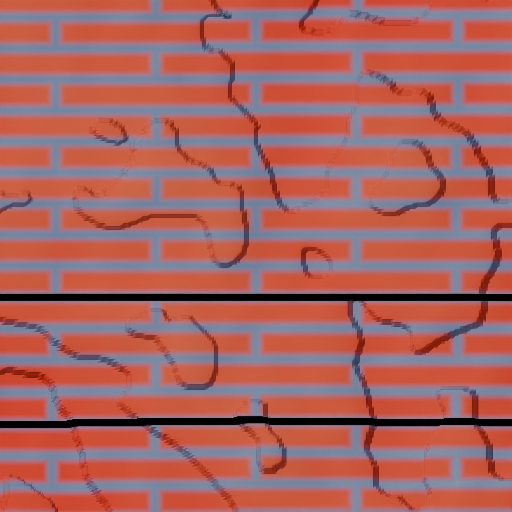}\hfill
    \includegraphics[width=0.26\linewidth]{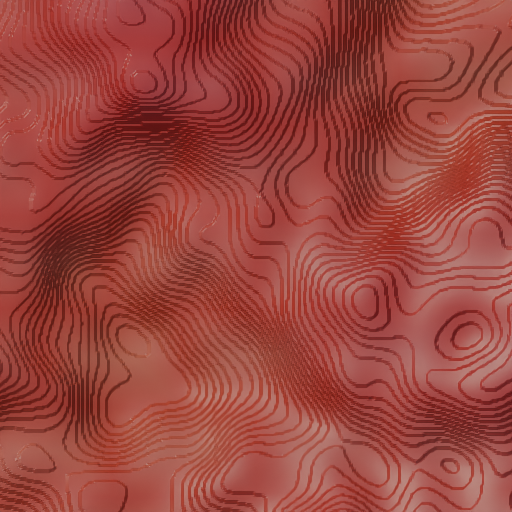}
    \caption{Three illustrative CMA-ES outputs from separate runs for ``a red brick wall'' in the flat layout, ordered by the authors' visual assessment rather than by optimization time. Quick-scorer gains over each run's pre-polish program are $+2.0$, $+3.7$, and $+0.7$, with the largest gain belonging to the middle image. The selected images illustrate scorer disagreement and do not establish a general failure rate.}
    \label{fig:cma}
\end{figure}

An exploratory variant searches continuous numeric literals selected by the engine using CMA-ES~\citep{hansen2016cmaes}, with frequencies in log space, bounded channel weights, and the base layer's alpha and view window excluded.
The recorded configuration uses $\sigma_0=0.25$ and a budget of $500$ evaluations per candidate against the quick scorer.
Figure~\ref{fig:cma} shows results from 3 separate runs for ``a red brick wall'' whose visual ordering differs from their quick-score improvements.
The panels are different outputs ordered by the authors' visual assessment, not successive steps of one optimization trajectory.
This is qualitative evidence of proxy mismatch, not a controlled estimate of how often parameter search fails.
Stored sweeps for the same prompt quantify the asymmetry: at a matched budget of $500$ evaluations, parameter search improved the quick scorer by a mean of $+1.6$ points, against $+0.2$ for seed-only search, and 3 of 13 parameter-search candidates failed to improve their start point.

Seed search preserves all non-seed source within each candidate, including explicit lattice parameters, but noise can still affect masks, color, and height enough to reduce visual fidelity.
Accepting a candidate only when its quick score improves protects that score, not appearance or external evaluation metrics.
The choice of seed-only search is therefore a restriction of the search space rather than a guarantee against degradation.
Independent judgments of the stored before-and-after programs are unavailable, so this comparison does not establish which search space gives better perceived materials.

\section{User Study Protocol and Results}
\label{app:userstudy}

We ran a blind online preference study, delivered through Qualtrics in English and Chinese.
Twenty prompts, five from each benchmark source, were drawn from the $141$-prompt benchmark with a seeded random selection subject to a Jaccard-similarity diversity guard, and the selection was frozen before recruitment.
The stimuli reused the seed-$123$ flat-layout evaluation outputs of the flagship and the three baselines: 80 $512{\times}512$ renders, each recorded in a trial manifest with its checksum.
\method occupied each display position in exactly five of the 20 items, the remaining methods filled the other positions at random, and the layout was fixed per prompt rather than redrawn per participant.
The session opened with informed consent and a background block recording image-creation experience and self-reported normal color vision.
Trial order was randomized per participant in two 10-trial blocks, with an instructed attention check between them.
Each trial showed the four unlabeled renders in a $2{\times}2$ grid and recorded which image best matched the prompt together with a $1$--$7$ prompt-fidelity rating for every image.
Participation was voluntary and began with an informed-consent screen.
The questionnaire did not request names, but participants could optionally provide an email address for follow-up, so responses were confidential rather than anonymous.
The reported analysis excluded the email field and used de-identified responses.

\paragraph{Analysis plan and outcomes.}
The frozen analysis treated the pooled forced-choice share and the mean rating per method as primary outcomes, with uncertainty estimated by a participant-cluster bootstrap ($10{,}000$ resamples), and tested the \method-versus-StableMaterials rating difference by Wilcoxon signed-rank tests at both the prompt and participant levels.
Thirty participants completed the study, all passed the attention check, and none was excluded, leaving $600$ paired decisions with no straight-line response patterns.
Table~\ref{tab:userstudy} reports the outcomes.
\method won $59.2\%$ of forced choices with a mean rating of $4.98$, StableMaterials followed at $19.3\%$ and $3.60$, IntrinsiX at $15.3\%$ and $3.01$, and MatFuse at $6.2\%$ and $2.27$.
The rating advantage over StableMaterials was positive at both test levels ($p{=}4.4\times 10^{-4}$ prompt-level, $p{=}6.5\times 10^{-7}$ participant-mean), and \method was the forced choice on every one of the 20 prompts by at least 4 of the 30 participants.
Ten participants ($33\%$) reported no image-creation experience and 20 ($67\%$) reported some, and \method's win rate was $60.0\%$ in the inexperienced group against StableMaterials' $21.5\%$, and $58.8\%$ in the experienced group against $18.2\%$.
Median per-trial completion time was $36.5$ seconds, and participants gave free-text reasons on $192$ of the $600$ trials.

\begin{table}[htbp]
    \caption{Blind preference study over 30 participants and 20 prompts. Brackets in both columns give $95\%$ participant-cluster bootstrap intervals. The forced-choice column shares sum to $100\%$ over the $600$ trials.}
    \label{tab:userstudy}
    \centering
    \footnotesize
    \setlength{\tabcolsep}{6pt}
    \begin{tabular}{l cc}
        \toprule
        Method          & Share of forced choices (\%)  & Mean fidelity ($1$--$7$)      \\
        \midrule
        \method (ours)  & $\mathbf{59.2}$ $[55.8,62.7]$ & $\mathbf{4.98}$ $[4.73,5.21]$ \\
        StableMaterials & $19.3$ $[16.5,22.3]$          & $3.60$ $[3.31,3.89]$          \\
        IntrinsiX       & $15.3$ $[13.3,17.2]$          & $3.01$ $[2.77,3.26]$          \\
        MatFuse         & $6.2$ $[4.3,8.2]$             & $2.27$ $[2.07,2.49]$          \\
        \bottomrule
    \end{tabular}
\end{table}

\paragraph{Participant and prompt consistency.}
All $30$ participants chose \method more often than any one baseline over the $20$ trials, with $8$--$17$ \method choices per person.
\method had a unique plurality of choices on $13$ of the $20$ prompts and tied for plurality on one more.
Its mean fidelity rating exceeded StableMaterials on $17$ of the $20$ prompts.
These post-hoc summaries describe the fixed study prompts and participants rather than population-wide unanimity.

\newpage
\paragraph{Variation across prompt sources.}
Table~\ref{tab:userstudy-sources} shows the pooled forced-choice shares and the paired mean rating difference between \method and StableMaterials within each prompt source.
Each source contributes only five prompts, so these differences are descriptive and should not be interpreted as source-level significance tests.

\begin{table}[htbp]
    \caption{Descriptive user-study results by prompt source. Each row pools five prompts and $150$ choices from the same $30$ participants. The four method columns give shares of forced choices in percent. Rating gap is the mean paired fidelity rating difference (\method{} minus StableMaterials) on the $1$--$7$ scale.}
    \label{tab:userstudy-sources}
    \centering
    \footnotesize
    \setlength{\tabcolsep}{4pt}
    \begin{tabular}{l r r r r r}
        \toprule
        Prompt source   & \method & StableMaterials & IntrinsiX & MatFuse & Rating gap \\
        \midrule
        GenProc         & $40.7$  & $35.3$          & $12.0$    & $12.0$  & $+0.35$    \\
        MatSynth        & $62.0$  & $8.7$           & $26.7$    & $2.7$   & $+1.95$    \\
        StableMaterials & $57.3$  & $18.7$          & $17.3$    & $6.7$   & $+1.13$    \\
        text2fabric     & $76.7$  & $14.7$          & $5.3$     & $3.3$   & $+2.06$    \\
        \bottomrule
    \end{tabular}
\end{table}

\paragraph{Prompt-sampling sensitivity.}
The intervals in Table~\ref{tab:userstudy} resample participants while holding the selected prompts fixed.
In a post-hoc crossed bootstrap with $10{,}000$ draws, we resampled participants and resampled five prompts within each of the four sources.
The resulting $95\%$ interval for \method's pooled choice share is $[47.3,70.7]\%$, and the interval for its choice-share margin over StableMaterials is $[22.0,56.3]$ percentage points.
This wider interval reflects variation across the selected prompts and does not establish performance outside their four sources.

\paragraph{Second study: the new arms.}
A second blind study with the same instrument and inclusion rule compares the new arms directly in two blocks: 12 three-way trials (DSL versus both Blender-Python arms, drawn from the $29$ prompts where all three delivered) and 8 two-way trials (DSL flagship final versus Material Apprentice's text-only mode, drawn from the $32$ prompts where both delivered a visible render, which conditions in Material Apprentice's favor).
Positions were Latin-square balanced and trial order randomized per participant, with the attention check between blocks.
All $21$ participants completed all trials, passed the attention check, and were included.
In the three-way block the DSL arm receives $65.5\%$ of choices (participant-cluster bootstrap $95\%$ CI $[59.9,70.6]$), against $21.0\%$ ($[15.5,26.6]$) for the unrestrained and $13.5\%$ ($[9.5,17.5]$) for the scaffolded Blender-Python arm, each different from the $1/3$ chance level at $p{\le}0.003$ (Wilcoxon over participant shares).
Mean fidelity ratings are $4.81$ versus $3.49$ and $3.22$, with within-trial rating gaps of $+1.32$ and $+1.58$ ($p{<}10^{-4}$).
In the two-way block the DSL arm receives $83.3\%$ ($[77.4,89.3]$) against $16.7\%$ for Material Apprentice's text-only mode ($p{=}5.2{\times}10^{-5}$ versus the $1/2$ chance level), with ratings $4.94$ versus $3.02$ and a within-trial gap of $+1.92$ ($p{<}10^{-4}$).
As disclosed in Appendix~\ref{app:ma}, Material Apprentice's appearance here is a lower bound on its evaluated image-plus-text mode, and its trials are conditioned on visible renders.

\section{Baseline Configurations}
\label{app:baselines}

Table~\ref{tab:baseline-config} gives the generation settings recorded for $423$ samples per diffusion baseline.
MatFuse evaluation selects its released \texttt{3\_cfg} sampler variant, whose recorded guidance scale is $5.0$.
The methods share Blender scene functions but have different output channels and routing.
StableMaterials height is used for bump shading, whereas \method height drives geometric displacement, and the compared baseline exports do not provide the transmission channel used in the ice example.
Consequently, staged results compare complete systems with these routing choices, rather than isolating the generative representation.
The shared-channel evaluation and the StableMaterials displacement control with fixed, documented calibration that address this attribution appear in Appendix~\ref{app:e2}.
Generation uses each system's released weights: StableMaterials from its published model with the LCM sampler, IntrinsiX as the released FLUX.1-dev LoRA, and MatFuse with its released checkpoint and the \texttt{3\_cfg} sampler.
Every per-run sampler setting is recorded in the baseline records, which also record the repositories, checkpoints, sampler settings, and map conventions.

\begin{table}[htbp]
    \caption{Recorded baseline generation settings at $512{\times}512$ map resolution and three run labels per prompt. Height-map availability does not imply matched rendering: StableMaterials uses bump shading while \method uses geometric displacement.}
    \label{tab:baseline-config}
    \centering
    \footnotesize
    \setlength{\tabcolsep}{5pt}
    \begin{tabular}{l cccc}
        \toprule
        Method          & Backbone           & Steps & Guidance   & Height map \\
        \midrule
        MatFuse         & latent diffusion   & 50    & 5.0        & \ding{55}  \\
        StableMaterials & SD-class LDM + LCM & 4     & 10.0 (LCM) & \ding{51}  \\
        IntrinsiX       & FLUX.1-dev + LoRA  & 28    & 3.5        & \ding{55}  \\
        \method (ours)  & LLM program        & n/a   & n/a        & \ding{51}  \\
        \bottomrule
    \end{tabular}
\end{table}

\section{Image-quality Diagnostic}
\label{app:iqa}

CLIP-IQA measures an image-quality proxy without using the target prompt and is reported separately from alignment metrics.
Table~\ref{tab:clipiqa} includes every main method and both layouts from the stored result snapshots.
StableMaterials has the highest flat-layout mean, while the GLM-based \method configuration has the highest staged mean.
The flagship's alignment gains therefore do not establish uniformly better image quality.
These scores also depend on the scene and routing differences described in Appendix~\ref{app:baselines}.

\begin{table}[htbp]
    \caption{CLIP-IQA image-quality diagnostic, averaged across $141$ prompts and three runs per method, with higher values indicating better predicted image quality. Best mean per layout is in \textbf{bold}. This metric does not measure prompt alignment.}
    \label{tab:clipiqa}
    \centering
    \footnotesize
    \setlength{\tabcolsep}{8pt}
    \begin{tabular}{l cc}
        \toprule
        Method                             & Flat CLIP-IQA  & Staged CLIP-IQA \\
        \midrule
        IntrinsiX                          & 46.51          & 24.08           \\
        MatFuse                            & 19.39          & 10.68           \\
        StableMaterials                    & \textbf{52.17} & 21.27           \\
        \midrule
        \method{} (gemma-4-26b-a4b-it)     & 47.51          & 28.95           \\
        \method{} (qwen3.6-35b-a3b)        & 44.65          & 28.42           \\
        \method{} (deepseek-v4-flash-0731) & 49.68          & 29.00           \\
        \method{} (glm-5.2)                & 49.19          & \textbf{30.12}  \\
        \method{} (gpt-5.6-luna)           & 44.09          & 24.25           \\
        \method{} (gemini-3.6-flash)       & 47.21          & 24.63           \\
        \bottomrule
    \end{tabular}
\end{table}

\section{Result Provenance}
\label{app:provenance}

The retained result snapshots cover all $141$ prompt strings, three run labels, two layouts, and five stored metrics for each of the nine main methods, with no non-finite score values.
The run tree contains $2538$ main program records, $1974$ component and cross-critic records, and $1269$ baseline records, and each retained main trajectory contains an initial program, five revisions, and a polished final program.
These counts establish completeness of the retained result slots, not the number of attempted generations or first-pass parse success, and the records do not retain failed requests, API request identifiers, or token usage.
The benchmark prompt list is reconstructed from the retained raw source pools together with the recorded selection seed.

Every quantitative table in this paper is regenerated from the stored per-run scores.
All $2538$ baseline scored images ($1269$ runs in two layouts) are retained, and our scored renders are regenerated on demand from the retained programs by the rendering pipeline in the released code, under the same engine version and export settings.
The ablation score file stores per-prompt scalars whose programs and settings are linked through the run tree.

\paragraph{Zero judge scores.} The MLLM scoring implementation returns zero after an exception or empty response as well as permitting a valid zero rating.
Exact-zero final judge counts in the flat/staged layouts are IntrinsiX $8/1$, MatFuse $7/2$, StableMaterials $2/1$, Gemma $0/1$, and Qwen $0/1$, with $0/0$ for the remaining four configurations.
A re-scoring audit of all $23$ entries under the identical judge configuration found that six re-score at zero, one re-scores at $1$, and 16 re-score between $2$ and $5$, so most recorded zeros were scoring failures rather than valid judgments.
The tables retain the recorded zeros, and replacing the 17 nonzero re-scores with their values shifts any reported mean by less than $0.04$ points, which changes no comparison.

\section{Per-source Breakdown}
\label{app:persource}

Table~\ref{tab:persource} groups the stored flat-layout scores by benchmark source.
The flagship has higher BLIPScore, CLIPScore, and VQAScore means than StableMaterials in each source.
Its judge mean is lower on the StableMaterials-source and \citet{hu2023generating} prompts, while its largest judge margins come from MatSynth and text2fabric.
These descriptive source differences do not isolate representation effects or establish the cause of a prompt-distribution effect.

\begin{table}[htbp]
    \caption{Flat-layout means by benchmark source for the flagship (\method, final program) and StableMaterials (SM). Best mean per pair is in \textbf{bold}. These descriptive comparisons have no per-source significance claim.}
    \label{tab:persource}
    \centering
    \footnotesize
    \setlength{\tabcolsep}{4.5pt}
    \begin{tabular}{l cc cc cc cc}
        \toprule
                               & \multicolumn{2}{c}{BLIPScore} & \multicolumn{2}{c}{CLIPScore} & \multicolumn{2}{c}{VQAScore} & \multicolumn{2}{c}{Judge}                                                            \\
        \cmidrule(lr){2-3}\cmidrule(lr){4-5}\cmidrule(lr){6-7}\cmidrule(lr){8-9}
        Source ($n$)           & ours                          & SM                            & ours                         & SM                        & ours           & SM    & ours           & SM             \\
        \midrule
        Hu et al. ($30$)       & \textbf{45.12}                & 30.30                         & \textbf{27.53}               & 26.12                     & \textbf{63.50} & 58.39 & 71.43          & \textbf{72.59} \\
        MatSynth ($11$)        & \textbf{73.72}                & 28.46                         & \textbf{30.83}               & 25.05                     & \textbf{59.04} & 42.69 & \textbf{76.30} & 38.82          \\
        StableMaterials ($50$) & \textbf{48.24}                & 34.34                         & \textbf{26.83}               & 25.17                     & \textbf{43.98} & 36.64 & 56.35          & \textbf{59.40} \\
        text2fabric ($50$)     & \textbf{66.55}                & 21.80                         & \textbf{31.09}               & 26.00                     & \textbf{59.21} & 46.85 & \textbf{72.97} & 50.39          \\
        \bottomrule
    \end{tabular}
\end{table}

\section{Prompts}
\label{app:prompts}

The system prompt is the operational specification summarized in Section~\ref{sec:method-prompt}: a role preamble (verbatim below), the DSL spec (program structure, channel ranges, expression vocabulary, the two pitfalls), the playbook of Table~\ref{tab:playbook}'s idioms, six few-shot examples, and output instructions requiring a single \texttt{<material>} block.

\begin{table}[htbp]
    \caption{The organic-texture playbook's 5 idioms, as they appear in the system prompt.}
    \label{tab:playbook}
    \centering
    \footnotesize
    \begin{tabular}{p{4.2cm}p{8.6cm}}
        \toprule
        Idiom                     & Use                                                                                                                                                             \\
        \midrule
        Multi-octave layering     & A low-frequency \dsl{fBm} for broad form plus a high-frequency one for fine grain, reused wherever each scale is needed.                                        \\
        Anisotropic striation     & \dsl{fBm}/\dsl{Worley} with \dsl{base\_freq\_x} $\neq$ \dsl{base\_freq\_y} elongates features along an axis (bark, brushed metal).                              \\
        Coordinate-domain warping & Add a noise to a coordinate term before a \dsl{Sin} pattern to turn banding into turbulent stratification, and drive \dsl{height} from the same warped pattern. \\
        Micro-gloss grain         & A high-frequency, low-amplitude \dsl{fBm} added only to \dsl{roughness} for skin, bark, or stone microstructure.                                                \\
        Substrate-matrix-first    & Continuous substrate in a bottom \dsl{Layer(1)}, discrete features as upper layers whose thresholded masks let the substrate show through.                      \\
        \bottomrule
    \end{tabular}
\end{table}

\paragraph{Role preamble (verbatim).}
\begin{quote}\small\ttfamily\frenchspacing\raggedright
    You are an expert technical artist. You write **layered-material DSL** programs: a compact, declarative language that compiles deterministically into physically based material maps. Given a natural-language description of a material, output a program that reproduces it.
\end{quote}

\paragraph{One few-shot example (of six, verbatim).}
The prompt is ``Weathered red brick wall with pale mortar.''
\begin{lstlisting}[basicstyle=\ttfamily\scriptsize,breaklines=true,columns=fullflexible,keepspaces=true]
View(0, 0, 6, 6)
Define(brick, Bricks(brick_width=1, brick_height=0.45, mortar=0.06, feather=0.005))
Define(tone, fBm(octaves=4, base_freq=2.5, to_01=True, seed=7))
Define(grime, Threshold(fBm(octaves=5, base_freq=1.5, to_01=True, seed=15), below_at=0.55, below_to=0, above_at=0.8, above_to=0.6))
Material(
  Layer(1)
    .basecolor(175, 170, 160)
    .roughness(0.95)
    .height((fBm(octaves=3, base_freq=10, to_01=True, seed=2) * 0.02)),
  Layer(brick)
    .basecolor((150 + (tone * 50)), (55 + (tone * 25)), (45 + (tone * 15)))
    .roughness((0.7 + (tone * 0.2)))
    .height((0.04 + (tone * 0.05))),
  Layer(grime)
    .basecolor(60, 55, 50)
    .roughness(1)
)
\end{lstlisting}

\paragraph{Critic output schema (verbatim field descriptions).}
The critic must return JSON with \texttt{match\_score} (``0-100, how well the image matches the description''), \texttt{differences} (``Concrete visual mismatches between image and description''), and \texttt{suggestions} (``Concrete appearance changes that would improve the match'').

\paragraph{Revision turn (verbatim).}
\begin{quote}\small\ttfamily\frenchspacing\raggedright
    A critic reviewed your material against the original description and scored the match XX/100. Their notes: \{differences + suggestions, bulleted\}. Revise the material to address these and reply with ONLY the updated <material>...</material>. When keeping a layer unchanged, copy it exactly, including any seed= values, so its noise stays the same.
\end{quote}

\section{Qualitative Figure Provenance}
\label{app:qualitative}

Figure~\ref{fig:comparison} uses selected final outputs from the flagship and baseline runs, recorded in the paper-figure export configuration.
The configuration selects the prompts ``holiday wrapping paper,'' ``Acoustic panels foam wedges checker tiles,'' and ``Cracked ice.''
The displayed row labels shorten these prompts, and each flat/staged pair renders the same selected material output.
This is a deliberate selection to expose pattern, relief, and transmission differences, not a random sample or a comparison with matched output-channel routing.

\section{Shared-Backbone Blender-Python Baseline}
\label{app:bpy}
This control holds the generation loop fixed and varies only the representation, asking the same backbone to author the same materials either as \method programs or as native Blender-Python scripts.
All numbers in this section use \texttt{gemini-3.6-flash} at seed $42$, because the second frozen backbone had not completed at scoring freeze.
This single-backbone, single-seed scope over $45$ prompts is a limitation of the control, so its margins should not be read as holding across backbones, seeds, or larger prompt sets.

\paragraph{Frozen protocol.}
The protocol and its reference document were frozen before any test-set generation call, at commit \texttt{59e3b09} with reference-doc sha256 \texttt{1a70019751087009215c285a54f60a8ade6f58f7c3c5d47a5fef9271779ff038}.
The reference document has $24{,}178$ characters and therefore meets or exceeds the DSL generator's system prompt in length, so the comparison does not favor the DSL through documentation size.
The test set contains $45$ prompts stratified from the $141$-prompt benchmark by one shared seeded draw ($10$ category, $4$ MatSynth, $16$ StableMaterials, $15$ text2fabric), and $5$ disjoint pilot prompts are excluded from every reported number by protocol.
Both arms fix temperature at $0.7$, reasoning effort high, a $3$-retry parse budget, $5$ critique--revision rounds, trajectory selection, and seed $42$.
Seed search is off on both sides, so the representation is not confounded with search budget.
The critic receives the render and the source code on both sides, with the DSL-specific statistics panel off on both sides, and the Blender-Python critic image is that arm's own plane swatch while ours is the CPU engine preview.
The scaffolded arm writes only the body of a \texttt{build\_material} function that constructs exactly one node-graph material on a harness-owned scene, executed by Blender $5.1.2$ Cycles in a fresh subprocess per attempt with a $120$-second timeout.
The unrestrained arm removes the scaffold and asks for a complete script including scene setup, frozen to the first $30$ prompts in dataset order to bound cost.
Render layouts are byte-identical to the main benchmark's flat swatch and studio room.

\paragraph{Scoring and status.}
Local metrics score the chosen program's final plane and scene pair, both conditional on delivering that pair and unconditional with failures scored $0$: BLIPScore, CLIPScore, and VQAScore appear in Table~\ref{tab:bpyquality}, and CLIP-IQA was scored but is not tabulated.
Table~\ref{tab:bpy} summarizes execution, cost, and program size per arm, and Table~\ref{tab:bpyquality} the quality metrics.
Judge scores come from a single scoring session covering all three arms, so every judge cell and judge delta in this section compares like with like, although the batch's absolute level is not comparable with the main evaluation's judge cells (Appendix~\ref{app:budget}).

\begin{table}[htbp]
    \caption{Direct Blender-Python generation under the frozen shared-backbone protocol (\texttt{gemini-3.6-flash}, seed $42$, $45$ stratified prompts, identical repair budget, critique rounds, selection, and rendering layouts, seed search off both sides). The unrestrained arm covers its frozen $30$-prompt scope. Valid renders counts runs with a final scored render pair, flat BLIP and flat judge are conditional on success, and first-attempt validity appears in Table~\ref{tab:bpyfunnel}: $82.2\%$ for initial DSL generations over the same $45$ prompts ($37$ of $45$, Wilson $[68.7,90.7]$, one sample per prompt under the main-run settings without repair) against $80\%$ and $57\%$ for the Blender-Python arms. USD per success is metered token usage at the documented list price for the Blender-Python arms, and the daggered DSL entry is a transfer estimate from the scaffolded arm's measured per-role token rates. Judge cells were scored in a single session shared by all three arms.}
    \label{tab:bpy}
    \centering
    \footnotesize
    \setlength{\tabcolsep}{5pt}
    \begin{tabular}{l ccccc}
        \toprule
         & Valid & Flat BLIP & Flat judge & USD per & Mean program \\
         & renders & (cond.) & (cond.) & success & bytes \\
        \midrule
        Ours-DSL (matched) & 45/45 & 47.6 & 76.5 & \$0.08$^{\dagger}$ & 2,559 \\
        Blender-Python (scaffolded) & 43/45 & 28.8 & 47.1 & \$0.09 & 3,541 \\
        Blender-Python (unrestrained) & 30/30 & 15.6 & 40.4 & \$0.06 & 2,729 \\
        \bottomrule
    \end{tabular}
\end{table}

\begin{table}[htbp]
    \caption{Per-attempt failure ladder over all generation attempts (excluding each run's final re-execution pass), with the exception taxonomy as a percentage of failed attempts. R5 and the recovery rows are per run. A \method program that parses executes by construction, so the DSL row reports run-level completion only.}
    \label{tab:bpyfunnel}
    \centering
    \footnotesize
    \setlength{\tabcolsep}{4pt}
    \resizebox{\linewidth}{!}{%
    \begin{tabular}{l cc}
        \toprule
         & Blender-Python scaffolded & Blender-Python unrestrained \\
        \midrule
        Generation attempts & 297 & 210 \\
        \midrule
        R0 tag-extract & 98.7\% & 96.7\% \\
        R1 ast-parse & 98.0\% & 96.7\% \\
        R2 execute & 93.3\% & 86.2\% \\
        R3 graph-valid & 93.3\% & 86.2\% \\
        R4 non-blank & 87.2\% & 85.7\% \\
        R5 scored (per run) & 43/45 & 30/30 \\
        Ours-DSL scored (run level) & \multicolumn{2}{l}{$45/45$ attempted runs (a parsed program executes by construction)} \\
        \midrule
        First attempt valid & 80\% (36/45) & 57\% (17/30) \\
        Repair recovery & 89\% (8/9) & 100\% (13/13) \\
        \midrule
        Taxonomy (\% of failed, scaffolded) & \multicolumn{2}{l}{blank 47\%, socket-KeyError 24\%, other 13\%, tag 11\%, ast 5\%} \\
        Taxonomy (\% of failed, unrestrained) & \multicolumn{2}{l}{NameError 43\%, tag 23\%, socket-KeyError 20\%, other 7\%, blank 3\%, AttributeError 3\%} \\
        \bottomrule
    \end{tabular}}
\end{table}

\begin{table}[htbp]
    \caption{Final-render quality of the chosen program per arm in both layouts, conditional on success and unconditional with failures scored $0$ over the arm's prompt scope. The $\Delta$ columns are paired per-prompt differences (ours minus Blender-Python scaffolded, two-sided Wilcoxon over shared prompts with $n{=}43$ conditional and $n{=}45$ unconditional) with $95\%$ percentile bootstrap intervals ($^{**}p{<}0.01$, $^{***}p{<}0.001$), and fifteen of the sixteen $\Delta$ cells are significant at $p{<}0.001$ with the conditional flat BLIP cell at $p{=}0.001$. Judge rows were scored in a single scoring session covering every arm, so all cells and deltas compare within a single batch. The unrestrained arm's staged column is the same HDRI render procedure as its flat column (disclosed below), so its staged cells are not comparable to the studio-room staged cells.}
    \label{tab:bpyquality}
    \centering
    \footnotesize
    \setlength{\tabcolsep}{3pt}
    \resizebox{\linewidth}{!}{%
    \begin{tabular}{l cc cc cc cc}
        \toprule
         & \multicolumn{2}{c}{Ours-DSL} & \multicolumn{2}{c}{BP scaffolded} & \multicolumn{2}{c}{BP unrestrained} & \multicolumn{2}{c}{$\Delta$ ours$-$BP scaff.} \\
        Metric & cond. & uncond. & cond. & uncond. & cond. & uncond. & cond. & uncond. \\
        \midrule
        Flat BLIP & 47.6 & 47.6 & 28.8 & 27.5 & 15.6 & 15.6 & +19.7$^{**}$ [+8.1,+31.1] & +20.1$^{***}$ [+8.9,+31.0] \\
        Flat CLIP & 29.4 & 29.4 & 24.7 & 23.6 & 22.5 & 22.5 & +4.9$^{***}$ [+3.4,+6.5] & +5.8$^{***}$ [+3.9,+7.9] \\
        Flat VQA & 54.0 & 54.0 & 42.1 & 40.3 & 35.7 & 35.7 & +12.6$^{***}$ [+6.8,+18.5] & +13.8$^{***}$ [+7.9,+19.7] \\
        Flat Judge & 76.5 & 76.5 & 47.1 & 45.0 & 40.4 & 40.4 & +29.3$^{***}$ [+20.6,+38.3] & +31.4$^{***}$ [+22.6,+40.6] \\
        Staged BLIP & 27.6 & 27.6 & 7.0 & 6.7 & 15.7 & 15.7 & +21.4$^{***}$ [+12.6,+31.2] & +21.0$^{***}$ [+12.6,+30.1] \\
        Staged CLIP & 25.0 & 25.0 & 20.8 & 19.8 & 22.5 & 22.5 & +4.5$^{***}$ [+2.9,+6.2] & +5.1$^{***}$ [+3.4,+7.0] \\
        Staged VQA & 52.0 & 52.0 & 35.6 & 34.0 & 35.8 & 35.8 & +16.9$^{***}$ [+12.0,+22.1] & +18.0$^{***}$ [+13.0,+23.1] \\
        Staged Judge & 61.3 & 61.3 & 30.7 & 29.4 & 42.6 & 42.6 & +30.0$^{***}$ [+21.7,+38.7] & +31.9$^{***}$ [+23.5,+40.3] \\
        \bottomrule
    \end{tabular}}
\end{table}

\paragraph{Per-attempt fragility.}
Table~\ref{tab:bpyfunnel} reports the per-attempt ladder.
The scaffolded arm fails $12.8\%$ of its $297$ generation attempts and the unrestrained arm $14.3\%$ of $210$, with first-attempt validity of $80\%$ and $57\%$, against $82.2\%$ for initial DSL generations measured the same way over the same prompts ($37$ of $45$).
The repair loop absorbs most of this fragility at the run level, recovering $8$ of $9$ and $13$ of $13$ runs whose first attempt failed, so $43$ of $45$ scaffolded and $30$ of $30$ unrestrained runs still deliver a final scored pair.
The failures concentrate in the native API: among failed scaffolded attempts, $47\%$ are blank renders and $24\%$ are socket-name KeyErrors, while the unrestrained arm adds undefined-name errors at $43\%$ of its failures.
This is consistent with the roughly $30\%$ first-shot direct-graph failure rate that Material Apprentice reports~\citep{gupta2026apprentice}.
A post-hoc visual-content diagnostic, computed after unblinding from the same final renders the metrics scored and not part of the frozen ladder, locates the remaining gap: a plane-render pixel standard deviation of at least $10$ passes for $89\%$ of DSL finals ($40$ of $45$) but only $60\%$ of scaffolded ($26$ of $43$) and $67\%$ of unrestrained ($20$ of $30$) finals, with medians of $28.1$ versus $11.9$ and $15.3$.
Many nominally successful Blender-Python materials are therefore blank-adjacent flat panels that clear the frozen non-blank rung (standard deviation above $1$) while carrying little visible texture.

\paragraph{Quality, size, and latency.}
Table~\ref{tab:bpyquality} reports the chosen programs' final-render means.
Conditional flat BLIPScore is $47.6$ for the matched DSL arm against $28.8$ scaffolded and $15.6$ unrestrained, the paired delta against the scaffolded arm is $+19.7$ ($95\%$ CI $[+8.1,+31.1]$), the staged layout repeats it at $+21.4$ $[+12.6,+31.2]$, and all sixteen paired cells favor the DSL, fifteen at $p{<}0.001$ and the conditional flat BLIPScore cell at $p{=}0.001$.
That single scoring session, shared by all three arms, agrees with the local metrics at a larger margin: the conditional flat judge delta is $+29.3$ $[+20.6,+38.3]$ and the staged one $+30.0$ $[+21.7,+38.7]$, both at $p{<}0.001$.
Blender-Python executes well at the run level, but the gap is quality per success, program size, and preview latency.
Chosen scripts average $88.4$ lines against $31.3$ for DSL programs, a $2.8\times$ ratio, with mean bytes closer at $3{,}541$ against $2{,}559$ because DSL masks serialize verbosely.
Each Blender-Python attempt costs about $12.5$ seconds of cold-subprocess wall, $7.6$ of them rendering, against roughly $70$ milliseconds for the standalone CPU preview.
Per-success API cost is comparable across arms at these matched polish-off settings (\$0.09 scaffolded, \$0.06 unrestrained, and about \$0.08 estimated for ours), so the headline cost asymmetry of Section~\ref{sec:results-stages} comes from the seed-search budget rather than the loop.

\paragraph{Deviations and disclosures.}
The ours-matched arm covers all $45$ prompts after the missing fill re-run completed, and because that driver's resume check looked for records at the wrong path it re-ran earlier prompts as duplicate slates, with the loader keeping the latest complete slate per prompt so exactly one slate per prompt enters every number.
Its USD entry is a transfer estimate that reuses the scaffolded arm's measured per-role token means, which biases the comparison against the DSL because Blender scripts are longer than DSL programs.
USD converts measured tokens at the documented list price, and wall-clock seconds are proxy-load confounded, so they are indicative only.
The ours-matched finals come from re-rendering the chosen programs through the shipped export pipeline, the same procedure as the flagship evaluation.
A harness filename collision overwrote the unrestrained arm's plane pass with its scene pass, repaired by a standard-deviation-verified rename plus one fresh re-execution, so that arm's flat and staged entries are two executions of the same HDRI render procedure over the model's own scene rather than swatch versus studio room.
Its staged cells are therefore not comparable with the studio-room staged cells of the other two arms.
The two scaffolded runs that never produced a final pair are retained in the record, score $0$ in the unconditional means, and are excluded from the conditional ones.

\section{Editing Benchmark}
\label{app:editing}
This suite measures whether authored programs support targeted post-generation edits at scale, extending the single-material demonstration of Section~\ref{sec:manual-edits}.
The specification, including the material cohort, the task list, the per-type attainment tolerances, and the per-task allowed-change masks, was fixed before any editing arm ran.

\paragraph{Cohort and tasks.}
Forty materials are drawn from the recorded seed-$42$ main runs of \texttt{gemini-3.6-flash} and \texttt{gpt-5.6-luna} ($20$ each and $10$ per benchmark source) by a seeded selection under a structural gate (at least two layers, a shared \dsl{Define} used at least twice, and an explicit noise seed) and a quality gate (final flat BLIPScore at or above the source-stratum median).
Each material receives the edit types whose ground truth is achievable on it, giving $172$ tasks over five families: roughness shift ($n{=}25$), base-color shift ($n{=}38$), widening a shared mask feature ($n{=}31$), noise-seed swap ($n{=}40$), and layer removal ($n{=}38$).
Every task carries a scripted ground-truth edit, and all $172$ ground-truth edits were verified to pass their own tolerances before any arm ran.
The manually edited tile material of Section~\ref{sec:manual-edits} is not part of this cohort, because the frozen pool admits only gate-passing seed-$42$ main runs, so the illustrative case and this benchmark share no material.

\paragraph{Success criteria.}
A task succeeds when the edited program attains the requested change and preserves everything else, with the denominator over all attempts.
Attainment is checked against the frozen per-type tolerances: a mean roughness change of at least $0.1$, a color-direction cosine of at least $0.8$ with a channel-mean magnitude of at least $16/255$, a mask-coverage change of at least $0.01$ with the pattern period held within $5\%$, a changed noise-seed multiset with non-seed source bytes equal, and a layer count equal to the ground truth with bounded residual coverage and height.
Preservation requires every channel outside the task's allowed-change mask to change by at most $1$ of $255$ levels on 8-bit channels and at most $1\%$ of the original maximum on HDR channels.
For seed swaps, attainment holds by construction once the seed multiset changes and the non-seed source is byte-equal, so that family's failures are preservation failures by necessity.
The editor is \texttt{gpt-5.6-luna} with at most $6$ LLM calls per task, $2.6$ on average.
These criteria are channel-space checks, and no judge or perceptual metric enters the score, so they measure edit exactness rather than perceived edit quality.

\begin{table}[htbp]
    \caption{Editing results under the frozen $172$-task specification. Joint requires attainment and preservation on a task, denominators count all attempts, and Att./Pres.\ are the marginal attainment and preservation rates (preservation is evaluated only after attainment passes, so Pres.\ equals Joint in every row). The DSL arm uses one editor run per task, and each raster arm takes the best of its three diffusion seeds per task, with any missing raster run counted as a failure. The joint and attainment cells of the raster arms' seed-swap and layer-removal rows carry $^{\dagger}$ because their attainment criteria are program-level properties that a regenerated raster set cannot express. The expressible-families row restricts the denominator to the three families a raster arm can express (roughness, base color, widen, $n{=}94$). The collateral row gives the mean absolute change on unedited channels over all attempts, in $1/255$ levels.}
    \label{tab:editing}
    \centering
    \footnotesize
    \setlength{\tabcolsep}{6pt}
    {(a) Joint success rate (\%), attainment and preservation on the same task.}\\[2pt]
    \begin{tabular}{l ccc}
        \toprule
        Edit type & DSL edit & MatFuse re-prompt & \multicolumn{1}{>{\centering\arraybackslash}p{2.2cm}}{StableMaterials re-prompt} \\
        \midrule
        Roughness ($n{=}25$) & 80.0 & 0.0 & 0.0 \\
        Base color ($n{=}38$) & 84.2 & 0.0 & 0.0 \\
        Widen ($n{=}31$) & 80.6 & 54.8 & 22.6 \\
        Seed swap ($n{=}40$) & 77.5 & 0.0$^{\dagger}$ & 0.0$^{\dagger}$ \\
        Layer removal ($n{=}38$) & 97.4 & 0.0$^{\dagger}$ & 0.0$^{\dagger}$ \\
        \midrule
        All ($n{=}172$) & 84.3 & 9.9 & 4.1 \\
        Expressible families only ($n{=}94$) & 81.9 & 18.1 & 7.4 \\
        \midrule
        Collateral change, all attempts ($1/255$ levels) & 0.08 & 34.8 & 68.1 \\
        \bottomrule
    \end{tabular}\\[6pt]
    {(b) Attainment / preservation rates (\%), preservation evaluated after attainment.}\\[2pt]
    \begin{tabular}{l ccc}
        \toprule
        Edit type & DSL edit & MatFuse re-prompt & StableMaterials re-prompt \\
        \midrule
        Roughness ($n{=}25$) & 84.0 / 80.0 & 84.0 / 0.0 & 76.0 / 0.0 \\
        Base color ($n{=}38$) & 84.2 / 84.2 & 57.9 / 0.0 & 78.9 / 0.0 \\
        Widen ($n{=}31$) & 83.9 / 80.6 & 61.3 / 54.8 & 61.3 / 22.6 \\
        Seed swap ($n{=}40$) & 100.0 / 77.5 & $0.0^{\dagger}$ / 0.0 & $0.0^{\dagger}$ / 0.0 \\
        Layer removal ($n{=}38$) & 100.0 / 97.4 & $0.0^{\dagger}$ / 0.0 & $0.0^{\dagger}$ / 0.0 \\
        \midrule
        All ($n{=}172$) & 91.3 / 84.3 & 36.0 / 9.9 & 39.5 / 4.1 \\
        Expressible families only ($n{=}94$) & 84.0 / 81.9 & 66.0 / 18.1 & 72.3 / 7.4 \\
        \bottomrule
    \end{tabular}
\end{table}

\paragraph{DSL-arm results.}
Table~\ref{tab:editing} reports the frozen DSL arm.
Overall joint success is $84.3\%$ ($95\%$ Wilson interval $[78.1,89.0]$) with attainment at $91.3\%$, all $172$ edited programs parsed and exported, and the $27$ failures split into $15$ attainment and $12$ preservation failures.
Every one of the $145$ joint successes preserves its frozen channels byte-exactly, so the numeric preservation tolerance was never needed within success.
Preservation failures concentrate on seed swaps, which contribute $9$ of the $12$ and fail when the model edits more than the seed.
Collateral change on unedited channels across all attempts averages $0.08$ of a $1/255$ level, with seed swaps highest at $0.25$.

\paragraph{Raster re-prompting arms.}
The MatFuse and StableMaterials arms receive the same edit instruction as text and regenerate the full map set, because a raster generator has no program to edit, and every raster output is converted to the same channel export format and scored by the same frozen evaluator.
Each raster arm runs three diffusion seeds per task and keeps the best of the three (a task's joint succeeds if any seed attains and preserves), a selection rule the DSL arm does not receive, and any missing raster run counts as a failure in the denominator.
The seed-swap and layer-removal attainment checks are program-level properties, a changed noise-seed multiset with all other source bytes equal and a reduced layer count, which a regenerated raster set cannot express, so those joint and attainment cells read $0$ for the raster arms and are daggered in Table~\ref{tab:editing} as inapplicable by construction.
Overall joint success is $9.9\%$ for MatFuse and $4.1\%$ for StableMaterials: both often attain the requested color or roughness change (all-task attainment $36.0$ and $39.5$) but the regenerated maps move every other channel as well, failing preservation, so joint success collapses to the widen family, where a full regeneration can pass both checks ($54.8$ and $22.6$).
Restricted to the three families a raster arm can express (roughness, base color, widen, $n{=}94$), joint success is $18.1\%$ for MatFuse and $7.4\%$ for StableMaterials against $81.9\%$ for the DSL arm, so the aggregate gap is not an artifact of the two program-level families.
Collateral change on unedited channels averages $34.8$ and $68.1$ levels of $255$ against $0.08$ for the DSL arm.
A single-seed variant (seed $42$ only) gives the same joint rates ($9.9$ and $4.1$) with slightly lower attainment ($35.5$ and $37.2$), so the best-of-three rule moves attainment, not joint success.
\paragraph{Blender-Python editing comparison.}
The same five edit types were frozen over the $40$ matched Blender-Python materials of Appendix~\ref{app:bpy} ($199$ tasks, instruction templates reused verbatim), and each arm edited its own programs under the matched backbone: the DSL arm through the same edit machinery as above, the Blender-Python arm by re-prompting its scaffolded script and re-executing it in the frozen sandbox.
Because the Python arm exposes no program-level ground truth, both arms are judged by one comparative protocol: a temperature-zero MLLM sees the before and after renders and answers whether the requested change was applied and whether everything else is unchanged, with joint success requiring execution, the applied change, and preservation.
This visual yardstick is stricter than the map-level checks above, so its levels are not comparable to Table~\ref{tab:editing}.
The DSL succeeds jointly on $54.8\%$ of tasks against $44.7\%$ for Blender-Python ($48$ versus $28$ discordant successes, sign test $p{=}0.057$), executes its edits more often ($98.5\%$ versus $92.0\%$), and the gap concentrates on the parameter edit ($45.0\%$ versus $17.5\%$ on widen), with seed swap slightly favoring Blender-Python ($47.5\%$ versus $52.5\%$) and preservation at parity ($94.9\%$ versus $94.5\%$).
Pixel-level change is smaller for Blender-Python edits (mean absolute difference $10.2$ versus $15.8$ of $255$), consistent with more conservative edits that less often deliver the requested change.

\section{Shared-Channel and Height-Routing Controls}
\label{app:e2}

The main comparison bundles two distinct confounds.
Channel richness is a property of a representation: which channels a system emits at all.
Height routing is an experimental choice: whether an emitted height map drives geometric displacement or bump shading.
This control separates the two by rerendering the stored outputs of all four systems under equalized channels and under crossed height routing.

\paragraph{Protocol.}
Four conditions are compared over the same $141$ prompts, seeds $42$, $123$, and $2026$, and both layouts, using the \texttt{gemini-3.6-flash} flagship to match the main significance family.
C0 (native) reuses each method's committed per-seed scores and recorded routing with no new renders: \method uses all channels with height as displacement, StableMaterials and IntrinsiX add metallic with StableMaterials height as bump at distance $0.1$, and MatFuse adds specular with no height map.
C1 (shared) rerenders every method with only the strict channel intersection that all systems emit, namely base color, roughness, and normal: StableMaterials and IntrinsiX drop metallic, MatFuse drops specular, and \method additionally drops height, transmission, metallic, ambient occlusion, IOR, and emissive.
C2 rerenders \method with its native channels but height routed as bump (per-map min-max normalization, bump distance $0.1$), mirroring StableMaterials' normalized map at its default strength.
C3 rerenders StableMaterials with its native channels but height routed as real displacement (midlevel $0.5$, scale $0.1$, its own bump amplitude).
A calibration probe measured the flagship's native relief, the peak-to-trough displacement Blender applies, at a median of $0.0167$ world units over $24$ stratified finals ($0.0170$ over all $423$), and this measured median was frozen as the fallback displacement scale, never needed because pilot checks at scale $0.1$ produced coherent silhouettes without tears or spikes.
Every new render goes through the evaluation pipeline's own renderer (\texttt{matloom.render}, Blender Cycles), so channel routing is the only variable, all maps are referenced in place from the recorded runs with no resampling or new generation, and ours are the same final programs scored in Table~\ref{tab:main}.
The control adds $5{,}076$ renders: $3{,}384$ for C1 ($4$ methods), $846$ for C2, and $846$ for C3, each $141$ prompts $\times$ $3$ seeds $\times$ $2$ layouts.

\begin{table}[htbp]
    \caption{Shared-channel comparison (C1): every method rerendered with only base color, roughness, and normal, through one renderer path over the stored maps with no resampling, mean over $141$ prompts and seeds $42$, $123$, and $2026$. Ours is the \texttt{gemini-3.6-flash} flagship's final programs, the same outputs as Table~\ref{tab:main}. Best value per column in \textbf{bold}.}
    \label{tab:e2-shared}
    \centering
    \footnotesize
    \setlength{\tabcolsep}{4.5pt}
        \begin{tabular}{l cccc cccc}
            \toprule
             & \multicolumn{4}{c}{Flat} & \multicolumn{4}{c}{Staged} \\
            \cmidrule(lr){2-5}\cmidrule(lr){6-9}
            Method & BLIP & CLIP & VQA & Judge & BLIP & CLIP & VQA & Judge \\
            \midrule
            IntrinsiX & 28.80 & 24.06 & 43.09 & 38.13 & 19.40 & 21.36 & 41.25 & 31.55 \\
            MatFuse & 8.23 & 20.09 & 30.57 & 19.66 & 8.61 & 19.19 & 30.84 & 17.41 \\
            StableMaterials & 27.77 & 25.36 & 44.96 & 43.82 & 20.07 & 23.34 & 44.45 & 37.92 \\
            Ours (gemini-3.6-flash) & \textbf{50.55} & \textbf{27.74} & \textbf{51.77} & \textbf{66.73} & \textbf{35.54} & \textbf{25.60} & \textbf{52.64} & \textbf{58.35} \\
            \bottomrule
        \end{tabular}
\end{table}

All twenty-four flagship-versus-baseline contrasts under this condition (three baselines by four metrics by two layouts) remain significant after Holm correction ($p{<}0.001$ in every cell), for example flat judge $+22.9$ ($95\%$ CI $[+19,+27]$) and staged judge $+20.4$ ($[+16,+25]$) against StableMaterials.
Render noise bounds every delta in this section: re-rendering $24$ native-routing cells per layout through the shared path and rescoring gives mean absolute changes of $0.29$--$0.30$ CLIPScore and $0.49$--$0.75$ BLIPScore with no per-method bias, so deltas below about $0.5$ CLIP or $1$ BLIP points are read as not resolvable above render noise.
The band was measured on CLIPScore and BLIPScore only, and the same-batch judge deltas reported below, $0.4$ to $6.1$ points, are read with the same caution.

\begin{table}[htbp]
    \caption{Native-minus-shared deltas per method, so positive cells are points carried by the channels the intersection removed. StableMaterials and IntrinsiX lose metallic, MatFuse loses specular, and \method loses height, transmission, metallic, ambient occlusion, IOR, and emissive ($n{=}141$ prompts). Local-metric cells compare the committed native scores with the shared rerenders, and judge cells are computed within a single scoring session that scored the shared rerenders and a native re-judge of every method, because judge scores from different sessions are not level-comparable (see the decomposition).}
    \label{tab:e2-delta}
    \centering
    \footnotesize
    \setlength{\tabcolsep}{4.5pt}
        \begin{tabular}{l cccc cccc}
            \toprule
             & \multicolumn{4}{c}{Flat} & \multicolumn{4}{c}{Staged} \\
            \cmidrule(lr){2-5}\cmidrule(lr){6-9}
            Method & BLIP & CLIP & VQA & Judge & BLIP & CLIP & VQA & Judge \\
            \midrule
            IntrinsiX & +1.1 & +0.2 & +0.5 & +0.9 & +1.6 & +0.4 & +0.7 & +0.5 \\
            MatFuse & +0.0 & +0.0 & +0.0 & +1.9 & -0.0 & +0.1 & -0.3 & -0.1 \\
            StableMaterials & +0.8 & +0.3 & +0.4 & +0.4 & +1.3 & +0.3 & +1.1 & +2.1 \\
            Ours (gemini-3.6-flash) & +5.5 & +1.1 & +2.9 & +6.1 & +0.6 & -0.3 & +1.6 & +3.2 \\
            \bottomrule
        \end{tabular}
\end{table}

\paragraph{Decomposition.}
The two controls answer different questions.
C1 measures channel richness, a representation property: removing the channels a system alone emits moves the baselines' judge means by at most $2.1$ points (IntrinsiX $+0.9$ flat and $+0.5$ staged, both n.s., MatFuse $+1.9$ flat at $p{=}0.07$ and $-0.1$ staged, StableMaterials $+0.4$ flat and $+2.1$ staged at $p{=}0.006$), while it costs \method $6.1$ flat judge points ($p{=}7.3\times 10^{-5}$) and $3.2$ staged ($p{=}0.057$) on top of its $5.5$ flat BLIPScore loss, the visible contribution of its height relief and transmission channels.
Under this strict intersection the flagship still leads every metric-layout cell against every baseline (Table~\ref{tab:e2-shared}), so the lead is not created by the extra channels even though those channels contribute about $6$ flat judge points.
The baselines' CLIP and BLIP deltas lie at or near the noise band and are not interpreted.
C2 and C3 measure height routing, an experimental choice made at render time.
Every judge comparison in this paper is computed within a single scoring session: judge scores recorded in different sessions are not level-comparable, so a judge delta that spans two sessions would reflect the session offset rather than the systems.

\begin{table}[htbp]
    \caption{Height-routing $2{\times}2$: the flagship and StableMaterials rerendered with their own native channels under both height routings (new renders, C2 and C3), next to their native rows (C0, committed scores, no new renders). Mean over $141$ prompts and seeds $42$, $123$, and $2026$. The re-routed rows were judge-scored in one session while the native rows' judge cells are the original evaluation's scores, so judge deltas between routings are read within that session, using its fresh native scores, in the text.}
    \label{tab:e2-routing}
    \centering
    \footnotesize
    \setlength{\tabcolsep}{4pt}
        \begin{tabular}{l cccc cccc}
            \toprule
             & \multicolumn{4}{c}{Flat} & \multicolumn{4}{c}{Staged} \\
            \cmidrule(lr){2-5}\cmidrule(lr){6-9}
            Condition & BLIP & CLIP & VQA & Judge & BLIP & CLIP & VQA & Judge \\
            \midrule
            Ours, displacement (native, C0) & 56.06 & 28.80 & 54.71 & 67.01 & 36.14 & 25.30 & 54.26 & 57.34 \\
            Ours, bump (C2) & 48.60 & 27.54 & 53.01 & 64.65 & 34.98 & 24.60 & 53.64 & 56.42 \\
            StableMaterials, bump (native, C0) & 28.57 & 25.66 & 45.36 & 57.41 & 21.33 & 23.59 & 45.53 & 54.29 \\
            StableMaterials, displacement (C3) & 30.79 & 25.84 & 46.31 & 45.49 & 21.21 & 23.59 & 46.78 & 39.87 \\
            \bottomrule
        \end{tabular}
\end{table}

\method with height as bump still leads every metric-layout cell against every baseline in either routing, and within a single scoring session its bump row exceeds StableMaterials' displacement row by $+19.2$ flat and $+16.6$ staged judge points.
Switching \method from displacement to bump costs it $7.5$ flat BLIPScore and, read within that same session, $8.2$ flat and $5.1$ staged judge points, while switching StableMaterials from bump to displacement gains it $2.2$ flat BLIPScore and moves its judge mean by about one point.
Height routing is thus second-order relative to the representation gap, and no routing swap changes the ordering on any cell.
We do not interpret StableMaterials' displacement cells mechanistically, because real displacement at scale $0.1$ is not its released configuration.

\paragraph{Scope.}
The control covers the flagship only, not all six backbones.
The intersection removes one channel from each baseline and six from \method, so it is conservative against \method on channel count and against the baselines on their metallic and specular contributions.
C0 reuses committed native scores while the new conditions rerender through the shared path, so the C1 deltas also absorb the renderer-path switch, which the noise band above bounds on the two rescored metrics.
Per-condition scores and the render driver are released with the artifacts.

\section{Budget, Best-of-N, and Cost Controls}
\label{app:budget}

\begin{figure}[htbp]
    \centering
    \includegraphics[width=\linewidth]{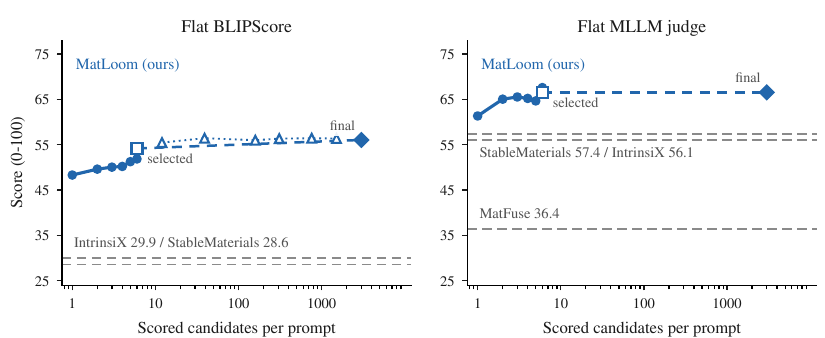}
    \caption{Flat-layout quality versus scored-candidate budget for the \texttt{gemini-3.6-flash} flagship, mean over per-prompt seed means ($141$ prompts $\times$ $3$ runs). All values are final external metrics, never the internal quick scorer. Circles mark rounds $0$--$5$, the open square the selected program, the open diamond the final polished output, and open triangles truncated seed-sweep replays at $K{=}1$ to $K{=}500$ variants per candidate, re-rendered and re-scored on final metrics. Triangles appear in the BLIPScore panel only: the replay points were also judge-scored, but in a later scoring session whose level is not comparable with this figure's other points (Appendix~\ref{app:e2}), so the judge panel keeps the trajectory and the replay judge values are reported as a single-session column in Table~\ref{tab:polish-budget}. Round $i$ costs $1{+}2i$ LLM calls and $i{+}1$ scored candidates, Selected is the quick-scorer argmax over the six trajectory programs ($12$ LLM calls), and Final adds seed search with at most $3{,}003$ scored candidates and no LLM calls. Dashed lines are the baselines' flat means under their single-sample protocol (three sampled seeds per prompt). MatFuse's flat BLIPScore ($8.3$) lies below the plotted range. Lines connect measured operating points and do not imply measured intermediate budgets between them.}
    \label{fig:budget-curve}
\end{figure}

This section completes the budget accounting behind the headline comparison: the matched round-$0$ comparison (Table~\ref{tab:budget0}, Section~\ref{sec:results-stages}), the same comparison across all program-generating backbones (Table~\ref{tab:round0}), best-of-$N$ selection for the diffusion baselines (Tables~\ref{tab:bestofn} and~\ref{tab:bestofn-oracle}), the per-prompt cost accounting (Table~\ref{tab:cost}), and the quality-versus-budget curve (Figure~\ref{fig:budget-curve}).

\paragraph{Round $0$ across backbones.}
Table~\ref{tab:round0} extends the budget-matched round-$0$ comparison of Table~\ref{tab:budget0} beyond the flagship, scoring every backbone's one-call initial program with the same stack.
Three of the five non-flagship backbones exceed StableMaterials' flat BLIPScore at a single call, significantly so for \texttt{gpt-5.6-luna}, \texttt{glm-5.2}, and \texttt{deepseek-v4-flash-0731} at $+11.6$ to $+12.7$ points (Holm-corrected $p{<}0.003$).
Gemma and Qwen trail StableMaterials on flat BLIPScore at $25.21$ and $27.01$ against $28.57$, and their flat CLIPScore sits significantly below it, so the round-$0$ advantage is a property of the stronger backbones rather than of the representation alone.

\begin{table}[htbp]
    \caption{Round-$0$ (one LLM call, one scored candidate) flat-layout means for every program-generating backbone, mean over per-prompt seed means ($141$ prompts, seeds $42$, $123$, and $2026$). Stars mark paired two-sided Wilcoxon significance versus StableMaterials with the backbone above it ($^{*}p{<}0.05$, $^{**}p{<}0.01$, $^{***}p{<}0.001$, Holm-corrected within the $15$ tests), and $\dagger$ marks cells significantly below. The flagship row repeats the committed round-$0$ means of Table~\ref{tab:stages}.}
    \label{tab:round0}
    \centering
    \footnotesize
    \setlength{\tabcolsep}{6pt}
    \begin{tabular}{l ccc}
        \toprule
        Backbone (round 0) & Flat BLIP & Flat CLIP & Flat VQA \\
        \midrule
        gpt-5.6-luna & 40.21$^{**}$ & 26.14 & 47.60 \\
        glm-5.2 & 40.41$^{**}$ & 26.45 & 48.13 \\
        deepseek-v4-flash-0731 & 41.28$^{***}$ & 26.19 & 47.79 \\
        gemma-4-26b-a4b-it & 25.21 & 23.64$^{\dagger}$ & 43.88 \\
        qwen3.6-35b-a3b & 27.01 & 23.59$^{\dagger}$ & 41.79 \\
        \midrule
        gemini-3.6-flash (flagship) & 48.31 & 27.70 & 53.34 \\
        StableMaterials & 28.57 & 25.66 & 45.36 \\
        \bottomrule
    \end{tabular}
\end{table}

\paragraph{Best-of-N selection.}
Each diffusion baseline samples three seeds per prompt, so Table~\ref{tab:bestofn} grants it best-of-$3$ selection over those samples.
StableMaterials additionally received a second sampling wave of seven seeds per prompt from the same released checkpoint, so its best-of-$10$ rows grant selection over all ten samples under the same two rules.
The MatFuse archive additionally stores two sampler variants beside the released CFG variant in every run directory, so its best-of-$9$ rows grant selection over all nine seed-by-variant samples under the same quick scorer, while its main-table rows use the CFG variant only (Table~\ref{tab:baseline-config}).
Table~\ref{tab:bestofn} selects each baseline's sample with the MobileCLIP2 quick scorer that our own trajectory selector uses, so the selection rule is symmetric.
Table~\ref{tab:bestofn-oracle} instead selects per metric with the evaluation score itself, an infeasible bound that requires the test labels, which we disclose rather than adopt.
Under the symmetric selector the flagship keeps a positive mean delta on seven of eight cells against StableMaterials (staged judge $-1.1$) and on all eight cells against IntrinsiX and MatFuse.
Extending StableMaterials to best-of-$10$ under the same scorer keeps the flagship ahead on seven of eight cells, with staged CLIPScore now the exception at $-0.2$ (a statistical tie), and turns both judge cells into significant leads, $+15.6$ flat ($p{=}5.7\times 10^{-8}$) and $+12.2$ staged ($p{=}6.3\times 10^{-6}$), paired within a single scoring session that scored both sides.
The flagship also keeps all eight cells against MatFuse under best-of-$9$ selection, with paired deltas between $+5.0$ and $+46.0$ points, all $p{<}0.001$.
Under the oracle the two judge cells against StableMaterials flip ($-3.5$ flat and $-9.2$ staged) and both IntrinsiX judge cells turn negative, while the flat BLIPScore, CLIPScore, and VQAScore leads against StableMaterials persist ($+14.7$, $+1.2$, and $+2.7$).
Over the same ten StableMaterials samples the oracle instead shrinks the judge cells against StableMaterials to $+6.1$ flat ($p{=}0.04$) and $+1.2$ staged (a statistical tie), both paired within a single scoring session, while the staged BLIPScore, staged CLIPScore, staged VQAScore, and flat VQAScore cells move to StableMaterials ($-6.9$, $-1.7$, $-4.9$, and $-3.5$).
MatFuse never approaches under either rule, with its smallest paired delta $+4.7$ points.
Table~\ref{tab:bestofn-abs} reports the underlying absolute means, with each cell giving the flat and staged values.

\begin{table}[htbp]
    \caption{Paired mean differences of the \texttt{gemini-3.6-flash} flagship against best-of-$N$ selection for each diffusion baseline, with the deployed MobileCLIP2 quick scorer ($n{=}141$ paired prompts, two-sided Wilcoxon, $^{*}p{<}0.05$, $^{**}p{<}0.01$, $^{***}p{<}0.001$), in the flat (a) and staged (b) layouts. Bracketed $95\%$ bootstrap intervals are shown for cells without stars. $N{=}3$ selects over the baseline's three sampled seeds, the MatFuse best-of-$9$ rows select over its nine stored seed-by-variant samples (three seeds times the basic, EMA, and CFG sampler variants; their judge cells pair the CFG-variant samples, the subset carrying original-evaluation judge scores), and the StableMaterials best-of-$10$ rows select over its ten sampled seeds (the three recorded seeds plus seven additional samples from the same released checkpoint). The first row of each pair compares the baseline against our final polished output, and the indented row against our pre-polish selected program. The daggered judge cells of the StableMaterials best-of-$10$ rows pair a fresh judge score of our final output with the same scoring session that scored those selections, while every other judge cell pairs within the original evaluation, and judge scores from different sessions are not level-comparable (Appendix~\ref{app:e2}), so daggered and undaggered judge cells are not comparable. The indented judge cells of the daggered rows are dashed because only the final output was re-judged.}
    \label{tab:bestofn}
    \centering
    \footnotesize
    \setlength{\tabcolsep}{5pt}
    {(a) Flat layout.}\\[2pt]
        \begin{tabular}{l cccc}
            \toprule
            Baseline & BLIP & CLIP & VQA & Judge \\
            \midrule
            IntrinsiX best-of-3 & $+$15.7$^{***}$ & $+$2.5$^{***}$ & $+$5.5$^{**}$ & $+$2.3 [$-$3,$+$8] \\
            \quad vs.\ Selected & $+$13.8$^{***}$ & $+$2.4$^{***}$ & $+$4.7$^{*}$ & $+$1.9 [$-$4,$+$7] \\
            MatFuse best-of-3 & $+$46.3$^{***}$ & $+$7.5$^{***}$ & $+$21.3$^{***}$ & $+$23.3$^{***}$ \\
            \quad vs.\ Selected & $+$44.4$^{***}$ & $+$7.4$^{***}$ & $+$20.5$^{***}$ & $+$22.8$^{***}$ \\
            MatFuse best-of-9 & $+$46.0$^{***}$ & $+$7.1$^{***}$ & $+$20.2$^{***}$ & $+$21.6$^{***}$ \\
            \quad vs.\ Selected & $+$44.1$^{***}$ & $+$7.0$^{***}$ & $+$19.5$^{***}$ & $+$21.2$^{***}$ \\
            StableMaterials best-of-3 & $+$19.8$^{***}$ & $+$1.8$^{***}$ & $+$6.8$^{***}$ & $+$4.9 [$+$0,$+$10] \\
            \quad vs.\ Selected & $+$18.0$^{***}$ & $+$1.7$^{***}$ & $+$6.1$^{***}$ & $+$4.5 [$-$0,$+$10] \\
            StableMaterials best-of-10 & $+$13.4$^{***}$ & $+$0.6 [$-$0,$+$1] & $+$2.3 [$-$1,$+$5] & $+$15.6$^{***\dagger}$ \\
            \quad vs.\ Selected & $+$11.5$^{***}$ & $+$0.5 [$-$0,$+$1] & $+$1.6 [$-$1,$+$5] & -- \\
            \bottomrule
        \end{tabular}\\[6pt]
    {(b) Staged layout.}\\[2pt]
        \begin{tabular}{l cccc}
            \toprule
            Baseline & BLIP & CLIP & VQA & Judge \\
            \midrule
            IntrinsiX best-of-3 & $+$9.9$^{***}$ & $+$1.8$^{***}$ & $+$7.2$^{***}$ & $+$1.9 [$-$3,$+$7] \\
            \quad vs.\ Selected & $+$9.3$^{***}$ & $+$1.8$^{***}$ & $+$7.0$^{***}$ & $+$1.7 [$-$4,$+$7] \\
            MatFuse best-of-3 & $+$26.1$^{***}$ & $+$5.3$^{***}$ & $+$22.0$^{***}$ & $+$20.1$^{***}$ \\
            \quad vs.\ Selected & $+$25.4$^{***}$ & $+$5.3$^{***}$ & $+$21.8$^{***}$ & $+$20.0$^{***}$ \\
            MatFuse best-of-9 & $+$25.4$^{***}$ & $+$5.0$^{***}$ & $+$20.4$^{***}$ & $+$19.7$^{***}$ \\
            \quad vs.\ Selected & $+$24.7$^{***}$ & $+$4.9$^{***}$ & $+$20.2$^{***}$ & $+$20.0$^{***}$ \\
            StableMaterials best-of-3 & $+$9.9$^{***}$ & $+$0.8 [$-$0,$+$2] & $+$6.3$^{***}$ & $-$1.1 [$-$6,$+$3] \\
            \quad vs.\ Selected & $+$9.2$^{***}$ & $+$0.7 [$-$0,$+$2] & $+$6.1$^{***}$ & $-$1.2 [$-$6,$+$3] \\
            StableMaterials best-of-10 & $+$5.3 [$-$0,$+$11] & $-$0.2 [$-$1,$+$1] & $+$3.1 [$+$0,$+$6] & $+$12.2$^{***\dagger}$ \\
            \quad vs.\ Selected & $+$4.6 [$-$1,$+$10] & $-$0.3 [$-$1,$+$0] & $+$2.9 [$-$0,$+$6] & -- \\
            \bottomrule
        \end{tabular}
\end{table}

\begin{table}[htbp]
    \caption{The same comparison under a per-metric oracle over the baseline's samples, an infeasible upper bound that requires the test labels, which we disclose rather than adopt. Layout, stars, brackets, pair structure, and dagger conventions follow Table~\ref{tab:bestofn}.}
    \label{tab:bestofn-oracle}
    \centering
    \footnotesize
    \setlength{\tabcolsep}{5pt}
    {(a) Flat layout.}\\[2pt]
        \begin{tabular}{l cccc}
            \toprule
            Baseline & BLIP & CLIP & VQA & Judge \\
            \midrule
            IntrinsiX best-of-3 & $+$11.5$^{**}$ & $+$1.9$^{***}$ & $+$1.5 [$-$2,$+$5] & $-$4.8$^{*}$ \\
            \quad vs.\ Selected & $+$9.7$^{**}$ & $+$1.8$^{**}$ & $+$0.8 [$-$3,$+$4] & $-$5.2$^{**}$ \\
            MatFuse best-of-3 & $+$43.4$^{***}$ & $+$6.9$^{***}$ & $+$18.8$^{***}$ & $+$16.5$^{***}$ \\
            \quad vs.\ Selected & $+$41.5$^{***}$ & $+$6.8$^{***}$ & $+$18.1$^{***}$ & $+$16.1$^{***}$ \\
            StableMaterials best-of-3 & $+$14.7$^{***}$ & $+$1.2$^{*}$ & $+$2.7$^{*}$ & $-$3.5 [$-$8,$+$1] \\
            \quad vs.\ Selected & $+$12.8$^{***}$ & $+$1.1$^{*}$ & $+$2.0 [$-$1,$+$5] & $-$3.9$^{*}$ \\
            StableMaterials best-of-10 & $+$0.9 [$-$5,$+$7] & $-$0.4 [$-$1,$+$0] & $-$3.5$^{**}$ & $+$6.1$^{*\dagger}$ \\
            \quad vs.\ Selected & $-$1.0 [$-$7,$+$5] & $-$0.5 [$-$1,$+$0] & $-$4.3$^{***}$ & -- \\
            \bottomrule
        \end{tabular}\\[6pt]
    {(b) Staged layout.}\\[2pt]
        \begin{tabular}{l cccc}
            \toprule
            Baseline & BLIP & CLIP & VQA & Judge \\
            \midrule
            IntrinsiX best-of-3 & $+$5.6$^{*}$ & $+$1.1$^{*}$ & $+$2.7 [$-$1,$+$6] & $-$6.6$^{**}$ \\
            \quad vs.\ Selected & $+$4.9$^{*}$ & $+$1.1$^{*}$ & $+$2.5 [$-$1,$+$6] & $-$6.8$^{**}$ \\
            MatFuse best-of-3 & $+$24.2$^{***}$ & $+$4.7$^{***}$ & $+$19.0$^{***}$ & $+$10.9$^{***}$ \\
            \quad vs.\ Selected & $+$23.5$^{***}$ & $+$4.7$^{***}$ & $+$18.8$^{***}$ & $+$10.8$^{***}$ \\
            StableMaterials best-of-3 & $+$4.0 [$-$2,$+$10] & $-$0.1 [$-$1,$+$1] & $+$2.3 [$-$1,$+$5] & $-$9.2$^{***}$ \\
            \quad vs.\ Selected & $+$3.3 [$-$3,$+$9] & $-$0.2 [$-$1,$+$1] & $+$2.1 [$-$1,$+$5] & $-$9.3$^{***}$ \\
            StableMaterials best-of-10 & $-$6.9$^{**}$ & $-$1.7$^{***}$ & $-$4.9$^{***}$ & $+$1.2$^{\dagger}$ [$-$3,$+$6] \\
            \quad vs.\ Selected & $-$7.5$^{**}$ & $-$1.7$^{***}$ & $-$5.1$^{***}$ & -- \\
            \bottomrule
        \end{tabular}
\end{table}

\begin{table}[htbp]
    \caption{Absolute means underlying Tables~\ref{tab:bestofn} and~\ref{tab:bestofn-oracle}, with each cell giving the flat and staged values. Ours rows are the final polished output and the pre-polish selected program of the \texttt{gemini-3.6-flash} flagship, and baseline rows are its quick-scorer or oracle best-of-$3$ selections, plus MatFuse's best-of-$9$ over its nine stored samples and StableMaterials' best-of-$10$ over its ten sampled seeds. The daggered StableMaterials best-of-$10$ judge cells were scored in a separate scoring session while the ours judge cells and the best-of-$3$ judge cells come from the original evaluation, so the apparent best-of-$10$ drop below the best-of-$3$ judge means (for example $57.26$ versus $62.09$ flat) is a batch effect, not a selection effect (Appendix~\ref{app:e2}). The paired judge deltas of Tables~\ref{tab:bestofn} and~\ref{tab:bestofn-oracle} therefore use re-judged finals from that same session (flat $72.85$, staged $61.51$) on our side.}
    \label{tab:bestofn-abs}
    \centering
    \footnotesize
    \setlength{\tabcolsep}{4pt}
        \begin{tabular}{l cccc}
            \toprule
            Method & BLIP & CLIP & VQA & Judge \\
            \midrule
            Ours (Final) & 56.06/36.14 & 28.80/25.30 & 54.71/54.26 & 67.01/57.34 \\
            Ours (Selected) & 54.17/35.50 & 28.71/25.25 & 53.97/54.07 & 66.56/57.20 \\
            \midrule
            IntrinsiX best-of-3 (MobileCLIP2) & 40.39/26.20 & 26.33/23.46 & 49.23/47.04 & 64.70/55.49 \\
            IntrinsiX best-of-3 (oracle) & 44.52/30.58 & 26.90/24.16 & 53.17/51.61 & 71.77/63.96 \\
            MatFuse best-of-3 (MobileCLIP2) & 9.76/10.08 & 21.35/19.96 & 33.42/32.31 & 43.71/37.19 \\
            MatFuse best-of-9 (MobileCLIP2) & 10.07/10.77 & 21.70/20.31 & 34.47/33.87 & 45.16/38.34 \\
            MatFuse best-of-3 (oracle) & 12.62/11.98 & 21.92/20.55 & 35.89/35.31 & 50.48/46.40 \\
            StableMaterials best-of-3 (MobileCLIP2) & 36.21/26.26 & 27.02/24.53 & 47.91/47.95 & 62.09/58.44 \\
            StableMaterials best-of-10 (MobileCLIP2) & 42.63/30.87 & 28.18/25.52 & 52.37/51.20 & 57.26/49.34$^{\dagger}$ \\
            StableMaterials best-of-3 (oracle) & 41.38/32.15 & 27.58/25.41 & 51.98/51.99 & 70.48/66.51 \\
            StableMaterials best-of-10 (oracle) & 55.13/43.03 & 29.19/26.96 & 58.26/59.12 & 66.77/60.27$^{\dagger}$ \\
            \bottomrule
        \end{tabular}
\end{table}

\paragraph{Cost accounting.}
Table~\ref{tab:cost} aggregates the flagship's $423$ recorded authoring runs and the stored baseline samples, where a run is one prompt at one seed.
Each flagship run makes $12$ recorded LLM calls, and the stage totals are $5.0$ hours for generation, $5.4$ for critique, $0.3$ for selection, and $33.5$ for seed search, so search takes $75\%$ of the $44.7$ hour total.
The median run takes $4.6$ minutes and scores at most $3{,}003$ candidates at a median $70$ milliseconds each.
The baselines make no LLM calls, score no candidates, and sample in $0.5$ to $44.4$ seconds per $512{\times}512$ sample on their original GPU stacks.
Timings were recorded on the different systems noted in the table, so this accounting quantifies budgets rather than cross-system latency.

\begin{table}[htbp]
    \caption{Recorded cost accounting over each method's $423$ runs ($141$ prompts $\times$ $3$ seeds), with the hardware row recording where each side's timings were taken. For \method, mean seconds per run is the $44.7$ hour total divided by $423$ (median $4.6$ minutes), and scored candidates are the seed-search evaluations, at most $3{,}003$ per run. For the baselines, seconds per run is the mean sampling time of one $512{\times}512$ sample on their original GPU stack. The baselines are feed-forward samplers, so their LLM-call and scored-candidate counts are zero: their entire generation budget is sampling time.}
    \label{tab:cost}
    \centering
    \footnotesize
    \setlength{\tabcolsep}{5pt}
    \begin{tabular}{l cccc}
        \toprule
        Method                      & Mean s / run & LLM calls / run & Scored / run & Total (h) \\
        \midrule
        Ours (gemini-3.6-flash)     & 381          & 12              & $\le\!3{,}003$ & 44.7      \\
        IntrinsiX                   & 44.4         & 0               & 0            & 5.2        \\
        MatFuse                     & 12.9         & 0               & 0            & 1.5        \\
        StableMaterials             & 0.5          & 0               & 0            & 0.1        \\
        \midrule
        \multicolumn{5}{l}{Hardware: \method{} on an Apple-silicon CPU with API LLMs, baselines on their original CUDA GPUs.} \\
        \bottomrule
    \end{tabular}
\end{table}

\paragraph{Quality versus budget.}
Figure~\ref{fig:budget-curve} plots flat-layout final metrics against the scored-candidate budget along the flagship's recorded trajectory: rounds $0$ through $5$, the selected program, and the final polished output.
Every point is an external evaluation metric, never the internal quick scorer, and each value is the mean of per-prompt seed means over $141$ prompts and $3$ runs.
The critique-revision loop ends at six scored candidates and $12$ LLM calls, and only seed search moves the budget to $3{,}009$ scored candidates, adding $+1.9$ BLIPScore and $+0.4$ judge over the selected program.
Truncated seed-sweep replays are now scored with these final metrics: the open triangles in the BLIPScore panel mark the re-rendered replay winners at $K{=}1$ to $K{=}500$ (Table~\ref{tab:polish-budget}).
The replay budgets were also judge-scored, in a single scoring session, and their flat judge means stay in a $1.0$-point band ($70.2$ to $71.2$) from $K{=}1$ to $K{=}500$, the same flat pattern the local metrics show, so the seed sweep does not move the judge either.
Those judge scores come from a session later than the main evaluation, and a judge's absolute level is not stable across sessions: re-judging StableMaterials' unchanged renders in that session moved their mean by $-13.8$ points, and our $K{=}500$ replay winners, whose programs nearly coincide with the full sweep's, read $70.3$ there against the original final mean of $67.0$.
The judge panel therefore keeps the trajectory points only, and the replay judges appear as a single-session column in Table~\ref{tab:polish-budget}.
Dashed lines mark the baselines' flat means under their single-sample protocol, and MatFuse's flat BLIPScore lies below the plotted range.

\section{Material Apprentice Native Run}
\label{app:ma}
This control runs the released Material Apprentice (MA) system~\citep{gupta2026apprentice}, the closest text-to-procedural baseline, natively on this benchmark under our rendering and scoring stack.
MA retrieves expert process traces, compiles them into shader-graph programs with an LLM, and executes them in Blender, so it shares the executable-program setting while targeting a different representation.

\paragraph{Protocol.}
The system under test is their public release pinned at commit \texttt{6811f66}, launched through \texttt{sceneprogexec} $0.2.6$ with pipeline package versions \texttt{sceneprogllm} $0.4.0$ and \texttt{sceneprogsyn} $0.1.1$ on Blender $4.5.4$ LTS.
All five pipeline modules (generator, retrieval matcher, compiler, sampler, evaluator) plus the dreamer and exporter use \texttt{gpt-5} at their defaults: internal seed $124$, temperature $0.8$, reasoning effort medium, five procedures, ten samples, three best-of-$10$ sampling rounds, and five compile-and-execute retries.
The protocol document, the prompt subset, and both transport-shim diffs were frozen and sha256-pinned before any generation call.
Every delivered \texttt{.blend} is rendered by our adapter, which mirrors the benchmark's two layouts exactly (same hero plane, camera, lights, samples, and shadow settings), and scored by the same stack as Table~\ref{tab:main}.

\paragraph{Prompt subset.}
At roughly one hour and $1.5$M prompt tokens per prompt, running all $141$ prompts was infeasible in the time available, so the comparison uses a frozen $45$-prompt subset: $10$ category, $4$ MatSynth, $16$ StableMaterials, and $15$ text2fabric prompts, a seeded draw mirroring the source proportions and committed before any run.
Claims in this section cover that subset only.

\paragraph{Disclosures.}
Three conditions bound the comparison.
First, we run MA's text-only mode, which dreams a reference image instead of receiving one, while their evaluated setting is image-plus-text conditioning, so their appearance cells here are off their evaluated distribution and we read them as a lower bound on their system rather than a defeat of it.
Second, our OpenAI-compatible proxy drops the Responses-API image tool and large concurrent multimodal chats, so two transport shims were required: the dreamer's image call is rerouted to the equivalent images endpoint with identical prompt, size, quality, and background parameters, and chat calls run under a concurrency cap of two with retry backoff (diffs sha256-pinned in the protocol, affecting transport only).
Prompts, models, sampling parameters, and pipeline logic are untouched.
Third, the robust content of this control is execution success and cost, not the appearance cells.

\paragraph{Execution and cost.}
MA executed all $45$ prompts end-to-end with zero failures: each run dreamed a reference, retrieved traces, compiled and executed a procedure, sampled three best-of-$10$ rounds, saved a \texttt{.blend}, and delivered both renders through our adapter.
Per prompt it made $112$ chat calls on average (range $84$ to $123$), consumed about $1.5$M prompt and $0.4$M completion tokens, and took $41$ to $84$ minutes end-to-end at two workers (median $61$).
The flagship's recorded runs on the same prompts take a median $4.6$ minutes with $12$ calls.
MA's cost profile is intrinsic to their multi-module pipeline, and their published first-shot direct-graph failure rate of roughly $30\%$ is corroborated by our Blender-Python funnel in Appendix~\ref{app:bpy}.

\begin{table}[htbp]
    \caption{Material Apprentice native run on the frozen $45$-prompt subset against the flagship's recorded final programs and StableMaterials on the same prompts (flat layout, means over per-prompt seed means, our rendering and scoring stack). MA ran one attempt per prompt under their default \texttt{gpt-5} configuration. LLM calls are per prompt (the mean over MA's $45$ prompts, range $84$--$123$). Min/prompt is end-to-end wall at two workers for MA (median $61$) and the recorded median authoring run for ours. MA's judge cell was scored in a separate session while the ours and StableMaterials judge cells come from the original evaluation, so the judge column's absolute levels are not comparable across rows, and the paired judge comparison in the text uses re-judged finals from that session on our side. MA's appearance cells are a lower bound (text-only mode, shimmed image transport), and execution success and cost are the robust content.}
    \label{tab:ma}
    \centering
    \footnotesize
    \setlength{\tabcolsep}{4pt}
    \resizebox{\linewidth}{!}{%
    \begin{tabular}{l c cccc c c}
        \toprule
         & Executed & \multicolumn{4}{c}{Flat means} & Min/ & LLM calls/ \\
         & (of 45) & BLIP & CLIP & VQA & Judge & prompt & prompt \\
        \midrule
        MA text-only (gpt-5) & 45/45 & 17.98 & 21.97 & 32.68 & 33.02 & 41--84 & 112 \\
        \method (gemini-3.6-flash) & 45/45 & 58.94 & 29.12 & 57.22 & 68.49 & 4.6 & 12 \\
        StableMaterials & 45/45 & 26.69 & 25.38 & 43.63 & 55.93 & $<$0.1 & 0 \\
        \bottomrule
    \end{tabular}}
\end{table}

\paragraph{Paired appearance deltas.}
Paired per prompt on the $45$-prompt subset, the flagship's final programs lead MA's flat means by $+41.0$ BLIPScore ($95\%$ CI $[+29.9,+51.9]$), $+7.2$ CLIPScore ($[+5.5,+8.9]$), and $+24.5$ VQAScore ($[+18.9,+30.6]$), all at $p{<}10^{-7}$, with staged deltas of $+28.6$, $+7.2$, and $+23.4$.
StableMaterials leads MA on flat CLIPScore ($+3.4$, $p{=}0.0017$) and VQAScore ($+11.0$, $p{=}0.0003$) with flat BLIPScore a statistical tie ($+8.7$, $p{=}0.13$).
MA's judge mean in that session is $33.0$, and paired within the same session our re-judged finals lead it by $+40.1$ flat ($95\%$ CI $[+30.2,+49.7]$, $p{=}2.4\times 10^{-7}$) and $+35.2$ staged ($[+26.1,+43.8]$, $p{=}3.3\times 10^{-7}$) over the $45$ shared prompts.
Given the lower-bound disclosures above, we do not headline these appearance margins.

\section{Representation and Method Details}
\label{app:rep-method-details}

\subsection{Formal Dependency Statement}
\label{app:rep-dependencies}

Let $G$ denote a fixed canonical program structure, $z$ its numeric parameters and explicit seed coordinates, and $F_k(u;G,z)$ the evaluated material channel $k$ at position $u$.
Let $D_k$ be the set of coordinates in $z$ that occur in the transitive source dependencies of $F_k$, including the coverage expressions used in compositing.
For two parameter assignments $z$ and $z'$ under the same engine version, settings, and sampling coordinates, finite deterministic evaluation gives
\begin{equation}
    z_{D_k}=z'_{D_k}
    \quad\Longrightarrow\quad
    F_k(u;G,z)=F_k(u;G,z').
    \label{eq:slate-dependencies}
\end{equation}
This is a consequence of evaluating unchanged dependencies, not a guarantee that an edit produces a perceptually local image change.
A roughness edit may alter reflected highlights across the rendered surface even when the color and height maps are unchanged.
Conversely, a coverage edit generally enters several composited channels, and a height edit also propagates to derived normals and approximate ambient occlusion.

\subsection{Channel Semantics and Validation}
\label{app:rep-channels}

Each layer carries coverage $\alpha\in[0,1]$, sRGB base color in $[0,255]$, surface-response channels including roughness and metallicity, emission, and height $h$ in coordinate units.
Each scalar component may be an expression over position.
For finite values, the engine clamps coverage and weights to $[0,1]$, color to its stated range, index of refraction to at least $1$, and emissive strength to nonnegative values.
Height is not clamped, and omitted channels take documented defaults.
The surface channels correspond to a subset of OpenPBR inputs~\citep{portsmouth2025openpbr} and are mapped to renderer parameters such as the Principled BSDF~\citep{burley2012principled}.

Coverage and material channels share arithmetic, coordinate transforms, noise (\dsl{fBm}, \dsl{Worley}), periodic patterns (\dsl{Bricks}, \dsl{Weave}), and filled or stroked shapes.
Definitions can refer to earlier definitions, forming an acyclic dependency graph that layer expressions reference.
Binary-mask union, intersection, complement, and difference use $\max(a,b)$, $\min(a,b)$, $1-a$, and $a(1-b)$.
Periodic primitives support tiling over compatible periods.
Arbitrary programs and sampling windows need not be seamless.

\subsection{Tile Program Dependencies}
\label{app:rep-tile-dependencies}

Listing~\ref{lst:example} makes source dependencies inspectable.
Its bottom layer describes grout, while the upper layer uses \dsl{tileMask} as coverage and as a factor in its height expression.
Changing the mortar parameter of this shared mask changes both the exposed grout region and the tile relief boundary.
The explicit multiplication by \dsl{tileMask} tapers height at the edge, so the taper comes from the source program rather than the compositing rule.

The upper layer's \dsl{glazeGrain} field appears only in its roughness expression, whereas \dsl{glazeRipple} appears in both base color and height.
With the remaining canonical source and sampling settings fixed, an edit confined to \dsl{glazeGrain} preserves the color and height maps, while an edit to \dsl{glazeRipple} can change both.
These are checkable properties of this program's dependencies, and do not imply that an automatically generated program will choose an equally useful decomposition.

\subsection{Serialization Controls}
\label{app:rep-serialization}

An inlined serialization removes named definitions by replacing every use with a copy of its resolved expression while keeping explicit noise seeds verbatim.
Of the $2538$ stored initial programs, $2514$ ($99\%$) use named definitions, inlining raises source length by a median factor of $1.4$, and the inlined form re-parses and exports channel maps byte-identical to the named-field form for every one of these programs.
This equivalence lets future generation and editing comparisons vary explicit reuse and serialization length while holding evaluation semantics fixed.

An explicit-channel form without layer grouping has no established equivalent yet.
Expanding the layer stack faithfully must preserve the zero-coverage branch, color-space conversion, and separate maximum-height rule, and a generic weighted sum would change the target semantics.
Direct Blender, native graph, or other material-generation systems provide complementary end-to-end comparisons, but changing their library and execution interface does not isolate the effect of named fields or layers.

The seed-variant mechanism replaces repeated occurrences of the same numeric seed together, preserving equality groups and all non-seed source within each candidate structure.
This is a source-level constraint.
Attribute preservation and perceptual diversity still require measurement, especially when a seeded field drives coverage or relief.

\subsection{Execution Details}
\label{app:rep-execution}

The Python reference evaluates fields on a pixel grid and exports maps.
A TypeScript port supports browser inspection and editing.
A canonicalized program with every stochastic seed explicit, fixed sampling coordinates, and a fixed engine version and settings defines repeatable field evaluations within that engine.
Omitted seeds must be resolved before comparing executions or attributing a change to an edit.

Parser checks cover syntax, named references, and constructors, while a finite regression suite checks serialization against the canonical EBNF (Appendix~\ref{app:grammar}).
Neither establishes finite fields or successful rendering for every accepted program.
For example, a well-formed square-root expression can be undefined at some positions.
Generator instructions also distinguish color visibility from maximum height and explain that constant height creates no slope, while linear height produces a tilted normal.
Finite cross-engine tests have explicit tolerances, differing Worley hashes, and excluded GPU shading (Appendix~\ref{app:engine}).
They do not establish identical images across renderers.

\subsection{Algorithmic Overview}
\label{app:method-algorithmic-overview}

\begin{figure}[htbp]
    \centering
    \input{figures/pipeline.tex}
    \caption{The three-stage pipeline in Figure~\ref{fig:overview}, summarized algorithmically.
        \textbf{I.~Generation} proposes a program and repairs parser errors.
        \textbf{II.~Critique and refinement} revises it using a quick render, material statistics, and source code where configured.
        \textbf{III.~Selection and seed search} scores the trajectory and explores noise realizations of up to three candidate programs while fixing each candidate's remaining source.
        The loop uses a lightweight preview renderer and requires no path tracer.}
    \label{fig:algorithmic-overview}
\end{figure}

\subsection{Prompt, Critic, and Renderer Mechanics}
\label{app:method-mechanics}

The generator receives an operational DSL reference describing structure, channel ranges, expressions, and compositing semantics.
Six few-shot examples span one to four layers, and an organic-texture playbook supplies idioms such as combining low- and high-frequency \dsl{fBm} fields (Appendix~\ref{app:prompts}).
The language reference is supplied as context, without constraining the model's token-level decoding.

The quick render evaluates the composited maps and height-derived normals, then shades a head-on swatch under a fixed directional light.
It uses approximate Blinn-Phong specular shading, Fresnel, sheen, subsurface, and transmission terms rather than path tracing.
It does not reproduce global illumination or physical refraction, so feedback can miss effects that become visible in a staged render.
The material statistics contain $28$ descriptors of the unlit channel maps, covering albedo, texture, roughness, metallicity, height, normal tilt, occlusion, and other channel summaries.
Each statistic is paired with a short definition.

Trajectory selection and seed search both use quick renders, but they score different render resolutions.
Both scorer calls use the text template ``A photo of $t$'' for a prompt $t$.
Shared preprocessing does not make $512{\times}512$ selection inputs equivalent to $256{\times}256$ seed-search inputs.
Within each seed sweep, all non-seed parameters and structure stay fixed, but the winning output can come from the initial, selected, or final trajectory program.

The preview renderer makes candidate search practical without a path tracer.
In the flagship recorded runs, a sweep of $1001$ seed variants renders $256{\times}256$ previews and scores them in a median of $70$ seconds, about $70$ milliseconds per candidate, and all stages ran sequentially in one process per prompt.
Appendix~\ref{app:runtime} reports stage-wise latency for every backbone.

\section{Additional Experimental Analyses}
\label{app:experiments-9page-extra}

This section provides further protocol and diagnostic details for the evaluation in Section~\ref{sec:eval}.

\subsection{Protocol Details}
\label{app:protocol-9page}

The representation targets surface appearance through spatial patterns, layers, and material channels.
We retain prompts with figurative or highly specific motifs to test where this representation becomes restrictive.
Every prompt is evaluated with both rendering layouts, and every scored run is retained in the quantitative tables.
The text-only backbones, DeepSeek and GLM, receive code and statistics without the quick render during critique.
The retained run tree contains $423$ complete conversations per backbone and no aborted attempts; per-turn repair counts are not logged, so first-pass validity within a turn is not measured.
The flagship's recorded median authoring time is $4.6$ minutes, with a median $68\%$ of time spent in polishing, the Stage III seed search (Appendix~\ref{app:runtime}).
Compact output size therefore should not be read as cheap authoring.

\subsection{Selected Refinement Trajectory}
\label{app:refinement-trajectory-9page}

\begin{figure}[htbp]
    \centering
    \setlength{\tabcolsep}{1.5pt}
    \begin{tabular}{@{}cccccc@{}}
        {\scriptsize Round 0}                             & {\scriptsize Round 1} & {\scriptsize Round 2} & {\scriptsize Round 3} & {\scriptsize Round 4} & {\scriptsize Round 5} \\[2pt]
        \includegraphics[width=0.16\linewidth]{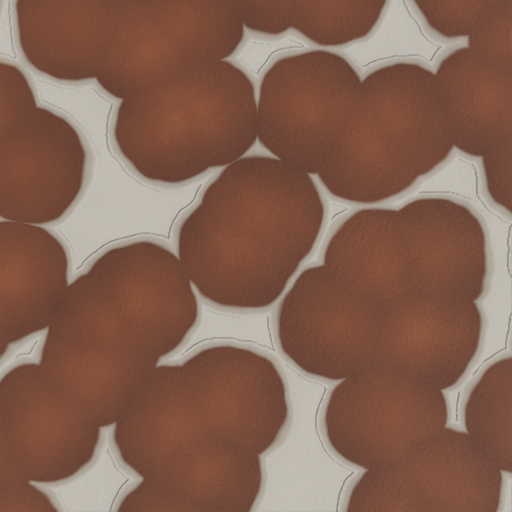} &
        \includegraphics[width=0.16\linewidth]{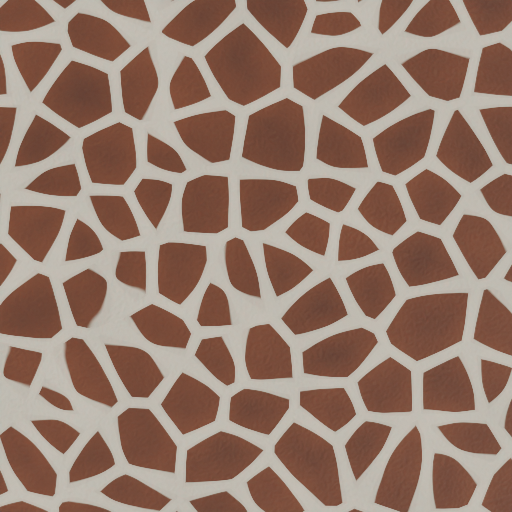} &
        \includegraphics[width=0.16\linewidth]{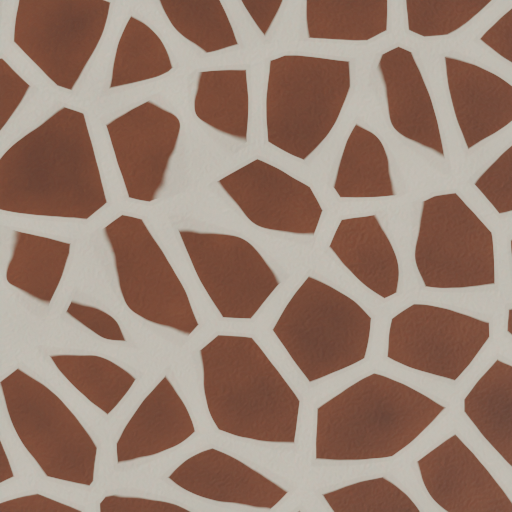} &
        \includegraphics[width=0.16\linewidth]{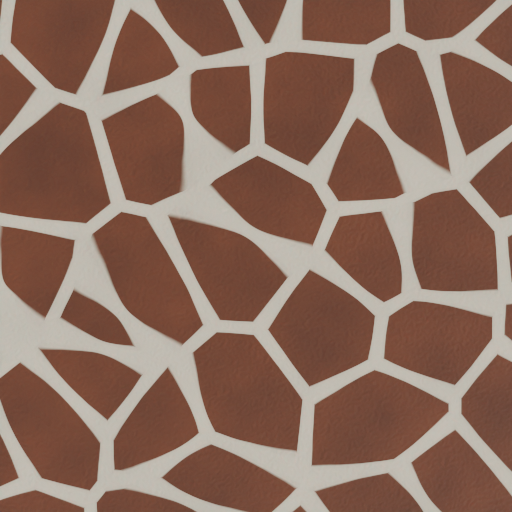} &
        \includegraphics[width=0.16\linewidth]{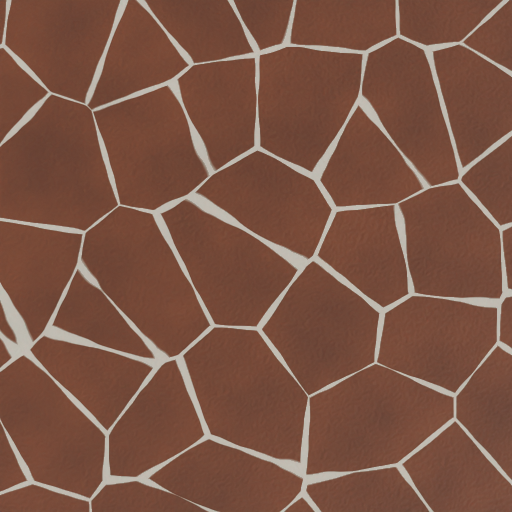} &
        \includegraphics[width=0.16\linewidth]{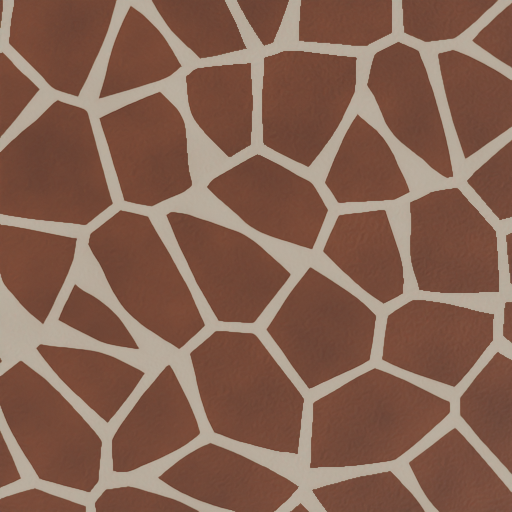}
    \end{tabular}
    \caption{Selected refinement trajectory for ``Giraffe skin with large polygonal patches.'' The figure-selection procedure filters for BLIPScore/judge gains and ranks candidates by visual change. This example's flat BLIPScore rises from $2.8$ to $99.8$, and its round-$5$ judge score is $16$ points above round $0$. The polished final output is omitted because it contains a smudge accepted by the quick scorer. This illustration does not estimate the frequency of successful or monotonic refinement.}
    \label{fig:refine}
\end{figure}

In this selected case, rounded and partly merged patches in round $0$ become distinct polygons in round $1$.
Later revisions change polygon scale and boundary width.
The sequence shows the type of structural change revision can make in one trajectory, while Table~\ref{tab:stages} and Appendix~\ref{app:refine} give aggregate score changes.

\subsection{Proxy Mismatch Example}
\label{app:proxy-mismatch-9page}

An exploratory parameter-search variant (Appendix~\ref{app:cma}) provides a qualitative diagnostic for proxy mismatch.
Three separate runs for ``a red brick wall'' produce regular brick rows, a wall with broad breaks and irregular cracks, and a wavy texture without recognizable brick rows.
The authors' visual ordering differs from the recorded quick-score gains of $+2.0$, $+3.7$, and $+0.7$ points, with the largest quick-score increase assigned to the middle example.
This motivates inspecting structural changes alongside proxy scores, but it does not establish a general failure rate for parameter search or guarantee that seed-only search preserves appearance.

\end{document}